\documentclass[10pt,twocolumn,letterpaper]{article}

\usepackage{cvpr}
\usepackage{times}
\usepackage{epsfig}
\usepackage{graphicx}
\usepackage{amsmath}
\usepackage{amssymb}
\usepackage{subfig}
\usepackage{authblk}
\usepackage{cite}
\usepackage{comment}
\usepackage{adjustbox}
\usepackage{textcomp}
\usepackage{array,multirow,graphicx}
\usepackage[table]{xcolor}
\usepackage{colortbl}
\usepackage{bbm}
\usepackage{makecell}
\usepackage[percent]{overpic}

\usepackage[pagebackref=true,breaklinks=true,letterpaper=true,colorlinks,bookmarks=false]{hyperref}

\DeclareMathOperator*{\argmin}{argmin}

\cvprfinalcopy % *** Uncomment this line for the final submission

\def\cvprPaperID{75} % *** Enter the ICCV Paper ID here

\definecolor{TableRed}{HTML}{800000}
\newcommand{\TextRed}[1]{\textcolor{TableRed}{#1}}

\begin{document}

%%%%%%%%% TITLE
\title{Multi-style Generative Network for Real-time  Transfer}

\author[ ]{
Hang Zhang
}
\author[ ]{
Kristin Dana
}
\affil[ ]{Department of Electrical and Computer Engineering, Rutgers University}%, New Brunswick, USA}
\affil[ ]{ {\tt\small \ zhang.hang@rutgers.edu, kdana@ece.rutgers.edu }}
\renewcommand\Authsep{  } 
\renewcommand\Authand{  }
\renewcommand\Authands{  }

\maketitle

\begin{abstract}
% recent work one style, traditional not limited

Despite the rapid progress in style transfer, existing approaches using feed-forward generative network for multi-style or arbitrary-style transfer are usually compromised of image quality and model flexibility. 
We find it is fundamentally difficult to achieve comprehensive style modeling using 1-dimensional style embedding. 
Motivated by this, we introduce CoMatch Layer that learns to match the second order feature statistics with the target styles.  
With the CoMatch Layer, we build a Multi-style Generative Network (MSG-Net), which achieves real-time performance.  
In addition, we employ an specific strategy of upsampled convolution which avoids checkerboard artifacts caused by fractionally-strided convolution.
Our method has achieved superior image quality comparing to state-of-the-art approaches. The proposed MSG-Net as a general approach for real-time style transfer is compatible with most existing techniques including content-style interpolation, color-preserving, spatial control and brush stroke size control.   MSG-Net is the first to achieve real-time brush-size control in a purely feed-forward manner for style transfer. 
Our implementations and pre-trained models for Torch, PyTorch and MXNet frameworks will be publicly available\footnote{links can be found at \url{http://hangzh.com/}}. 

\end{abstract}

%section 1
\section{Introduction}

\begin{figure}
\subfloat[Input]
{
  \begin{overpic}[width=0.32\linewidth, height=0.2\linewidth]{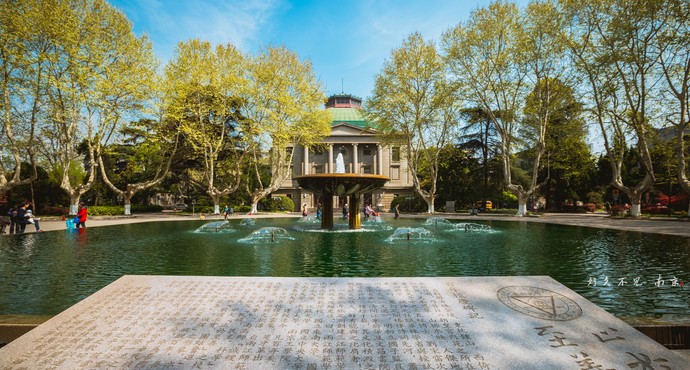}
     \put(70,32){\frame{\includegraphics[width=0.1\linewidth,height=0.1\linewidth]{figure/styles/candy.jpg}} }
  \end{overpic}
}
\subfloat[Mean \& Var]
{
\includegraphics[width=0.32\linewidth, height=0.2\linewidth]{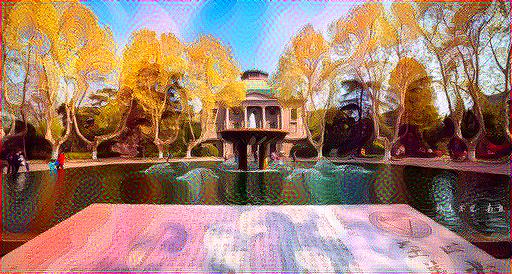}
}
\subfloat[Gram Matrix]
{
\includegraphics[width=0.32\linewidth, height=0.2\linewidth]{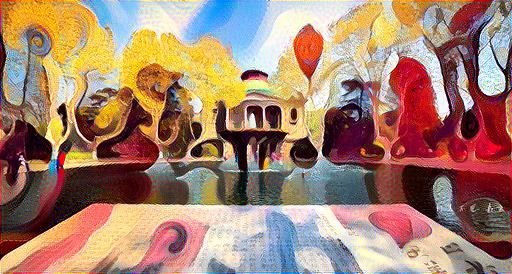}
}
\caption{Comparing 1D and 2D style representation using an optimization-based approach~\cite{gatys2016image}. (a) Input image and style. (b) Style transfer result minimizing difference of CNN featuremap mean and variance. (c) Style transfer result minimizing the difference in Gram matrix representation. }
\label{fig:1d2d}
\end{figure}

% what is style transfer, why it is interesting
Style transfer can be approached as reconstructing or synthesizing  texture  based on the target image semantic content~\cite{Li_2016_CVPR}. 
Many pioneering works have achieved success in classic texture synthesis  starting with methods that  resample pixels~\cite{efros1999texture,efros2001image,wei2000fast,kwatra2003graphcut} or match multi-scale feature statistics\cite{Debonet97,heeger1995pyramid,portilla2000parametric}.
These methods employ traditional image pyramids obtained by handcrafted multi-scale linear filter banks~\cite{simoncelli1995steerable,Burt83} and perform texture synthesis by matching the feature statistics to the target style. 
In recent years, the concepts of texture synthesis and  style transfer have been revisited within the context of deep learning.
Gatys {\it et al.}~\cite{gatys2015texture} 
shows that using feature correlations ({\it i.e.}\ Gram Matrix) of convolutional neural nets (CNN) successfully captures the image styles. 
%and provides an explicit representation that separates image style and content information. 
This framework has brought a surge of interest in texture synthesis and style transfer using iterative optimization ~\cite{gatys2015texture,gatys2016image,Li_2016_CVPR} or training feed-forward networks~\cite{ulyanov2016texture,Johnson2016Perceptual,li2016precomputed,ulyanov2017improved}.  
%This method is {\it optimization-based} because the new texture image is generated by applying gradient descent that manipulates a white noise image to match the Gram Matrix representation of the target image. 
%Optimization-based approaches have no scalability problem but require expensive computation to generate images using gradient descent.
Recent work extends  style flexibility using feed-forward networks and achieves multistyle or arbitrary style transfer~\cite{dumoulin2016learned,huang2017arbitrary,Chen2017StyleBank,chen2016fast}. 
These approaches typically encode image styles into 1-dimensional space, {\it i.e.}\ tuning the featuremap mean and variance (bias and scale) for different styles.  
However, the comprehensive appearance of image style is fundamentally difficult to represent in 1D embedding space. 
Figure~\ref{fig:1d2d} shows style transfer results using  the optimization-based approach~\cite{gatys2016image} and we can see Gram matrix representation produces more appealing image quality comparing to mean and variance of CNN featuremap. 

\begin{figure}
\centering
%%%%%%%%%%%%%%%%%%%%%%%%%%
\subfloat
{
  \begin{overpic}[height=0.217\linewidth]{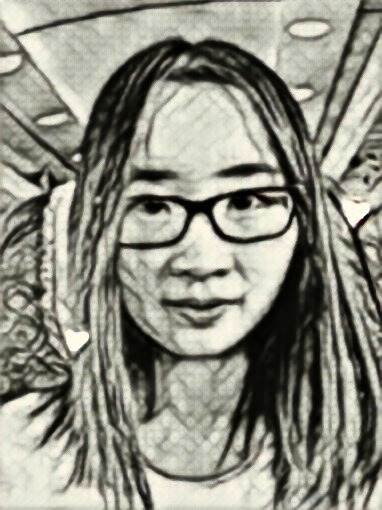}
     \put(45,70){\frame{\includegraphics[width=0.07\linewidth,height=0.07\linewidth]{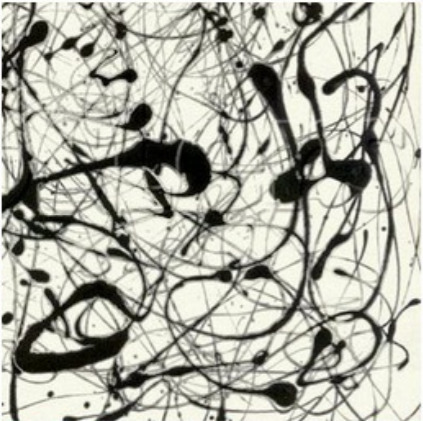}}}
  \end{overpic}
}
%%%%%%%%%%%%%%%%%%%%%%%%%%
\subfloat
{
  \begin{overpic}[height=0.217\linewidth]{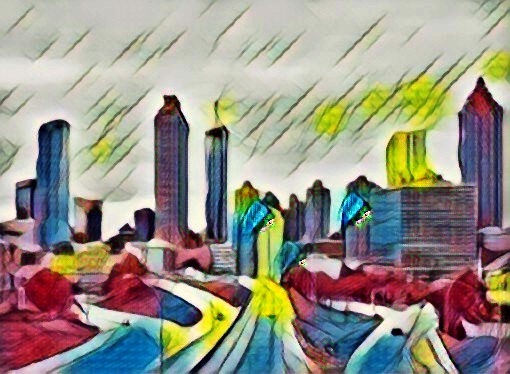}
     \put(80,52){\frame{\includegraphics[width=0.07\linewidth,height=0.07\linewidth]{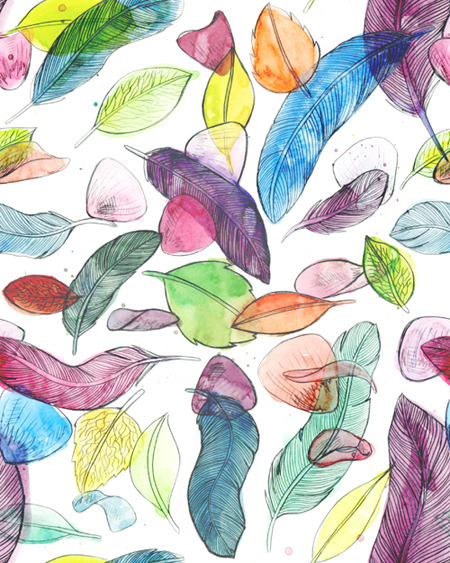}}}
  \end{overpic}
}
%%%%%%%%%%%%%%%%%%%%%%%%%%
\subfloat
{
  \begin{overpic}[height=0.217\linewidth]{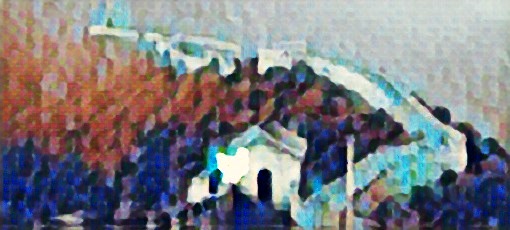}
     \put(85,31.5){\frame{\includegraphics[width=0.07\linewidth,height=0.07\linewidth]{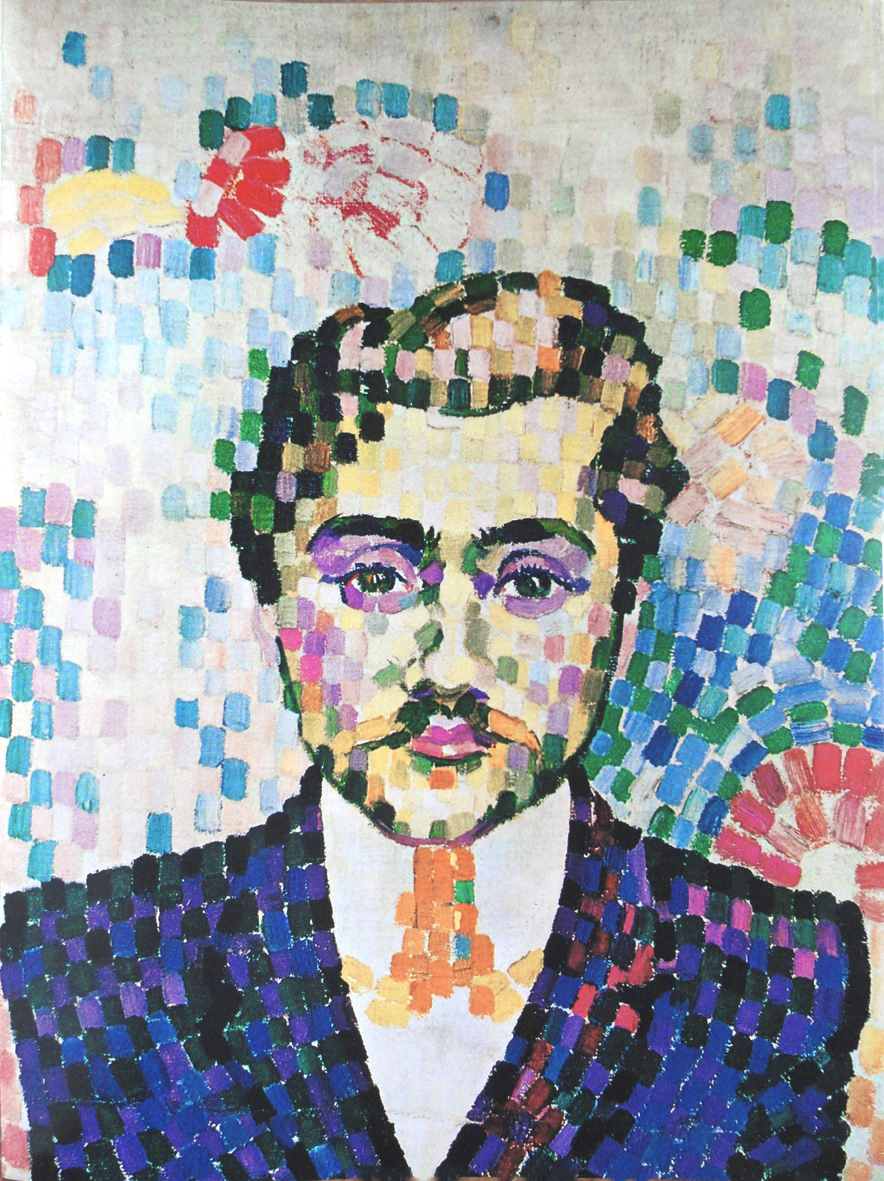}}}
  \end{overpic}
}
%%%%%%%%%%%%%%%%%%%%%%%%%%
\par
\vspace{-0.7em}
\subfloat
{
  \begin{overpic}[height=0.21\linewidth]{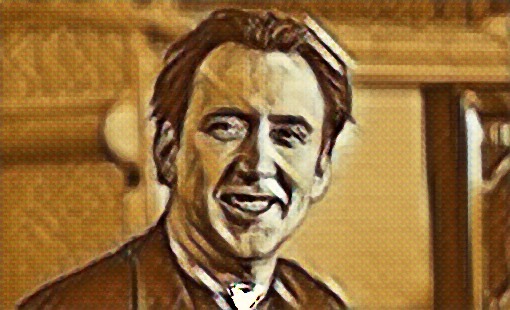}
     \put(82,43){\frame{\includegraphics[width=0.07\linewidth,height=0.07\linewidth]{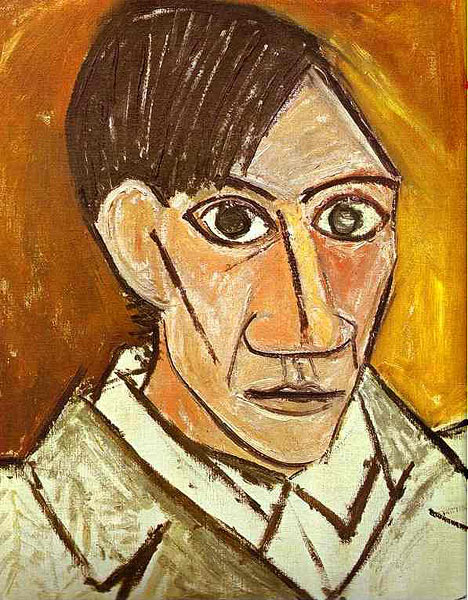}}}
  \end{overpic}
}
%%%%%%%%%%%%%%%%%%%%%%%%%%
\subfloat
{
  \begin{overpic}[height=0.21\linewidth]{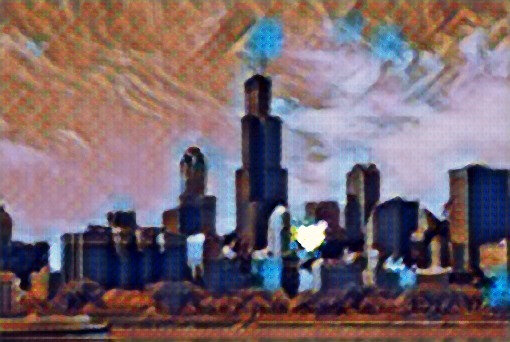}
     \put(78,47){\frame{\includegraphics[width=0.07\linewidth,height=0.07\linewidth]{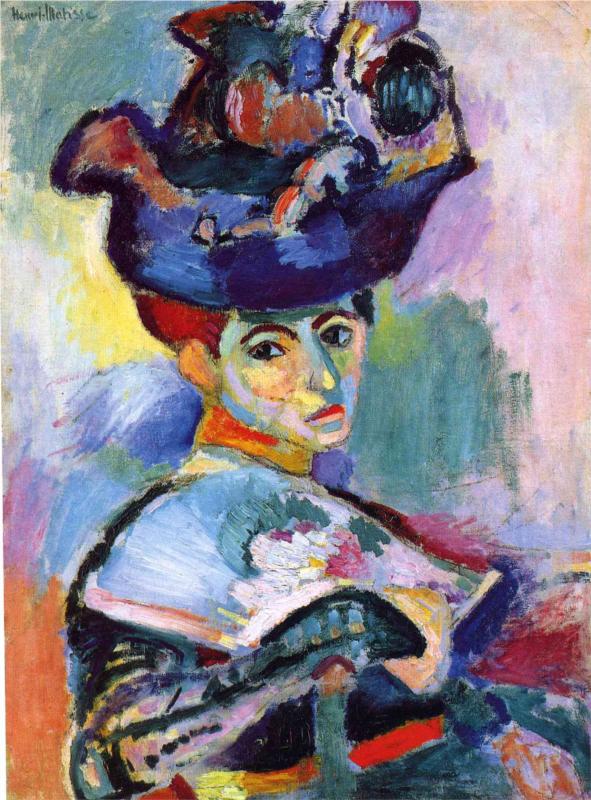}}}
  \end{overpic}
}
%%%%%%%%%%%%%%%%%%%%%%%%%%
\subfloat
{
  \begin{overpic}[height=0.21\linewidth]{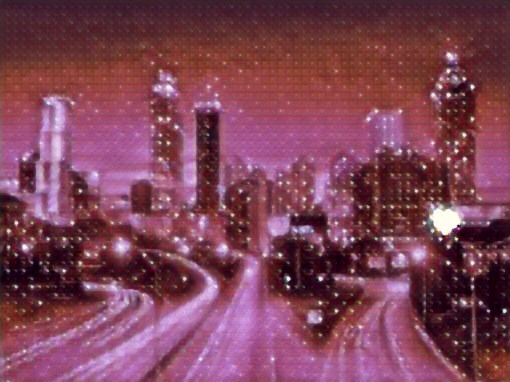}
     \put(75,52){\frame{\includegraphics[width=0.07\linewidth,height=0.07\linewidth]{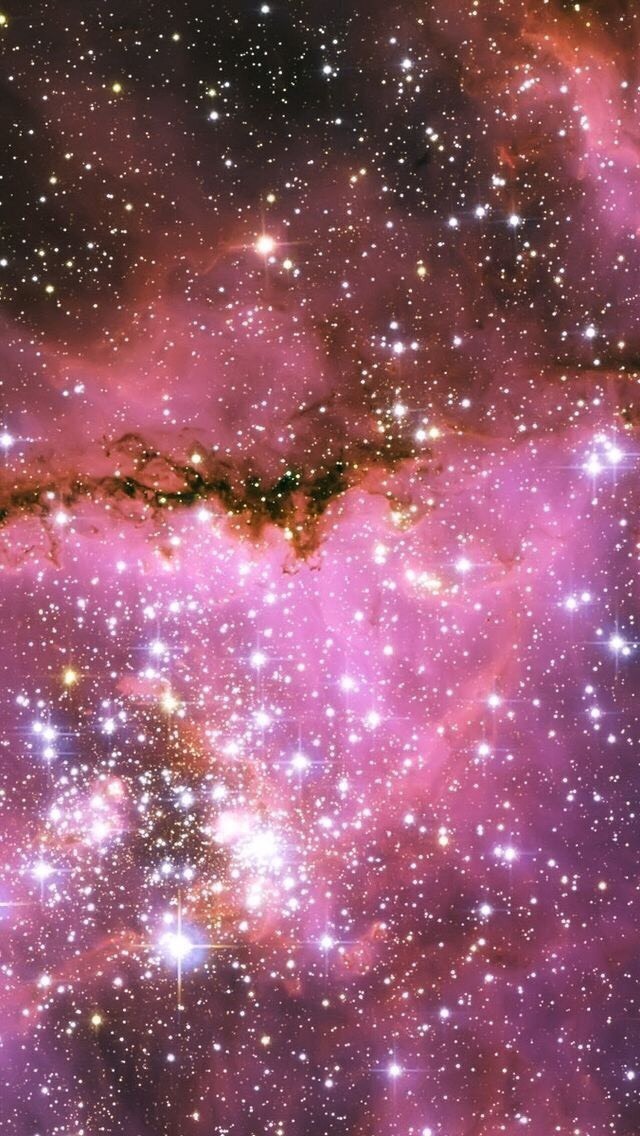}}}
  \end{overpic}
}
%%%%%%%%%%%%%%%%%%%%%%%%%%
\par
\vspace{-0.7em}
\subfloat
{
  \begin{overpic}[height=0.19\linewidth]{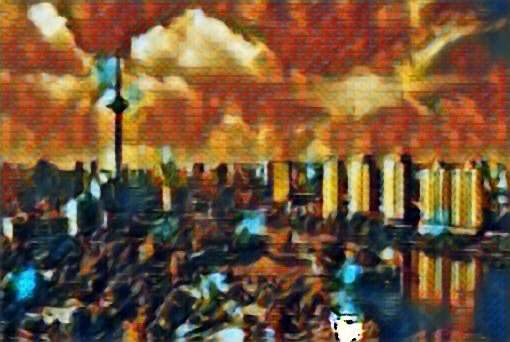}
     \put(77,44){\frame{\includegraphics[width=0.07\linewidth,height=0.07\linewidth]{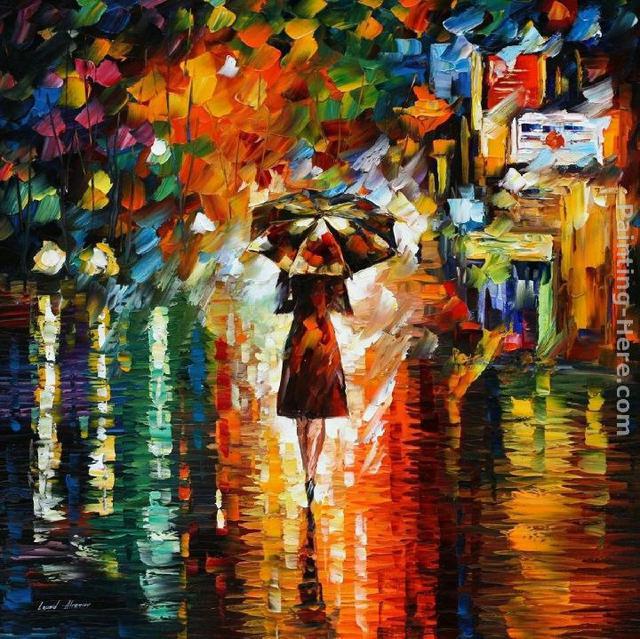}}}
  \end{overpic}
}
%%%%%%%%%%%%%%%%%%%%%%%%%%
\subfloat
{
  \begin{overpic}[height=0.19\linewidth]{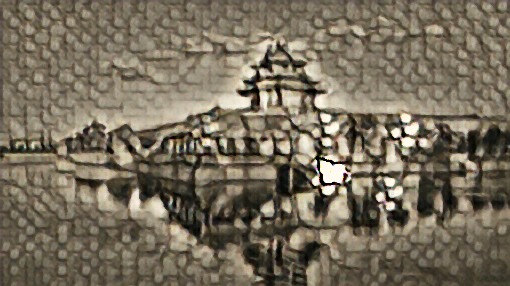}
     \put(85,37){\frame{\includegraphics[width=0.07\linewidth,height=0.07\linewidth]{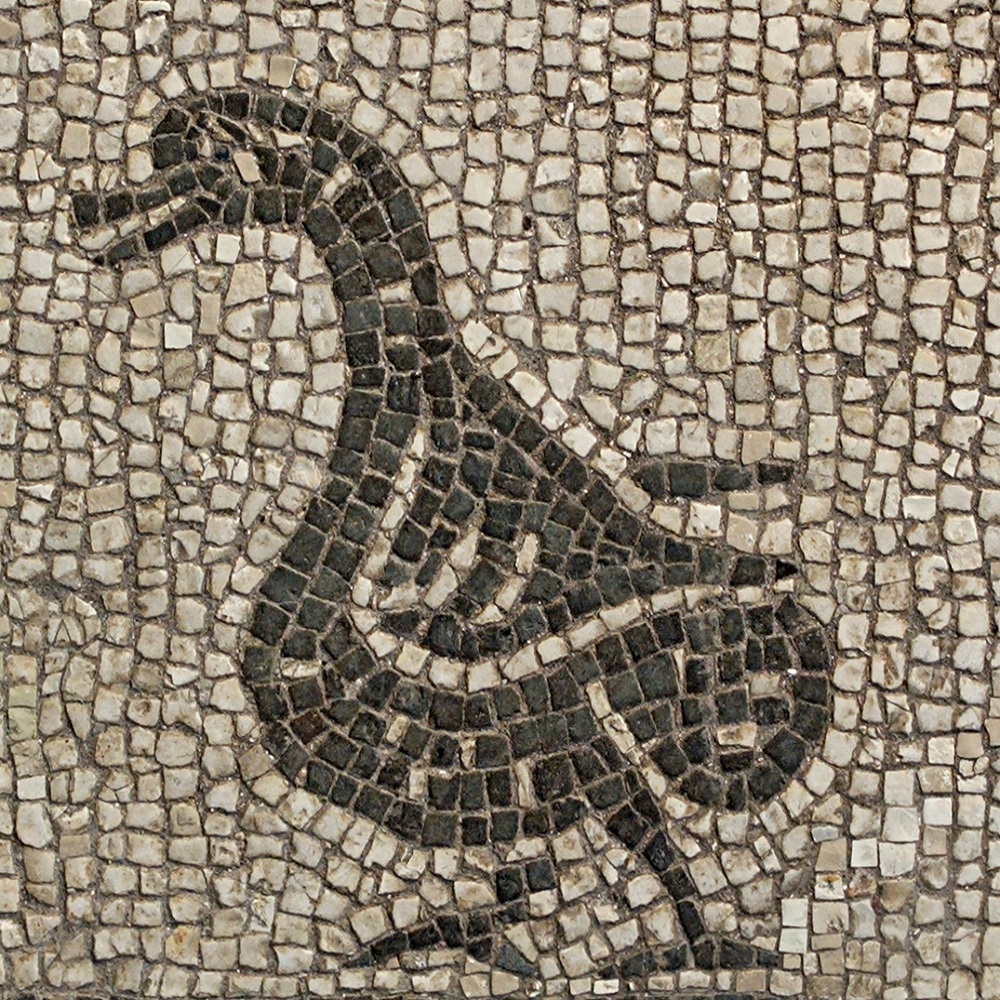}}}
  \end{overpic}
}
\subfloat
{
  \begin{overpic}[height=0.19\linewidth]{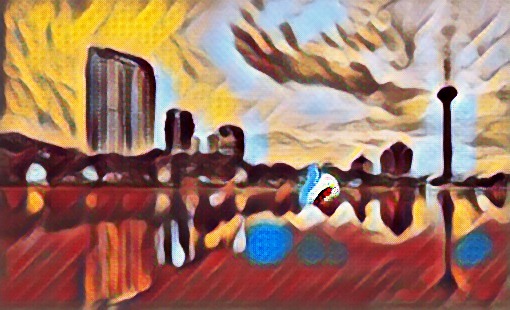}
     \put(78,40){\frame{\includegraphics[width=0.07\linewidth,height=0.07\linewidth]{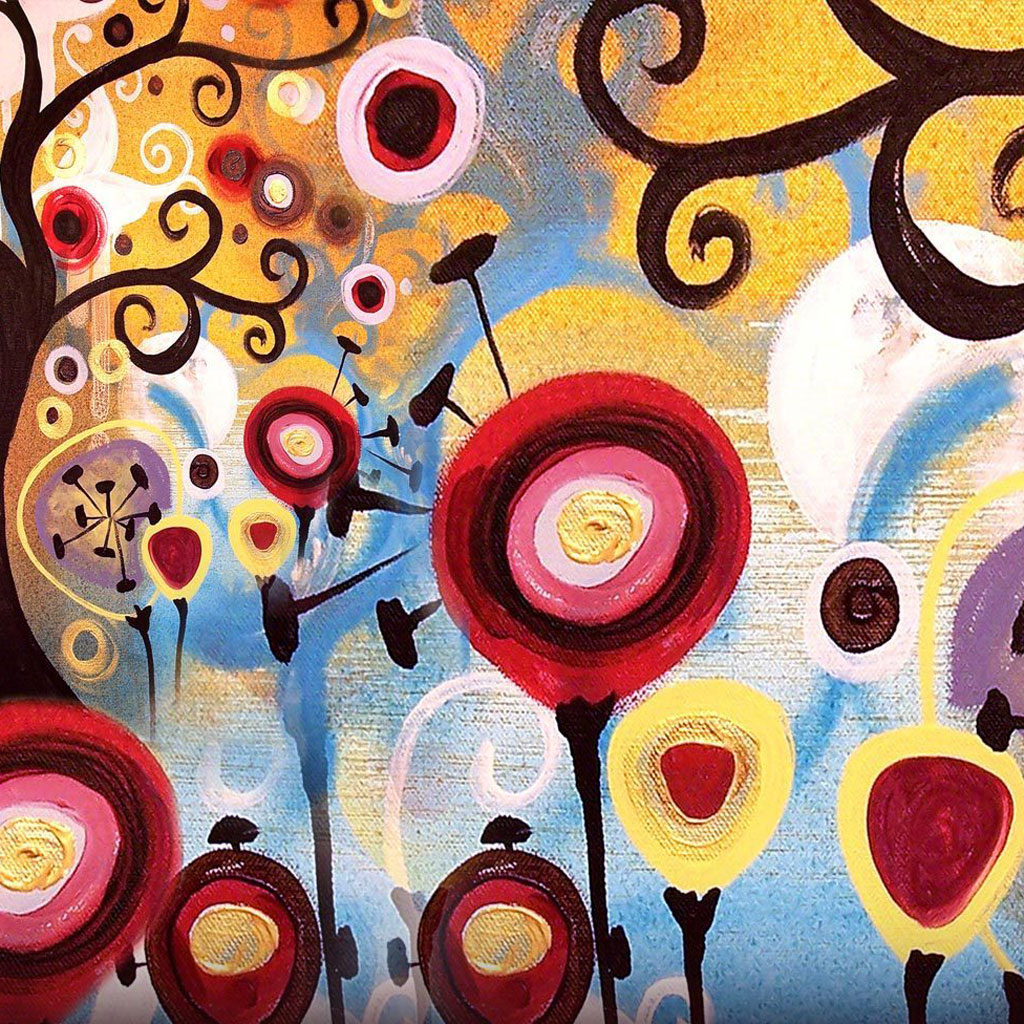}}}
  \end{overpic}
}
%%%%%%%%%%%%%%%%%%%%%%%%%%
\caption{%The problem of multi-style transfer in real-time using a {\bf single} network is solved in this paper. 
Examples of transferred images and the corresponding styles using the proposed MSG-Net. 
}
\end{figure}

\begin{figure*}
\centering
\includegraphics[width=0.9\linewidth,height=0.3\linewidth]{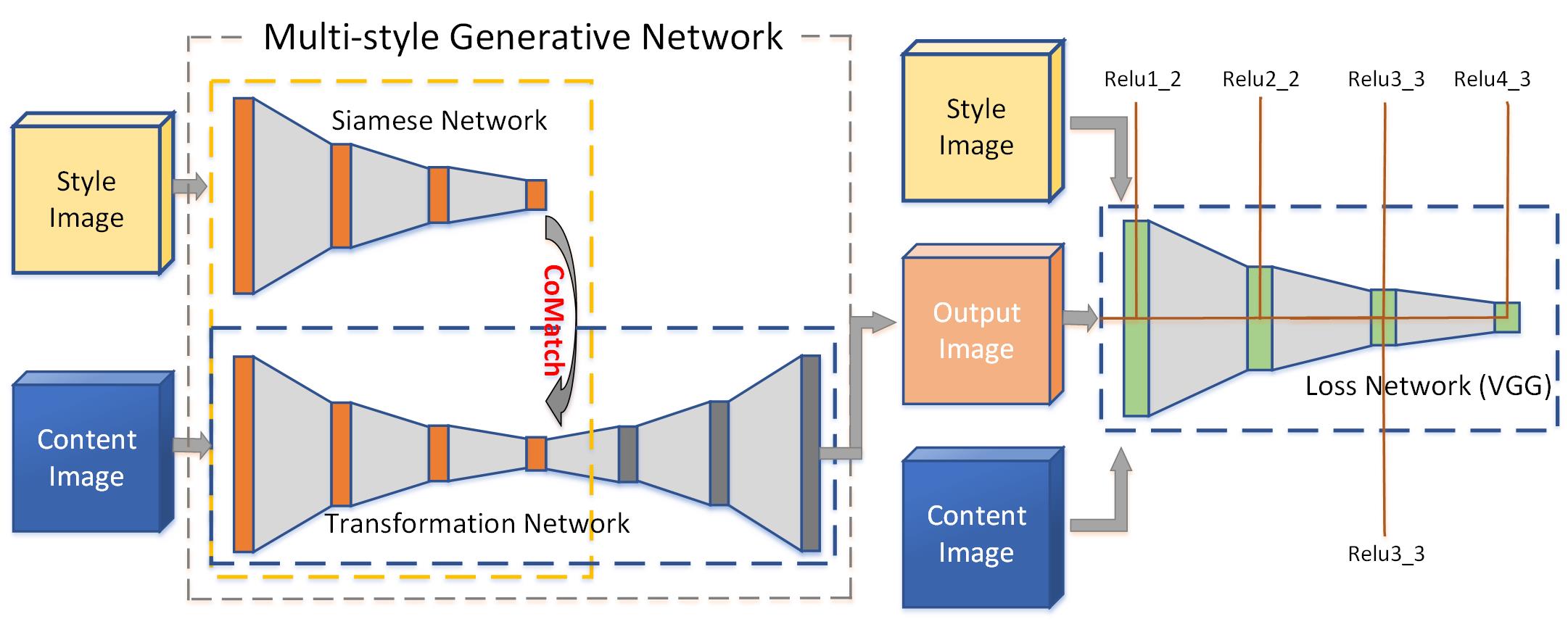}
\caption{An overview of MSG-Net, Multi-style Generative Network. The transformation network explicitly matches the features statistics of the style targets captured by a Siamese network 
using the proposed CoMatch Layer  (introduced in Section~\ref{sec:3}). %Detailed architecture of the transformation network is shown in Table~\ref{tab:msg-net}. 
A pre-trained loss network provides the supervision of MSG-Net learning by minimizing the content and style differences with the targets as discussed in Section~\ref{sec:learning}.}
\label{fig:overview}
\end{figure*}

%{\it What limits the diversity of styles in existing generative network? } Existing work learns a generative network taking an input image $x_c$ and providing the transferred output $G(x_c)$, in which the feature statistics of the style image are implicitly learned from the loss function without informing the network about the style target $x_s$~\cite{ulyanov2016texture,Johnson2016Perceptual,li2016precomputed}. 
%For existing approaches, there is a fundamental difficulty in deriving a representation that simultaneously preserves the semantic content of input image $x_c$ and matches the style of the target image $x_s$ (see extended discussion in Section~\ref{sec:relation}). In order to build a multi-style generative network $G(x_c,x_s)$, where $x_s$ can be chosen from a diverse set of styles, the generator network should explicitly match the feature statistics of the style target images at run time.

In addition to the image quality, concerns about the flexibility of current feed-forward generative models have been raised in Jing {\it et al.}~\cite{jing2017neural}, and they point out that no generative methods can adjust the brush stroke size in real-time. Feeding the generative network with high-resolution content image usually results in unsatisfying images as shown in Figure~\ref{fig:brush}. 
The generative network as a fully convolutional network (FCN) can accept arbitrary input image sizes. Resizing the style image  changes the relative brush size and the multistyle generative network matching the image style at run-time should naturally enable brush-size control by changing the input style image size. 
{\it What limits the current generative model from being aware of the brush size? } The 1D style embedding (featuremap mean and variance) fundamentally limits the potential of exploring finer behavior for style representations. 
Therefore, a 2D method is desired for finer representation of image styles. 

As the {\bf first contribution} of the paper, we introduce an {\it CoMatch Layer} which embeds style with a 2D representation and learns to match the second-order feature statistics (Gram Matrix) of the style targets inherently during the training. 
The CoMatch Layer is differentiable and end-to-end learnable with existing generative network architectures without additional supervision.    
The proposed CoMatch Layer enables multi-style generation from a single feed-forward network. 

% contributions
The {\bf second contribution} of this paper is building {\it Multi-style Generative  Network (MSG-Net)} with the proposed CoMatch Layer and a novel {\it Upsample Convolution}. The MSG-Net as a feed-forward network runs in real-time after training. 
Generative networks typically have a decoder part recovering the image details from downsampled representations. Learning fractionally-strided convolution~\cite{long2015fully} typically brings checkerboard artifacts. 
For improving the image quality, we employ a strategy we call {\it upsampled convolution}, which successfully avoids the checkerboard artifacts by applying an integer stride convolution and outputs an upsampled featuremap (details in Section~\ref{sec:net}). 
In addition, we extend the Bottleneck architecture~\cite{he2016deep} to an {\it Upsampling Residual Block}, which reduces computational complexity without losing style versatility by preserving larger number of channels. Passing identity all the way through the generative network enables the network to extend deeper and converge faster. 
The experimental results show that MSG-Net has achieved superior image fidelity and test speed compared to previous work. 
We also study the scalability of the model by extending 100-style MSG-Net to 1K styles using a larger model size and longer training time, and we observe no obvious quality differences. 
In addition, MSG-Net as a general multi-style strategy is compatible to most existing techniques and progress in style transfer, such as content style trade-off and interpolation~\cite{dumoulin2016learned}, spatial control, color preserving and brush-size control~\cite{gatys2016preserving,gatys2016controlling}. 

To our  knowledge, MSG-Net is the first to achieve real-time brush-size control in a purely feed-forward manner for multistyle transfer. 

%The paper is organized as follows. We briefly describe the prior work on content and style representation using a CNN framework in Section~\ref{sec:2}. We introduce our proposed CoMatch Layer in Section~\ref{sec:3} and our novel generative architecture Multi-style Generative Network in Section~\ref{sec:4}. The comparison to other approaches is discussed in Section~\ref{sec:relation}. Finally, the experimental results and comparisons are presented in Section~\ref{sec:5}. 

\subsection{Related Work}
\label{sec:relation}

\paragraph{Relation to Pyramid Matching. }
Early methods for texture synthesis were developed using multi-scale image pyramids \cite{heeger1995pyramid, Debonet97, wei2000fast, portilla2000parametric}. 
The discovery in these earlier methods was that realistic texture images could be synthesized from manipulating a white noise image %could be manipulated 
so that its feature statistics %(characterized by either histograms  or conditional histograms) 
were matched with the target at each pyramid level. 
Our approach is inspired by classic	 methods, which match feature statistics within the feed-forward network, 
but it  leverages the advantages of deep learning networks while placing the computational costs into the training process (feed-forward vs. optimization-based). 
%but it puts the computation burden  into the training process.

\paragraph{Relation to Fusion Layers. } Our proposed CoMatch Layer is a kind of fusion layer that takes two inputs (content and style representations). Current work in  fusion layers with CNNs include feature map concatenation and element-wise sum ~\cite{isola2016image,zhang2016stackgan,feichtenhofer2016convolutional}. However, these approaches are not directly applicable,  since there is no  separation of style from content. For style transfer, the generated images should not carry  semantic information of the style target nor styles of the content image. 
%In addition, input representation size  must match in prior fusion methods; but for style transfer, the content representation has the dimension of ${C_i\times H_i\times W_i}$ and the orderless style representation (Gram Matrix) has the dimension of $C_i\times C_i$.

\paragraph{Relation to Generative Adversarial Training. } 
The Generative Adversarial Network (GAN)~\cite{goodfellow2014generative}, which  jointly trains an adversarial  generator and discriminator simultaneously,  has  catalyzed a surge of interest in  the study of image generation~\cite{zhang2016stackgan,isola2016image,che2016mode,radford2015unsupervised,huang2016sgan,sindagi2017generating,zhang2017image}. 
Recent work on image-to-image GAN~\cite{isola2016image} adopts a conditional GAN to provide a general solution for some image-to-image generation problems. For those problems, it was previously hard to define a loss function. 
However, the style transfer problem cannot be tackled using the conditional GAN framework, due to missing ground-truth image pairs. 
Instead, we follow the work~\cite{ulyanov2016texture,Johnson2016Perceptual} to adopt a discriminator/loss network that minimizes the perceptual difference of synthesized images with content and style targets and provides the supervision of the generative network learning. 
The initial idea of employing Gram Matrix to trigger style synthesis is inspired by a recent work~\cite{che2016mode} that suggests using an encoder instead of random vector in GAN framework.

\paragraph{Recent Work in Multiple or Arbitrary Style Transfer.} Recent/concurrent work explores multiple or arbitrary style transfer~\cite{dumoulin2016learned,chen2016fast,huang2017arbitrary,dumoulin2016learned}. A style swap layer is proposed in~\cite{chen2016fast}, but gets lower quality and slower speed (compared to existing feed-forward approaches). 
An adaptive instance normalization is introduced in ~\cite{huang2017arbitrary} to match the mean and variance of the feature maps with the style target. Instead, our CoMatch Layer matches the second order statistics of Gram Matrices for the feature maps. 
We also explore the scalability of our approach in the Experiment Section~\ref{sec:5}.

%section 2 & 3

\section{Content and Style Representation}
\label{sec:2}

% the figure shows the features statistics form different levels are useful
%\input{sections/4lvls}

CNNs pre-trained on a very large dataset such as ImageNet can be regarded as descriptive representations of image statistics containing both semantic content and style information. 
Gatys {\it et\ al.}~\cite{gatys2016image} provides explicit representations that independently model the image content and style from CNNs, which we briefly describe in this section for completeness. 

The semantic content of the image can be represented as the activations of the descriptive network at $i$-th scale $\mathcal{F}^i(x)\in\mathbb{R}^{C_i\times H_i\times W_i}$ with a given the input image $x$, where the $C_i$, $H_i$ and $W_i$ are the number of feature map channels, feature map height and width. 
The texture or style of the image can be represented as the distribution of the features using Gram Matrix $\mathcal{G}(\mathcal{F}^i(x))\in\mathbb{R}^{C_i\times C_i}$ given by
\begin{equation}
\mathcal{G}\left(\mathcal{F}^i(x)\right) = 
\sum_{h=1}^{H_i}\sum_{w=1}^{W_i}\mathcal{F}^i_{h,w}(x){\mathcal{F}^i_{h,w}(x)}^T .
\end{equation}
The Gram Matrix is orderless and describes the feature distributions. 
For zero-centered data, the Gram Matrix is the same as the covariance matrix  scaled by the number of elements $C_i\times H_i\times W_i$. It can 
 be calculated efficiently by first reshaping the feature map  $\Phi\left(\mathcal{F}^i(x)\right)\in\mathbb{R}^{C_i\times (H_iW_i)}$, where $\Phi()$ is a reshaping operation. Then the Gram Matrix can be written as
$\mathcal{G}\left(\mathcal{F}^i(x)\right) = \Phi\left(\mathcal{F}^i(x)\right)\Phi\left(\mathcal{F}^i(x)\right)^T$.

\begin{figure}[t]
\centering
\includegraphics[width=0.9\linewidth]{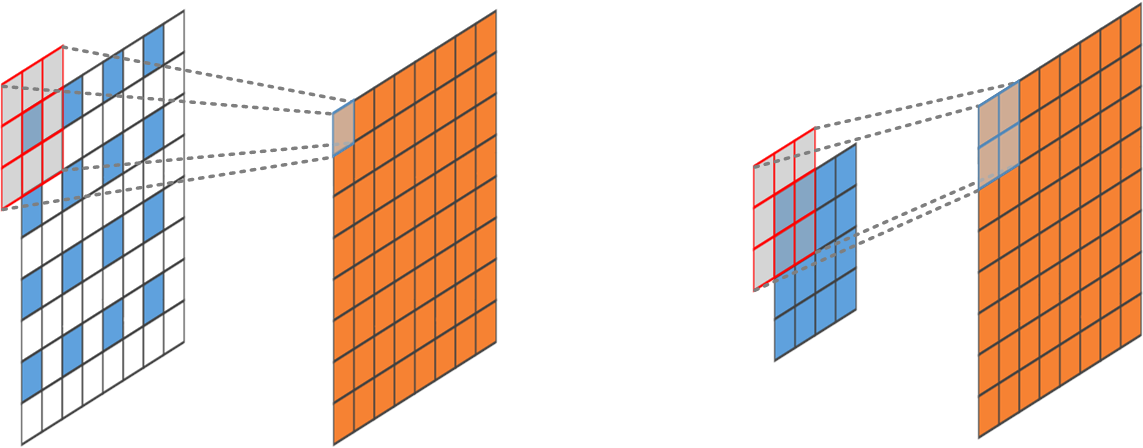}
\caption{Left: fractionally-strided convolution. Right: Upsampled convolution, which reduces the checkerboard artifacts by applying an integer stride convolution and outputting an upsampled featuremaps.  }
\label{fig:upconv}
\end{figure}

\section{CoMatch Layer}
\label{sec:3}
In this section, we introduce {\it CoMatch Layer}, which explicitly matches second order feature statistics based on the given styles. 
For a given content target $x_c$ and a style target $x_s$, 
the content and style representations at the $i$-th scale using the descriptive network can be written as $\mathcal{F}^i(x_c)$ and $\mathcal{G}(\mathcal{F}^i(x_s))$, respectively.
A direct solution $\hat{\mathcal{Y}^i}$ is desirable which preserves the semantic content of input image and matches the target style feature statistics: 
\begin{equation}
\begin{split}
\hat{\mathcal{Y}^i} = & \argmin_{\mathcal{Y}^i} \{ \|\mathcal{Y}^i-\mathcal{F}^i(x_c)\|^2_F \\ + 
& \alpha \|\mathcal{G}(\mathcal{Y}^i)-\mathcal{G}\left(\mathcal{F}^i(x_s)\right)\|^2_F \}.
\end{split}
\label{eq:model}
\end{equation}
where $\alpha$ is a trade-off parameter that balancing the contribution of the content and style targets.

\begin{figure}[t]
\centering
\includegraphics[width=0.9\linewidth]{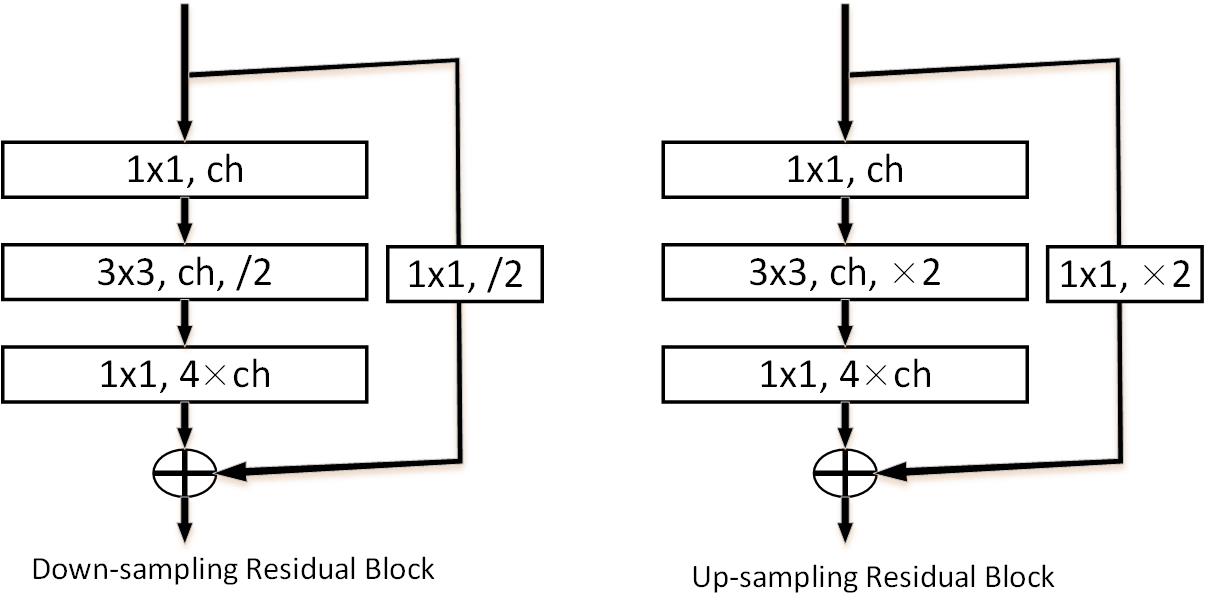}
\caption{We extend the original down-sampling residual architecture (left) to an up-sampling version (right). We use a 1$\times$1 fractionally-strided convolution as a shortcut and adopt reflectance padding. }
\label{fig:bottleneck}
\end{figure}

\begin{figure}
\subfloat
{
  \begin{overpic}[width=0.32\linewidth,height=0.2\linewidth]{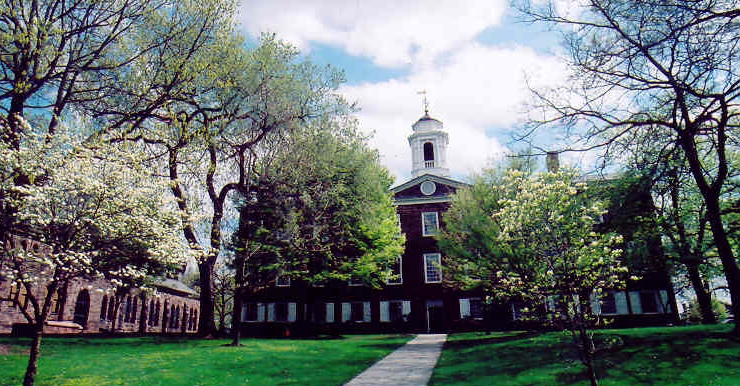}
     \put(62,23){\frame{\includegraphics[width=0.13\linewidth,height=0.13\linewidth]{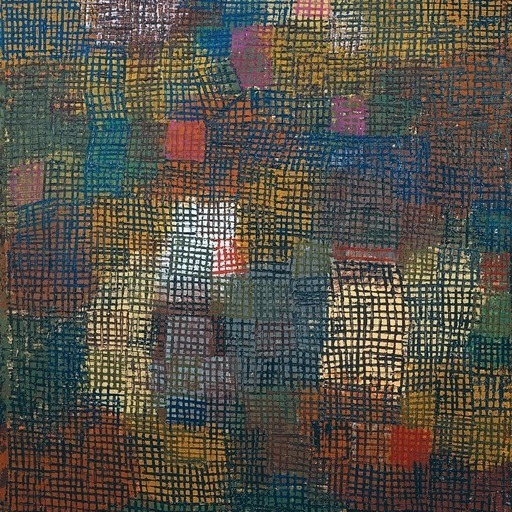}} }
  \end{overpic}
}
\subfloat
{
\includegraphics[width=0.32\linewidth,height=0.2\linewidth]{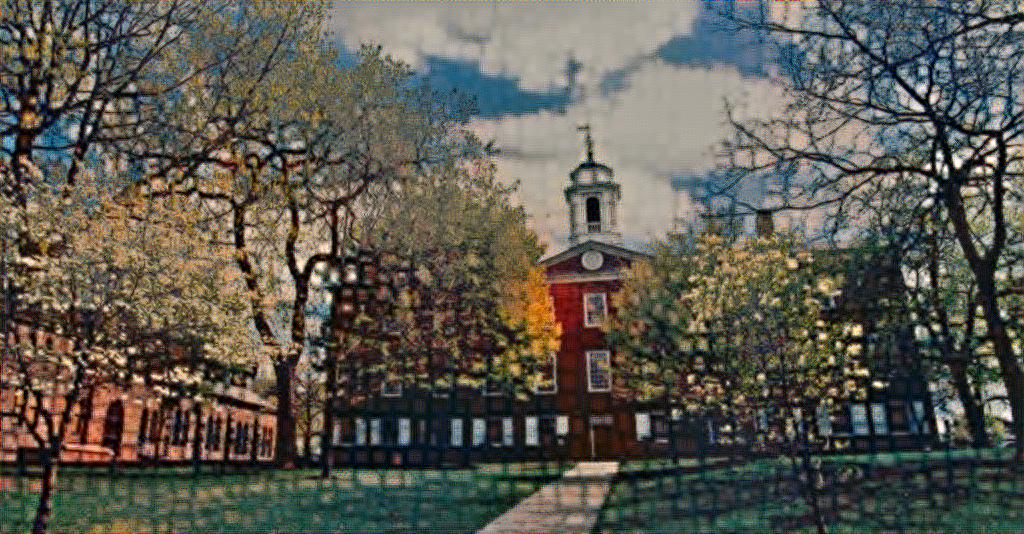}
}
\subfloat
{
\includegraphics[width=0.32\linewidth,height=0.2\linewidth]{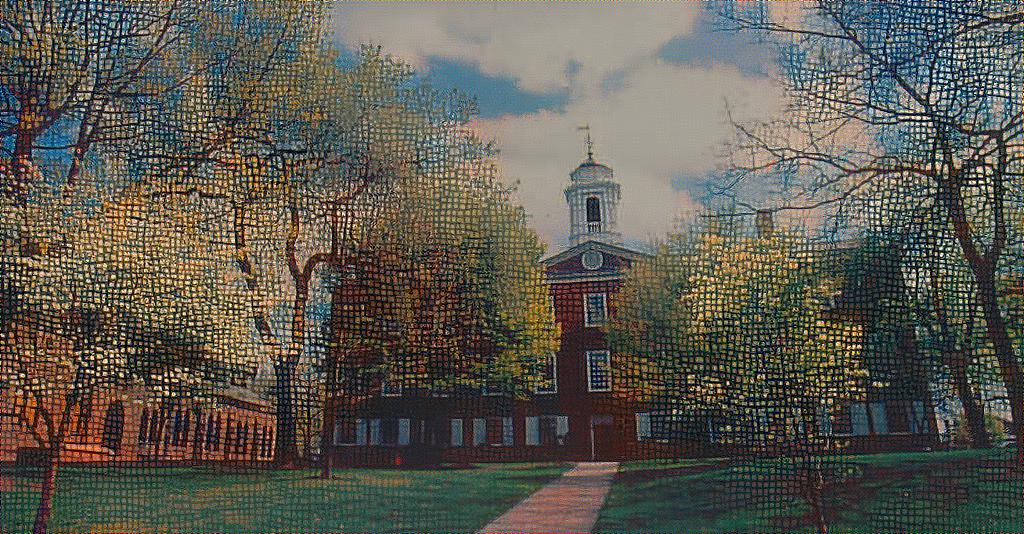}
}
\par
\vspace{-0.8em}
%%%%%%%%%%%%%%%%%%%%%%%%%%%55
\setcounter{subfigure}{0}
\subfloat[input]
{
  \begin{overpic}[width=0.32\linewidth]{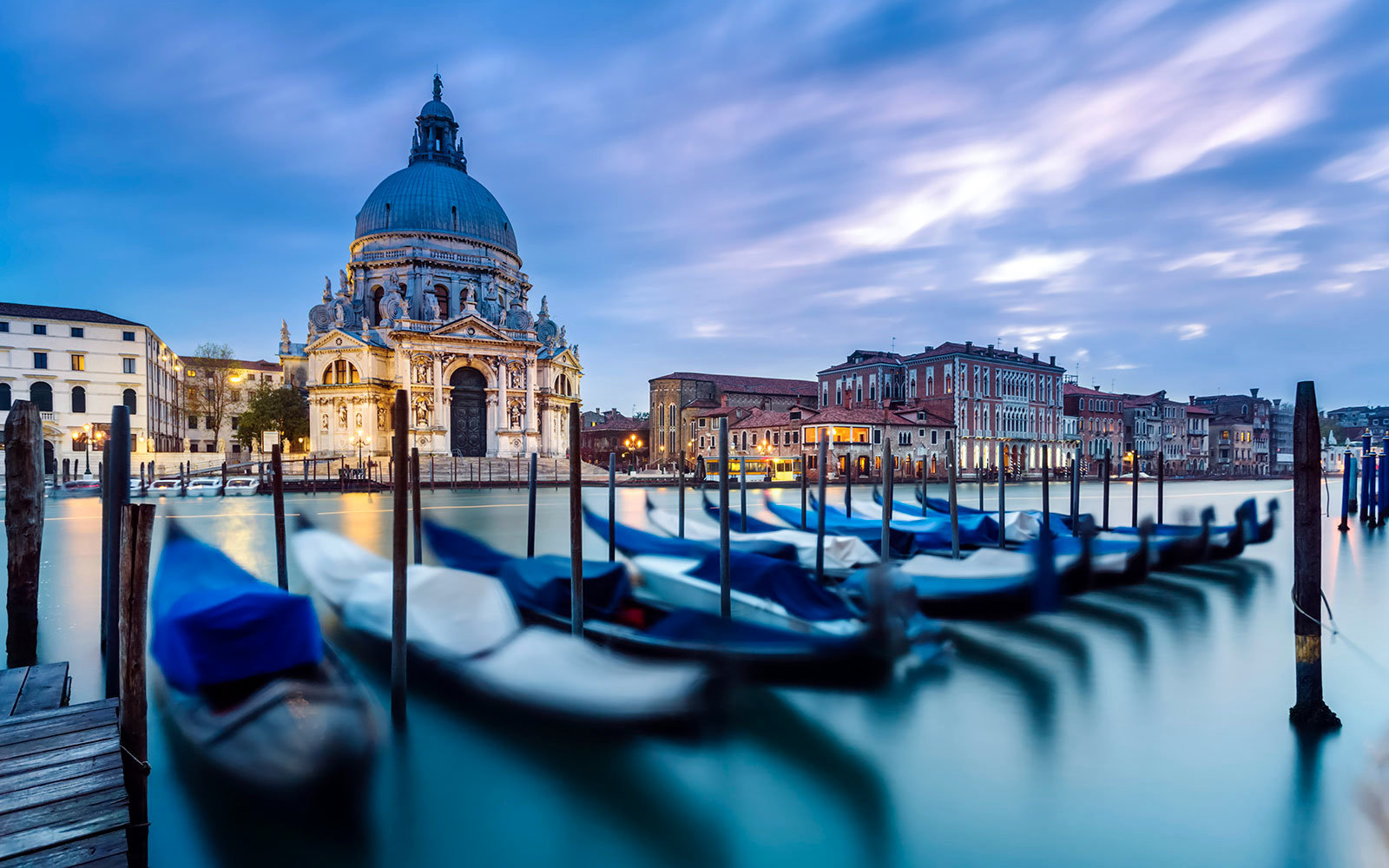}
     \put(62,23){\frame{\includegraphics[width=0.13\linewidth,height=0.13\linewidth]{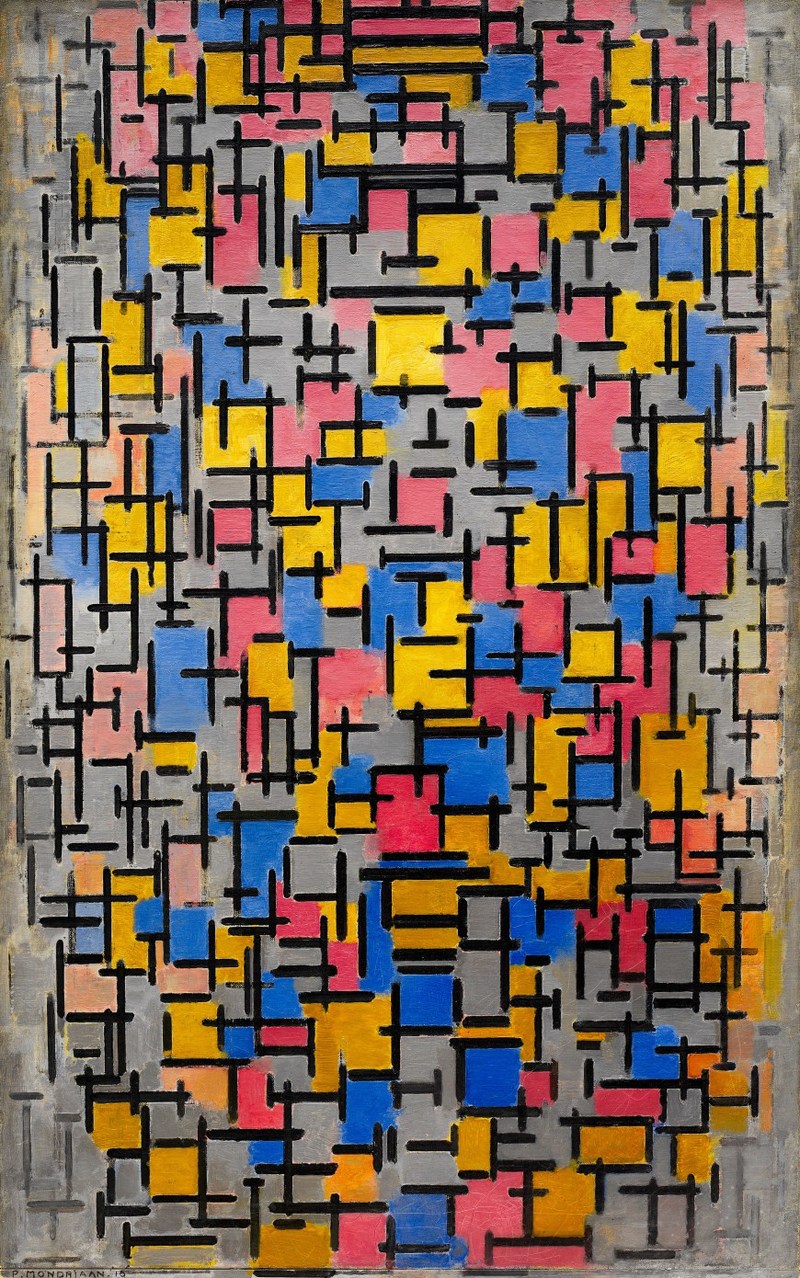}} }
  \end{overpic}
}
\subfloat[MSG-Net (\small\TextRed{ours})]
{
\includegraphics[width=0.32\linewidth]{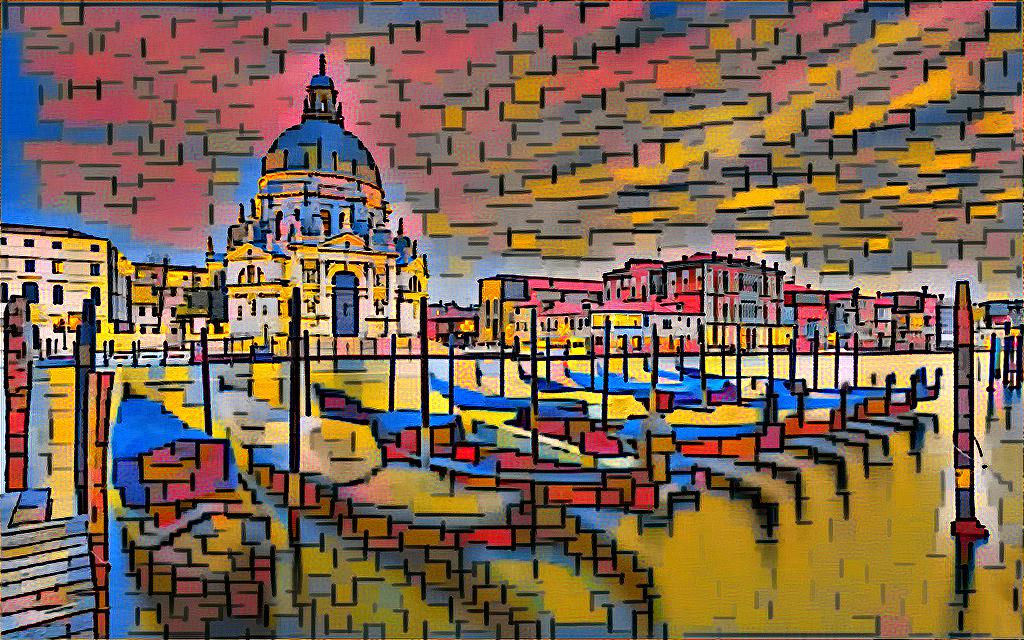}
}
\subfloat[baseline]
{
\includegraphics[width=0.32\linewidth]{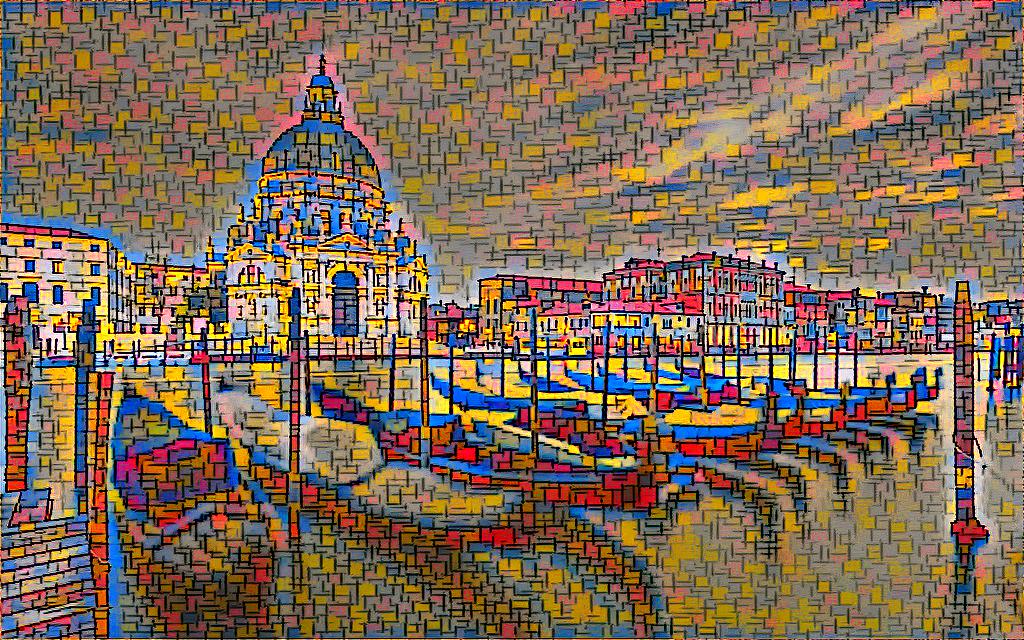}
}
\caption{Comparing Brush-size control. a) High-resolution input image and dense styles. b) Style transfer results using MSG-Net with brush-size control. c) Standard generative network~\cite{Johnson2016Perceptual} without brush-size control. See also Figure~\ref{fig:brush2}}
\label{fig:brush}
\vspace{-0.7em}
\end{figure}

The minimization of the above problem is solvable by using an iterative approach, but it is infeasible to achieve it in real-time or make the model differentiable. 
However, we can still approximate the solution and put the computational burden to the training stage. We introduce an approximation which tunes the feature map based on the target style:
\begin{equation}
\hat{\mathcal{Y}^i} = \Phi^{-1}\left[  \Phi\left(\mathcal{F}^i(x_c)\right)^TW\mathcal{G}\left(\mathcal{F}^i(x_s)\right) \right]^T,
\end{equation}
where $W\in\mathbb{R}^{C_i\times C_i}$ is a learnable weight matrix and $\Phi()$ is a reshaping operation to match the dimension, so that $\Phi\left(\mathcal{F}^i(x_c)\right)\in\mathbb{R}^{C_i\times(H_iW
_i)}$. For intuition on the functionality of $W$, suppose  $W={\mathcal{G}\left(\mathcal{F}^i(x_s)\right)}^{-1}$, %then $\mathcal{Y}^i=\mathcal{F}^i(x_c)$, 
then the  first term in Equation~\ref{eq:model} (content term) is minimized.  Now let $W=\Phi\left(\mathcal{F}^i(x_c)\right)^{-T}{\mathcal{L}(\mathcal{F}^i(x_s))}^{-1}$, where $\mathcal{L}\left(\mathcal{F}^i(x_s)\right)$ is obtained by the Cholesky Decomposition of $\mathcal{G}\left(\mathcal{F}^i(x_s)\right)=\mathcal{L}\left(\mathcal{F}^i(x_s)\right) {\mathcal{L}\left(\mathcal{F}^i(x_s)\right)}^T$, then the second term of Equation~\ref{eq:model} (style term) is minimized.
We let $W$  be learned directly from the loss function to dynamically balance the trade-off.
The CoMatch Layer is differentiable and can be inserted in the existing generative network and directly learned from the loss function without any additional supervision. 
%with respect to both layer input and the layer weights. 
%Therefore, the CoMatch Layer can be learned by standard Stochastic Gradient Decent (SGD) solver. %We provide explicit expression for the derived backpropagation equations in the supplementary material. 

%section 4
\section{Multi-style Generative Network}
\label{sec:4}

\subsection{Network Architecture}
\label{sec:net}

Prior feed-forward based single-style  transfer work learns a generator network that takes only the content image as the input and outputs the transferred image,  i.e. the generator network can be expressed as $G(x_c)$, which implicitly learns the feature statistics of the style image from the loss function. 
We introduce a {\it Multi-style Generative Network} which takes both content and style target as inputs.  i.e. $G(x_c, x_s)$.
The proposed network explicitly matches the feature statistics of the style targets at  runtime.

%Details about the network architecture:
%As part of the Generator Network, we adopt a 16-layer pre-trained VGG network~\cite{simonyan2014very} as a descriptive network $\mathcal{F}$ 
As part of the Generator Network, we adopt a Siamese network sharing weights with the encoder part of transformation network, 
which captures the feature statistics of the style image $x_s$ at different scales, and outputs the Gram Matrices $\{\mathcal{G}(\mathcal{F}^i(x_s))\} (i=1,...K)$ where $K$ is the total number of scales. 
Then a transformation network takes the content image $x_c$ and matches the feature statistics of the style image at multiple scales with CoMatch Layers.

\paragraph{Upsampled Convolution. } Standard CNN for image-to-image tasks typically adopts an encoder-decoder framework, because it is efficient to put heavy operations (style switching) in smaller featuremaps and also important to keep a larger receptive field for preserving semantic coherence. The decoder part learns a fractionally-strided convolution to recover the detail information from downsampled featuremaps. However, the fractionally strided convolution~\cite{long2015fully} typically introduces checkerboard artifacts~\cite{odena2016deconvolution}. Prior work suggests  using upsampling followed by convolution to replace the standard fractionally-strided convolution~\cite{odena2016deconvolution}. 
However, this strategy will decrease the receptive field and it is inefficient to apply convolution on an upsampled area.
For this, we use {\it upsampled convolution}, which has an integer stride, and outputs  upsampled featuremaps. For an upsampling factor of 2, the upsampled convolution will produce a 2$\times$2 outputs for each convolutional window as visualized in Figure~\ref{fig:upconv}.  Comparing to fractionally-strided convolution, this method has the same computation complexity and 4 times parameters. 
This strategy successfully avoid upsampling artifacts in the network decoder. 

\paragraph{Upsample Residual Block.} 
Deep residual learning has achieved great success in visual recognition~\cite{he2016deep,he2016identity}.
Residual block architecture plays an important role by reducing the computational complexity without losing diversity by preserving the large number of feature map channels. 
We extend the original architecture with an upsampling version as shown in Figure~\ref{fig:bottleneck} (right), which has a fractionally-strided convolution~\cite{long2015fully} as the shortcut and adopts reflectance padding to avoid artifacts of the generative process.
This upsampling residual architecture allows us to pass identity all the way through the network, %(as shown in Table~\ref{tab:msg-net})
 so that the network  converges faster and extends deeper. 

\begin{figure}
\subfloat
{
\includegraphics[width=0.18\linewidth,height=0.2\linewidth]{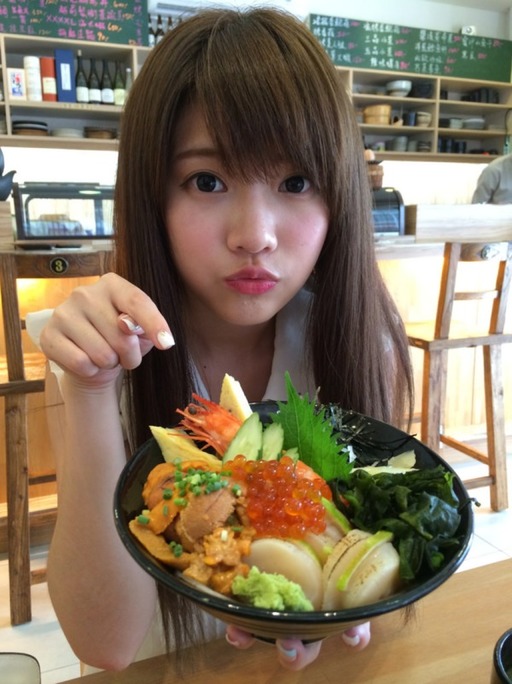}
}
\subfloat
{
\includegraphics[width=0.18\linewidth,height=0.2\linewidth]{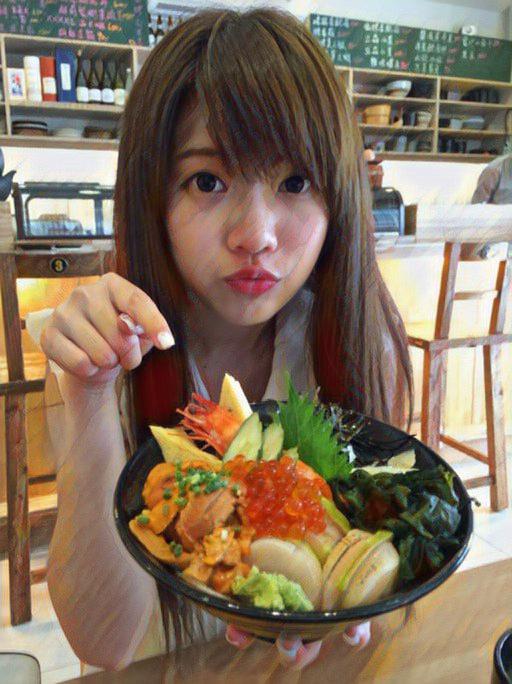}
}
\subfloat
{
\includegraphics[width=0.18\linewidth,height=0.2\linewidth]{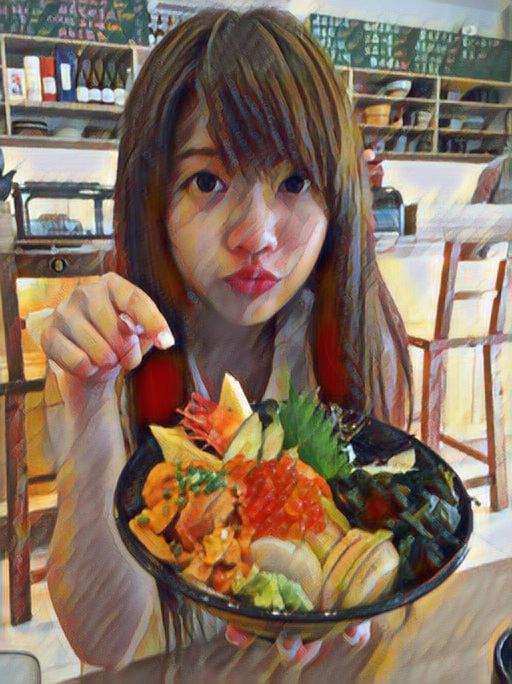}
}
\subfloat
{
\includegraphics[width=0.18\linewidth,height=0.2\linewidth]{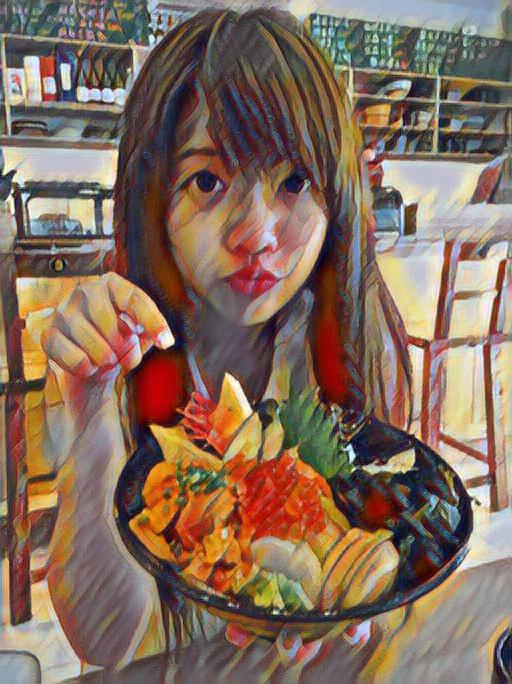}
}
\subfloat
{
\includegraphics[width=0.18\linewidth,height=0.2\linewidth]{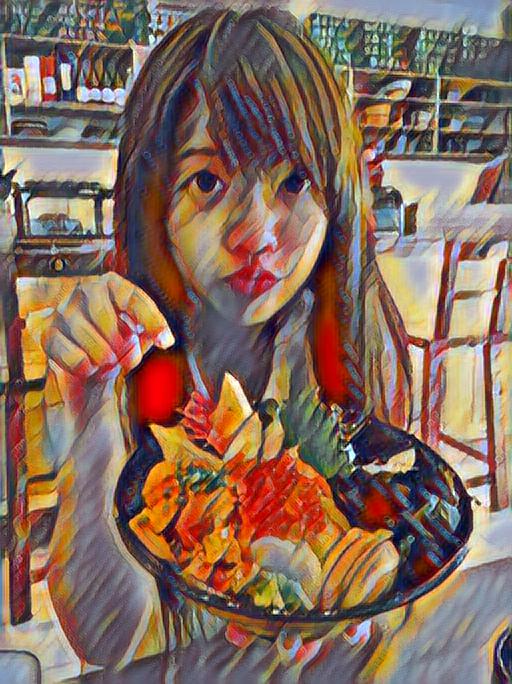}
}\par
\vspace{-0.8em}
%%%%%%%%%%%%%%%%%%%%%%%%%%%%%%%%%%%%%%%%%
\subfloat
{
\includegraphics[width=0.18\linewidth,height=0.2\linewidth]{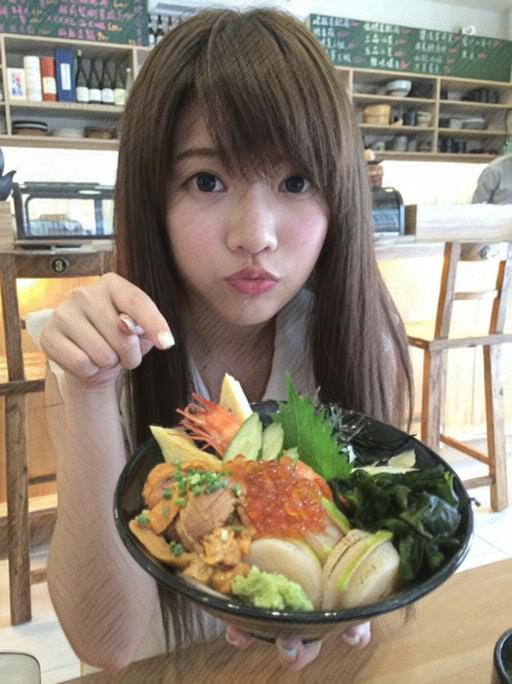}
}
\subfloat
{
\includegraphics[width=0.18\linewidth,height=0.2\linewidth]{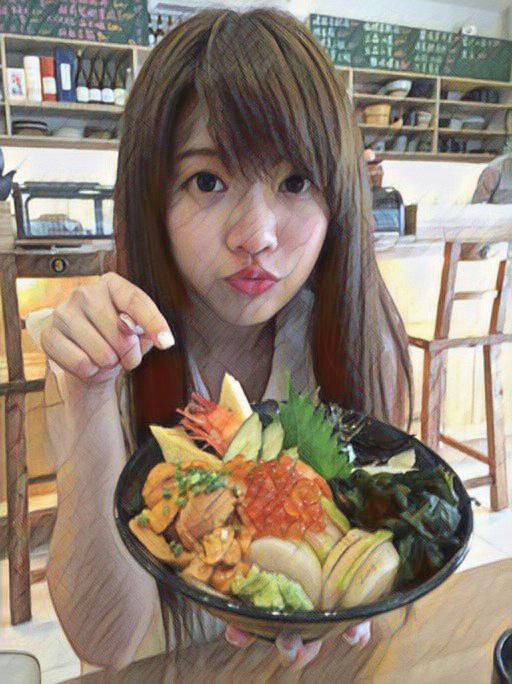}
}
\subfloat
{
\includegraphics[width=0.18\linewidth,height=0.2\linewidth]{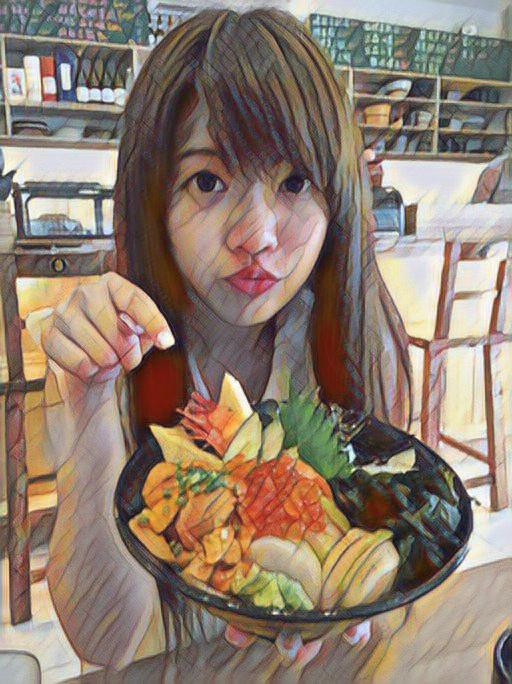}
}
\subfloat
{
\includegraphics[width=0.18\linewidth,height=0.2\linewidth]{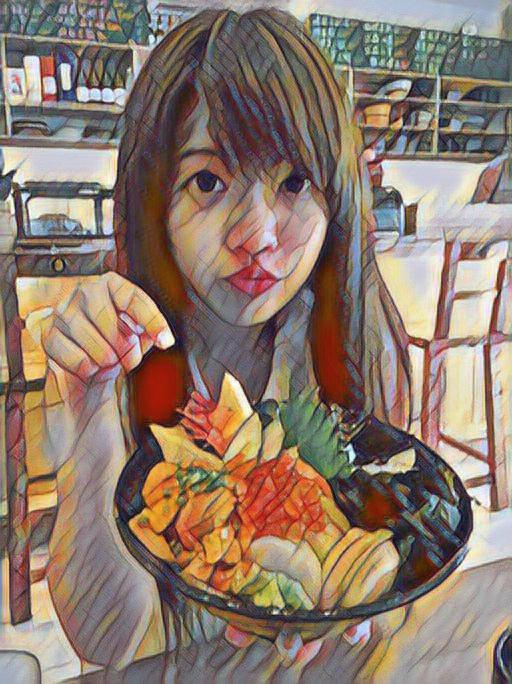}
}
\subfloat
{
\includegraphics[width=0.18\linewidth,height=0.2\linewidth]{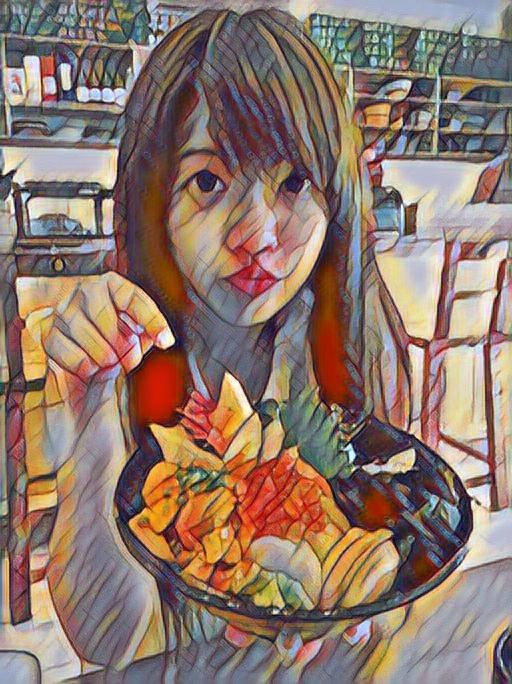}
}\par
\vspace{-0.8em}
%%%%%%%%%%%%%%%%%%%%%%%%%%%%%%%%%%%%%%%%%5
\subfloat
{
\includegraphics[width=0.18\linewidth,height=0.2\linewidth]{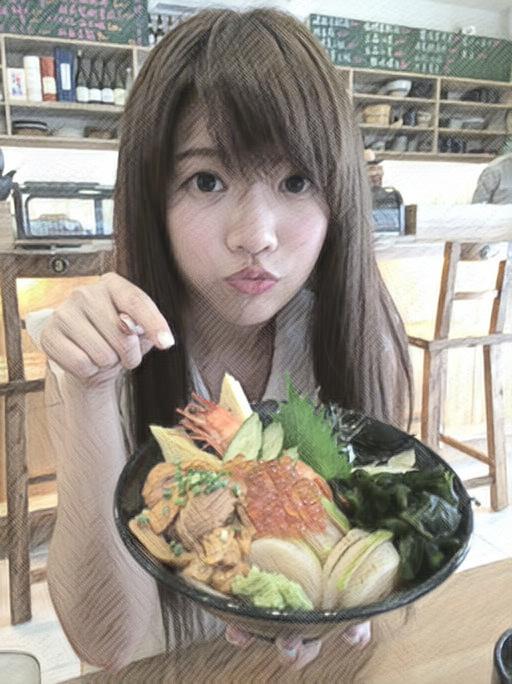}
}
\subfloat
{
\includegraphics[width=0.18\linewidth,height=0.2\linewidth]{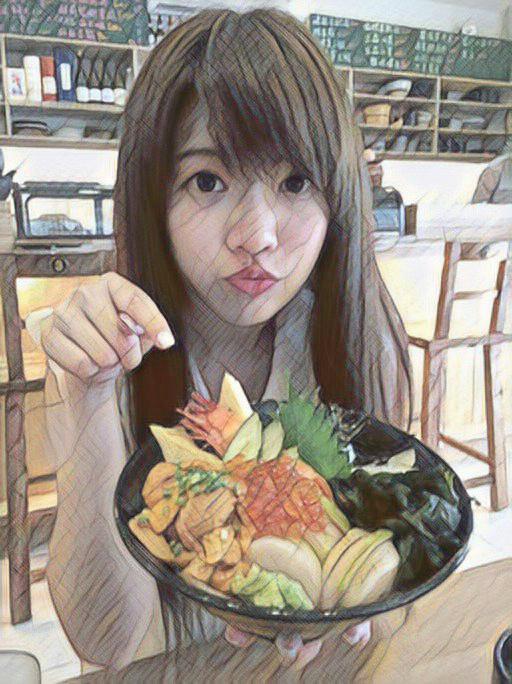}
}
\subfloat
{
\includegraphics[width=0.18\linewidth,height=0.2\linewidth]{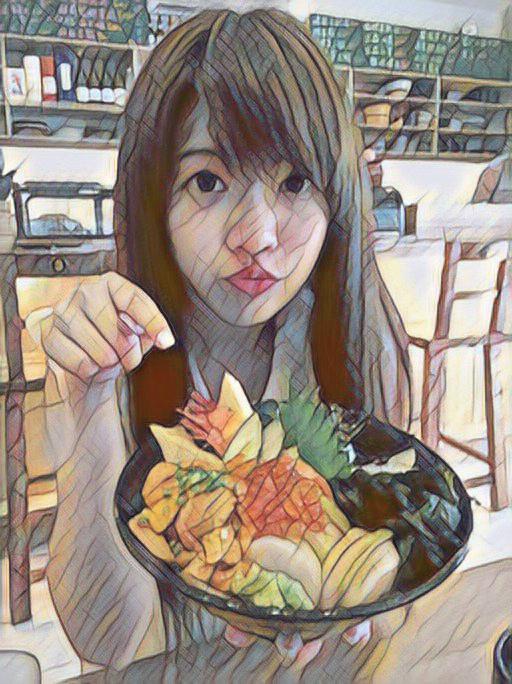}
}
\subfloat
{
\includegraphics[width=0.18\linewidth,height=0.2\linewidth]{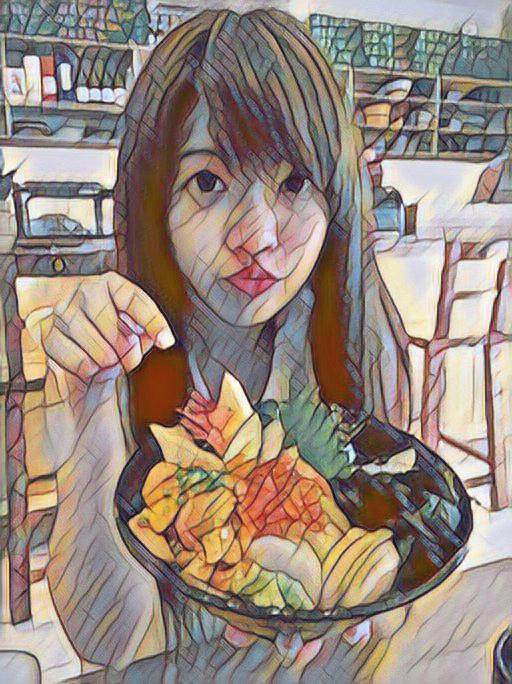}
}
\subfloat
{
\includegraphics[width=0.18\linewidth,height=0.2\linewidth]{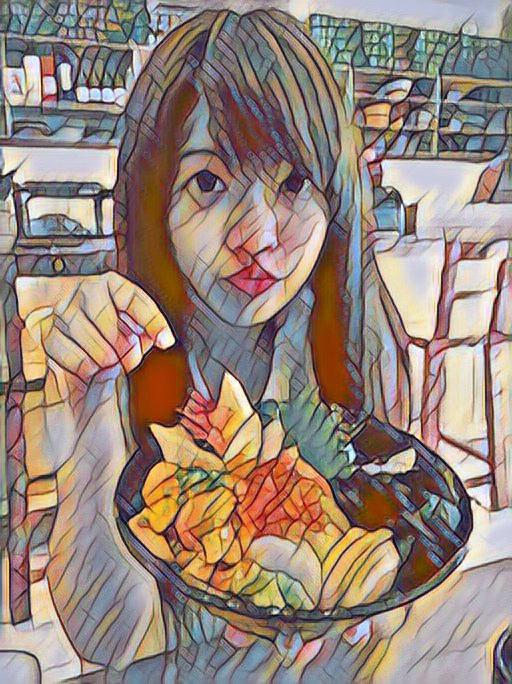}
}\par
\vspace{-0.8em}
%%%%%%%%%%%%%%%%%%%%%%%%%%%%%%%%%%%%%%%%%5
\subfloat
{
\includegraphics[width=0.18\linewidth,height=0.2\linewidth]{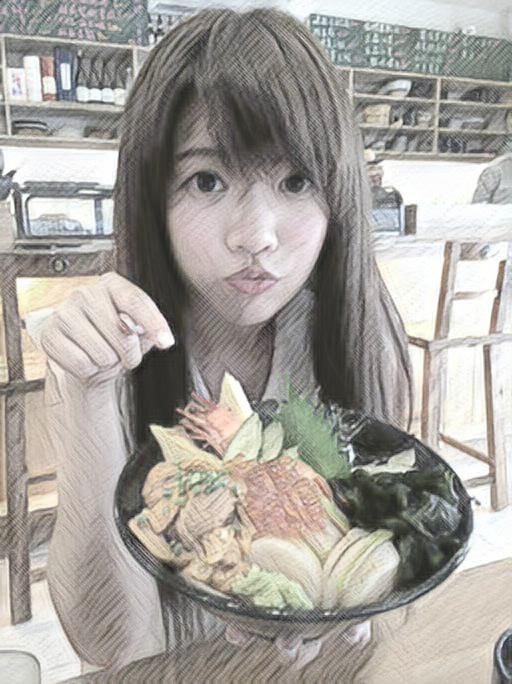}
}
\subfloat
{
\includegraphics[width=0.18\linewidth,height=0.2\linewidth]{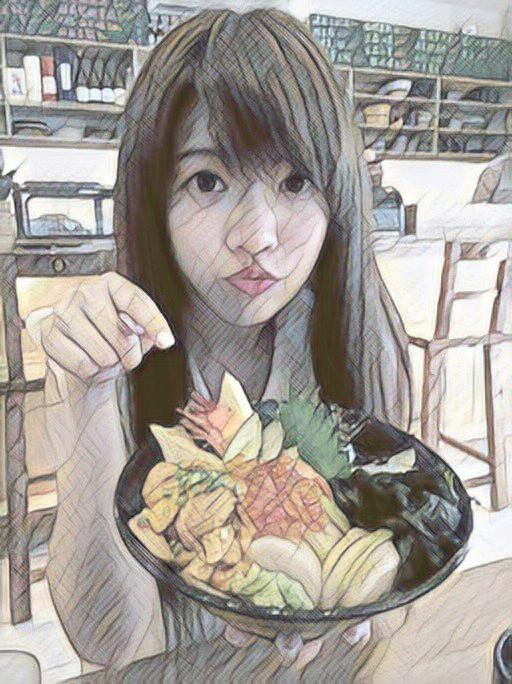}
}
\subfloat
{
\includegraphics[width=0.18\linewidth,height=0.2\linewidth]{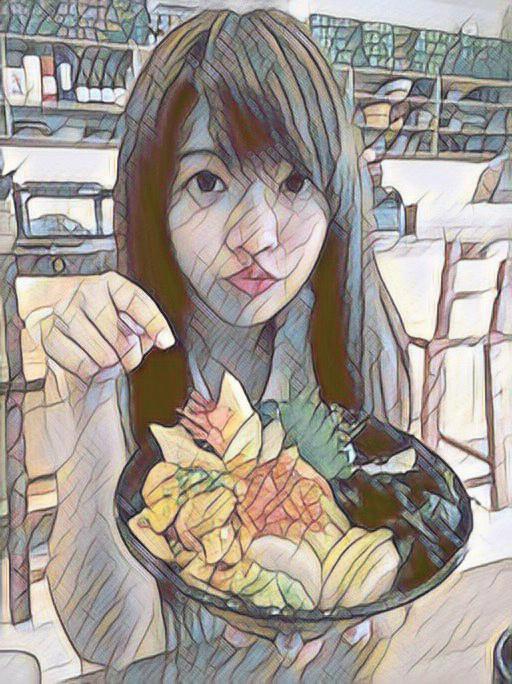}
}
\subfloat
{
\includegraphics[width=0.18\linewidth,height=0.2\linewidth]{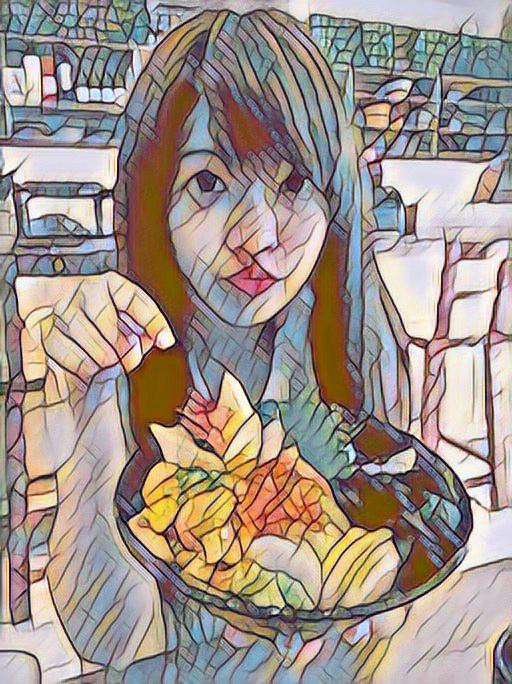}
}
\subfloat
{
\includegraphics[width=0.18\linewidth,height=0.2\linewidth]{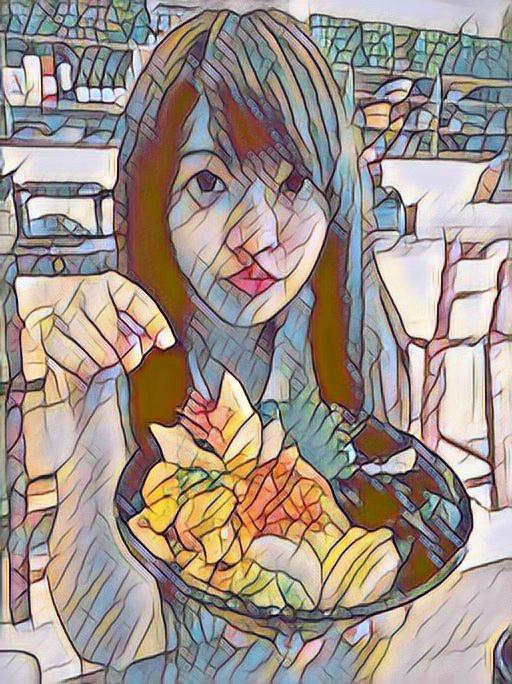}
}\par
\vspace{-0.8em}
%%%%%%%%%%%%%%%%%%%%%%%%%%%%%%%%%%%%%%%%%5
\subfloat
{
\includegraphics[width=0.18\linewidth,height=0.2\linewidth]{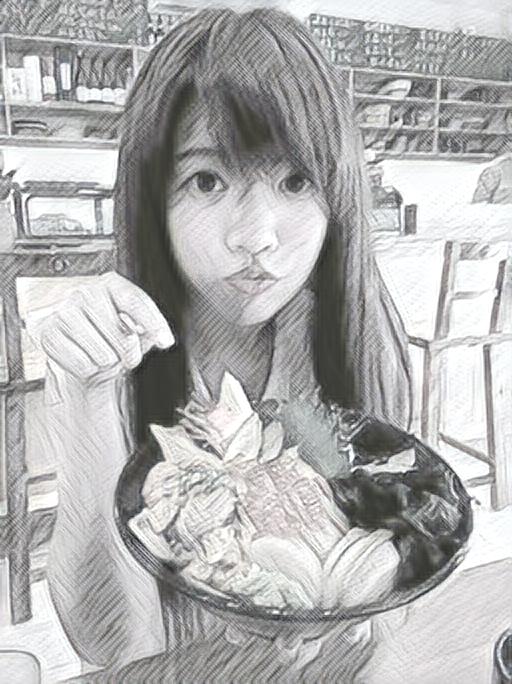}
}
\subfloat
{
\includegraphics[width=0.18\linewidth,height=0.2\linewidth]{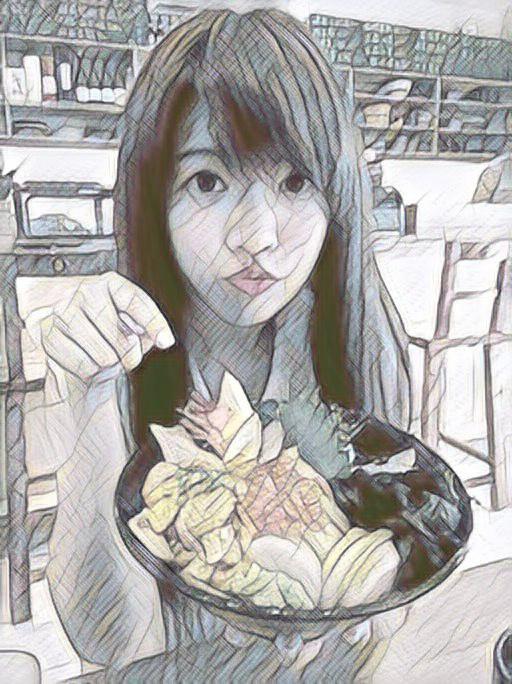}
}
\subfloat
{
\includegraphics[width=0.18\linewidth,height=0.2\linewidth]{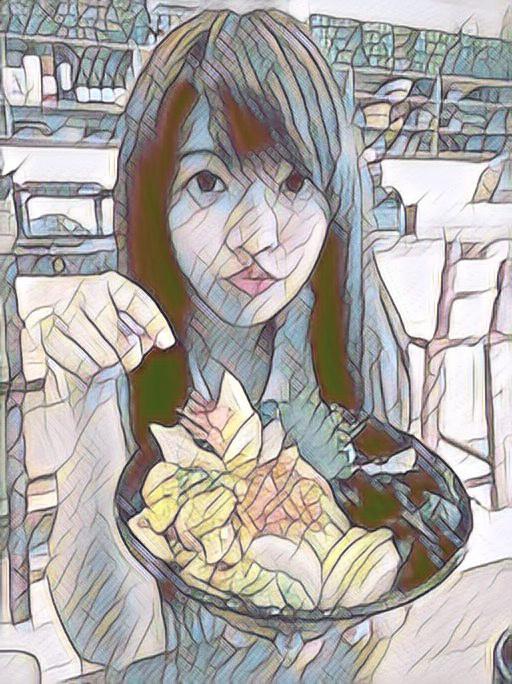}
}
\subfloat
{
\includegraphics[width=0.18\linewidth,height=0.2\linewidth]{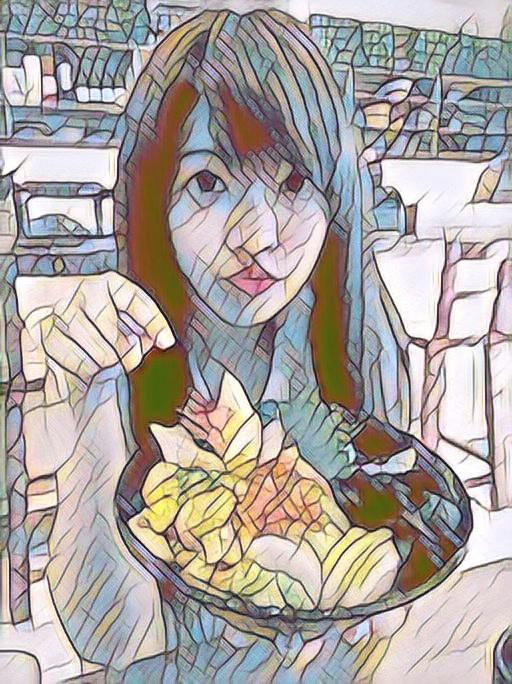}
}
\subfloat
{
\includegraphics[width=0.18\linewidth,height=0.2\linewidth]{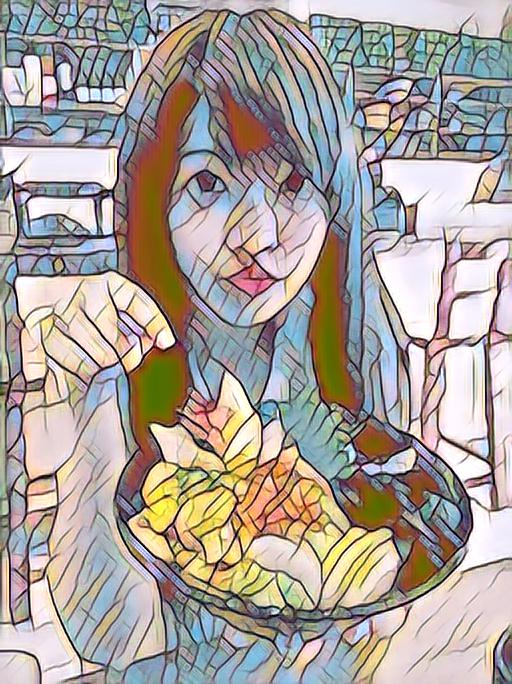}
}\par
%%%%%%%%%%%%%%%%%%%%%%%%%%%%%%%%%%%%%%%%%5
\caption{Content and style trade-off and interpolation. }
\label{fig:interp}
\vspace{-0.7em}
\end{figure}

% the table show the network architecture is here

\begin{figure}
\subfloat
{
  \begin{overpic}[width=0.48\linewidth,height=0.28\linewidth]{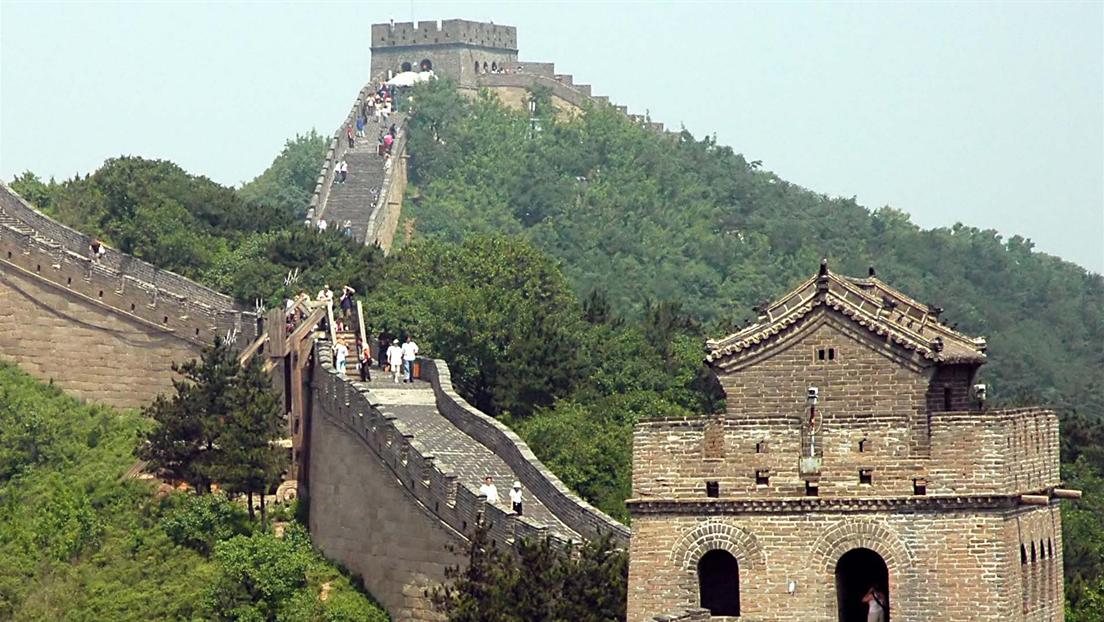}
     \put(74,33){\frame{\includegraphics[width=0.13\linewidth,height=0.13\linewidth]{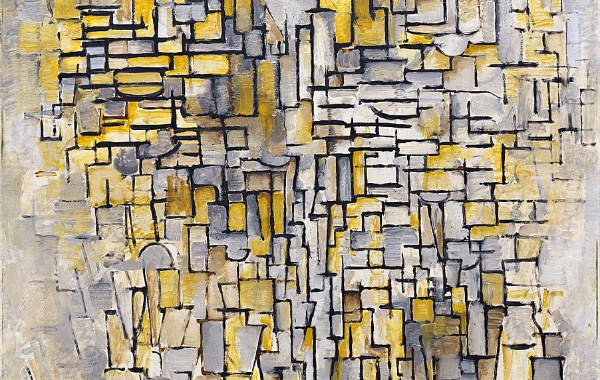}} }
  \end{overpic}
}
\subfloat
{
\includegraphics[width=0.48\linewidth,height=0.28\linewidth]{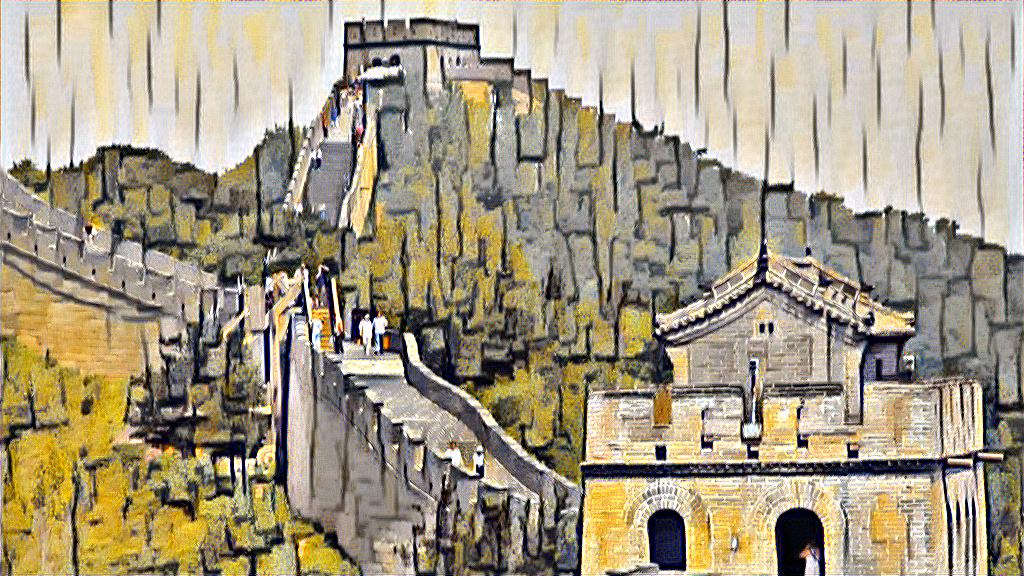}
}
\par
\vspace{-0.8em}
\subfloat
{
\includegraphics[width=0.48\linewidth,height=0.28\linewidth]{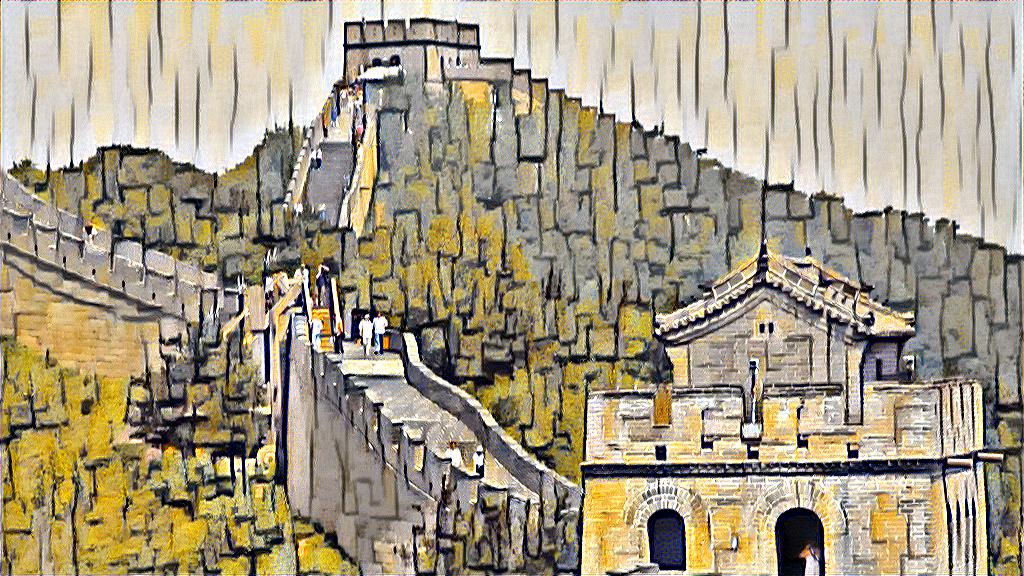}
}
\subfloat
{
\includegraphics[width=0.48\linewidth,height=0.28\linewidth]{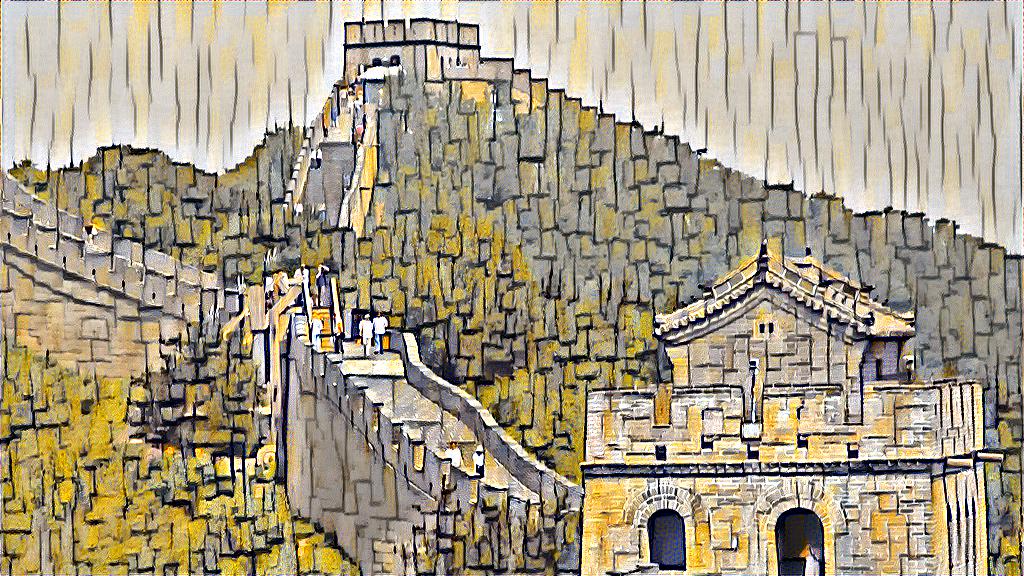}
}
\caption{Brush-size control using MSG-Net. Top left: High-resolution input image and dense style. Others: Style transfer results using MSG-Net with brush-size control. }
\label{fig:brush2}
\vspace{-0.7em}
\end{figure}

\paragraph{Brush Stroke Size Control. } Feeding the generative model with high-resolution image usually results in unsatisfying style transfer outputs, as shown in Figure~\ref{fig:brush} (c). Controlling brush stroke size can be achieved using optimization-based approach~\cite{gatys2016controlling}. Resizing the style image changes the brush-size, and feed-forward generative model matches the feature statistics at runtime should naturally achieve brush stoke size control. However, prior work is mainly limited by the 1D style embedding, because this finer style behavior cannot be captured using merely featuremap mean and variance. With MSG-Net, the CoMatch Layer matching the second order statistics elegantly solves the brush-size control. During training, we train the network with different style image sizes to learn from different brush stroke sizes. 
After training, the brush stroke size can be an option to the user by changing style input image size. Note that the MSG-Net can accept different input sizes for style and content images. 
Example results are shown in Figure~\ref{fig:brush2}.

\begin{figure*}
\captionsetup[subfloat]{labelformat=empty}
\centering
%%%%%%%%%%%%%%%%%%%%%%%%%%
%\begin{adjustbox}{width=\linewidth}
\subfloat
{
  \begin{overpic}[width=0.16\linewidth]{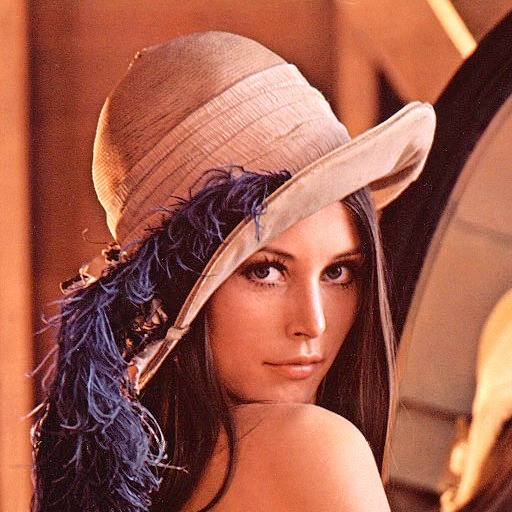}
     \put(58,58){\frame{\includegraphics[width=0.07\linewidth,height=0.07\linewidth]{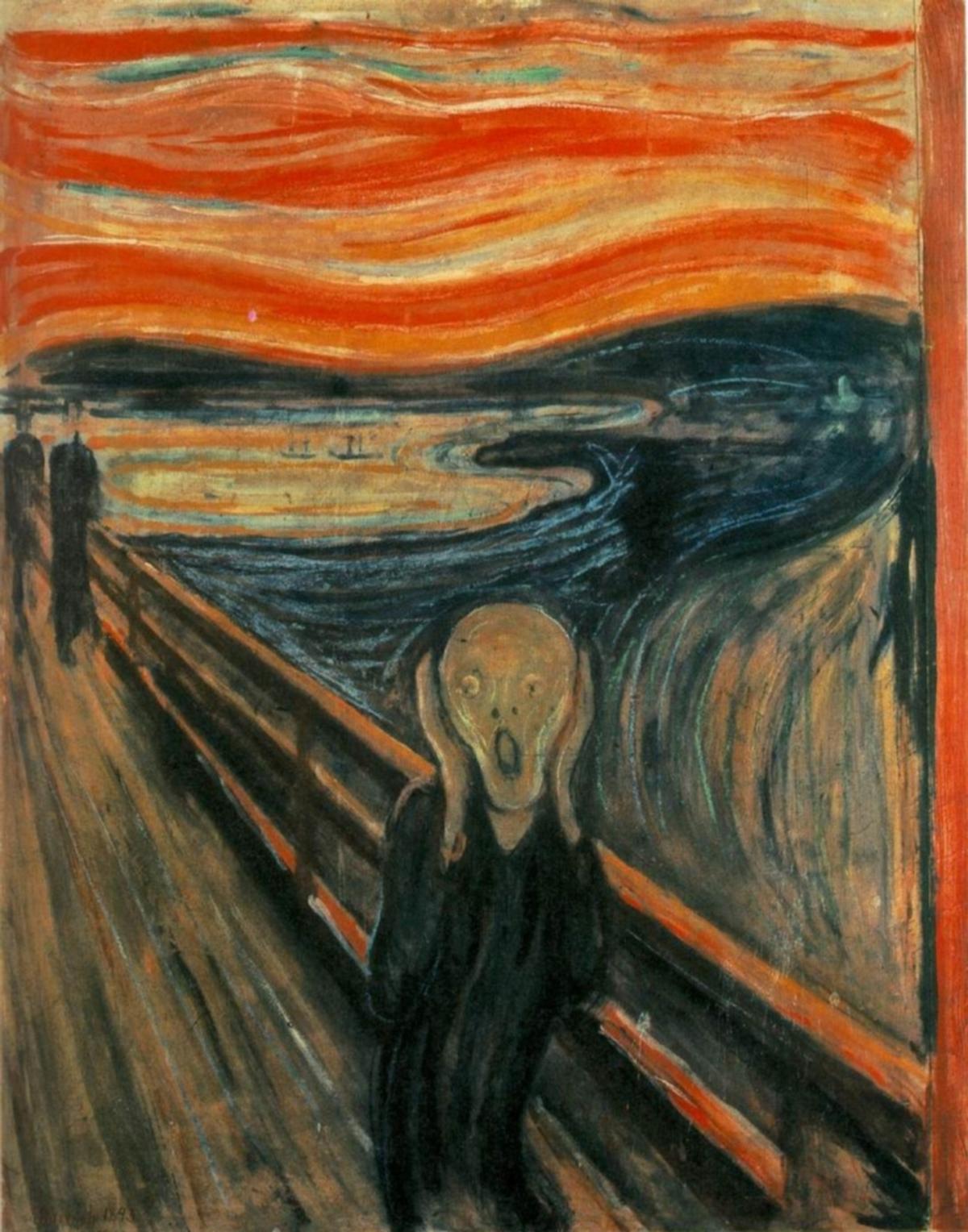}}}
  \end{overpic}
}
\subfloat
{
\includegraphics[width=0.16\linewidth]{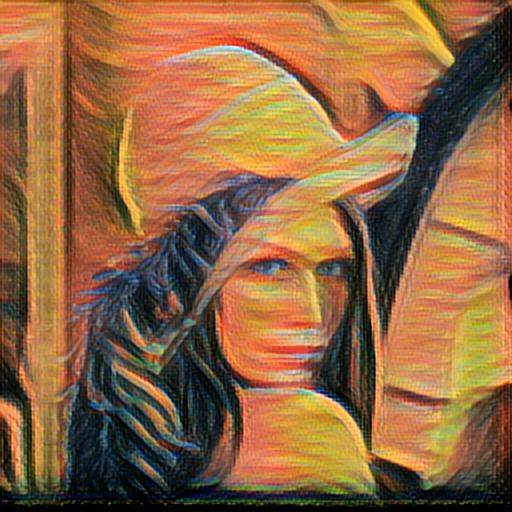}
}
\subfloat
{
\includegraphics[width=0.16\linewidth]{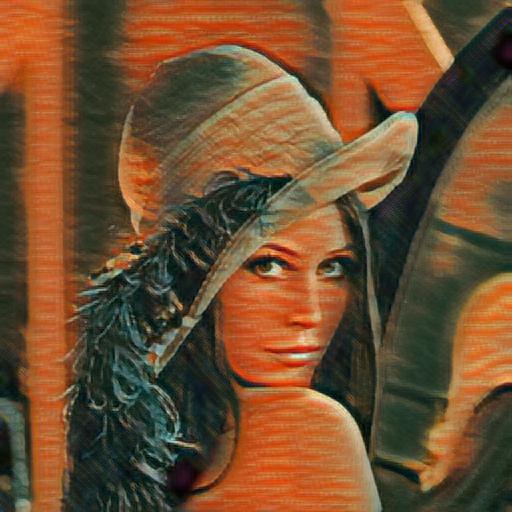}
}
\subfloat
{
\includegraphics[width=0.16\linewidth]{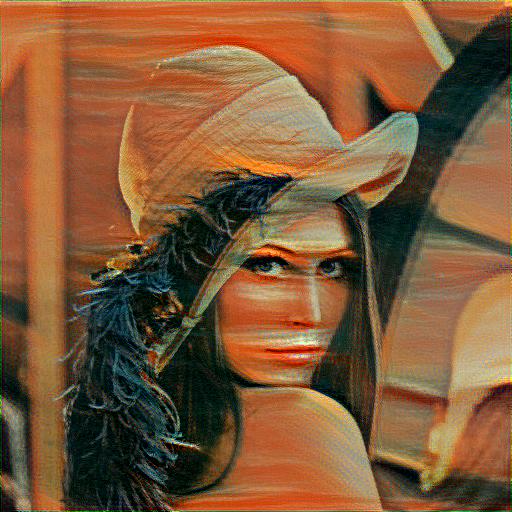}
}
\subfloat
{
\includegraphics[width=0.16\linewidth]{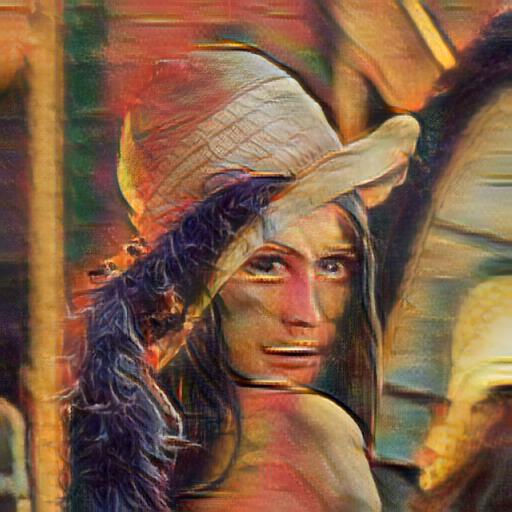}
}
\subfloat
{
\includegraphics[width=0.16\linewidth]{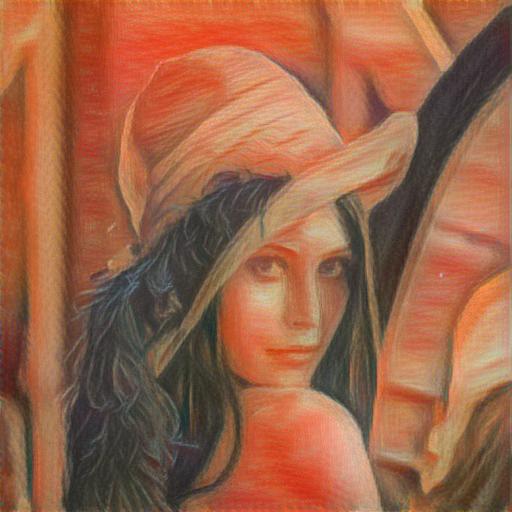}
}
\vspace{-0.8em}
%%%%%%%%%%%%%%%%%%%%%%%%%%
\par
\subfloat
{
  \begin{overpic}[width=0.16\linewidth,height=0.12\linewidth]{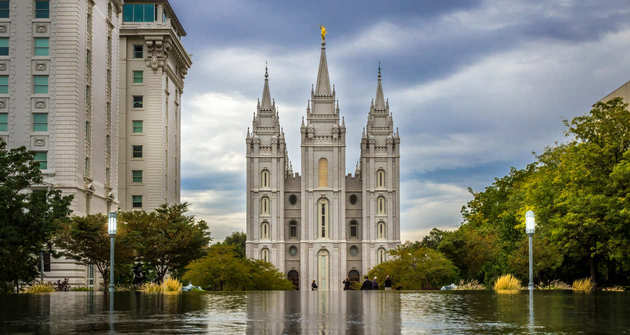}
     \put(58,32){\frame{\includegraphics[width=0.07\linewidth,height=0.07\linewidth]{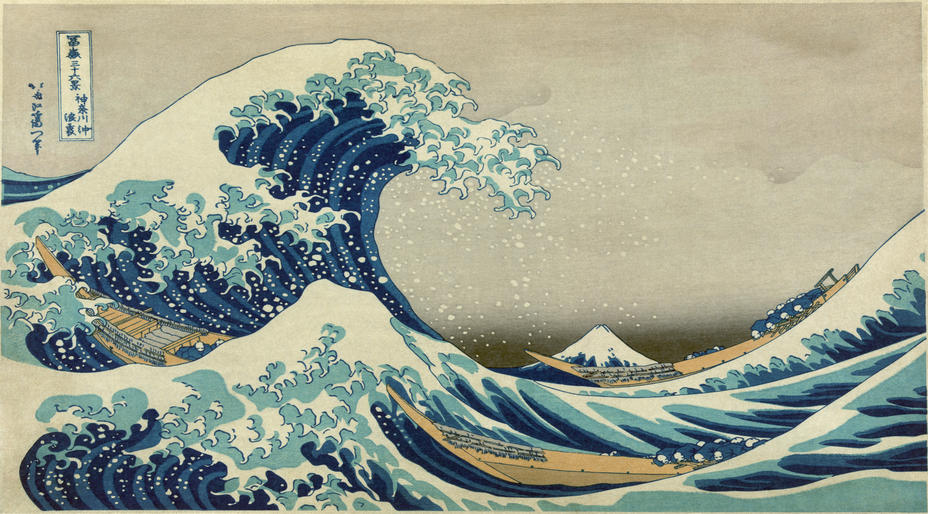}}}
  \end{overpic}
}
\subfloat
{
\includegraphics[width=0.16\linewidth,height=0.12\linewidth]{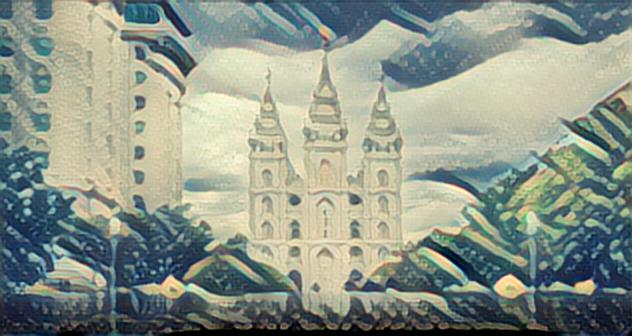}
}
\subfloat
{
\includegraphics[width=0.16\linewidth,height=0.12\linewidth]{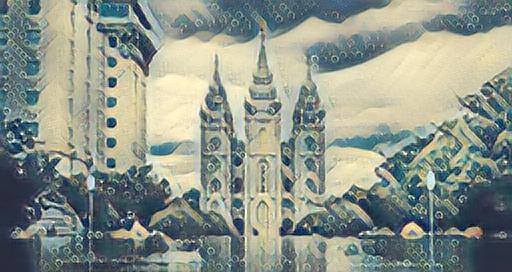}
}
\subfloat
{
\includegraphics[width=0.16\linewidth,height=0.12\linewidth]{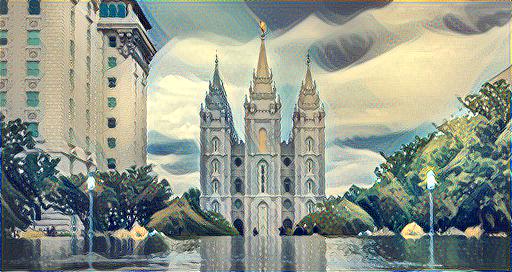}
}
\subfloat
{
\includegraphics[width=0.16\linewidth,height=0.12\linewidth]{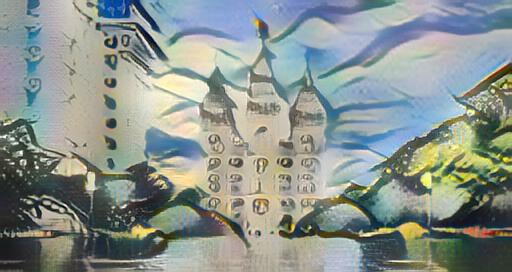}
}
\subfloat
{
\includegraphics[width=0.16\linewidth,height=0.12\linewidth]{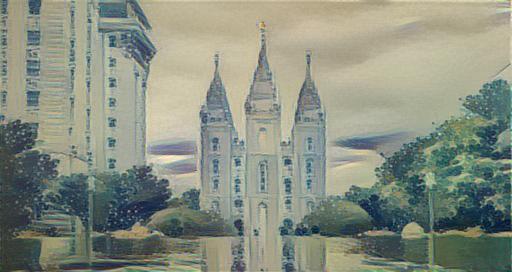}
}
\vspace{-0.8em}
%%%%%%%%%%%%%%%%%%%%%%%%%%
\par
\subfloat
{
  \begin{overpic}[width=0.16\linewidth,height=0.12\linewidth]{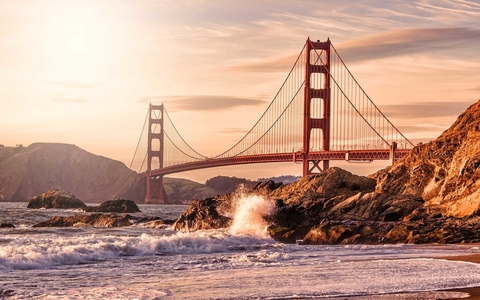}
     \put(58,32){\frame{\includegraphics[width=0.07\linewidth,height=0.07\linewidth]{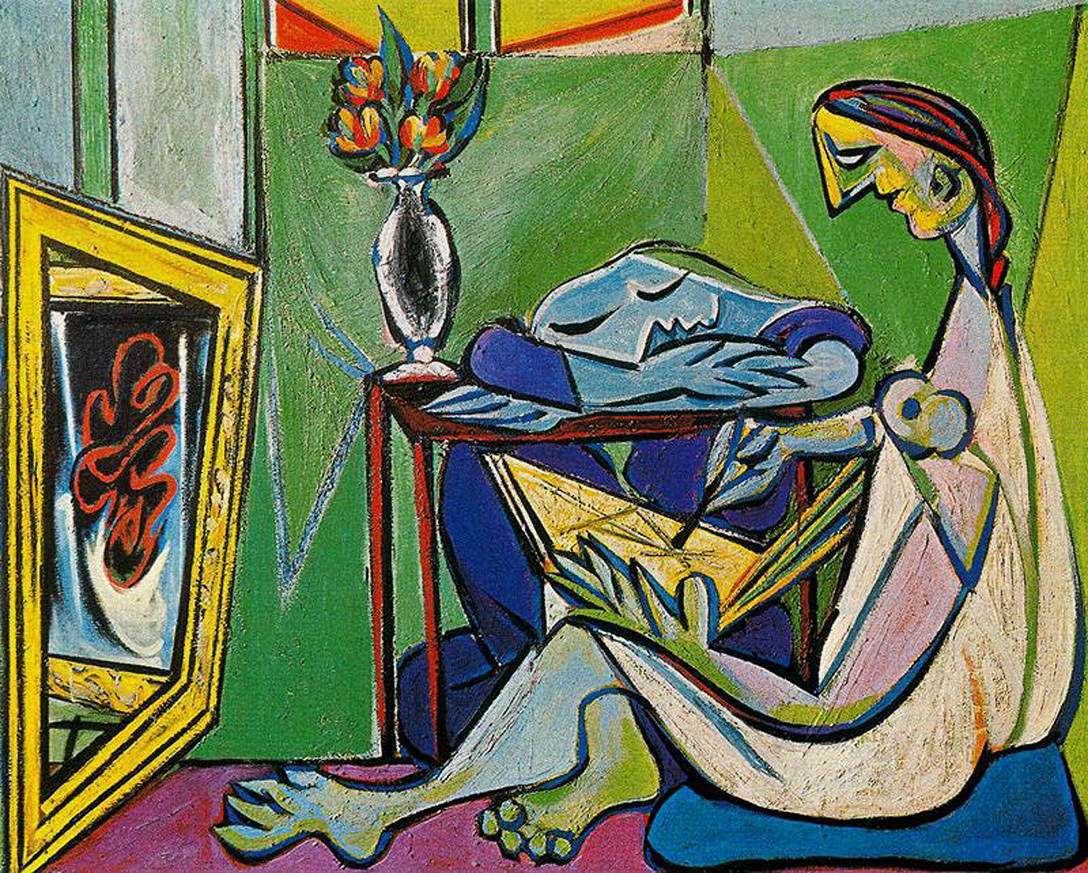}}}
  \end{overpic}
}
\subfloat
{
\includegraphics[width=0.16\linewidth,height=0.12\linewidth]{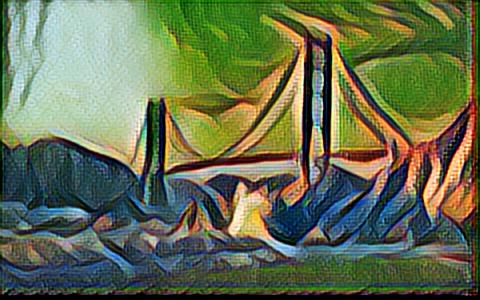}
}
\subfloat
{
\includegraphics[width=0.16\linewidth,height=0.12\linewidth]{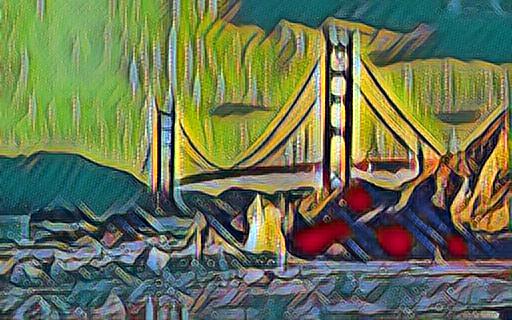}
}
\subfloat
{
\includegraphics[width=0.16\linewidth,height=0.12\linewidth]{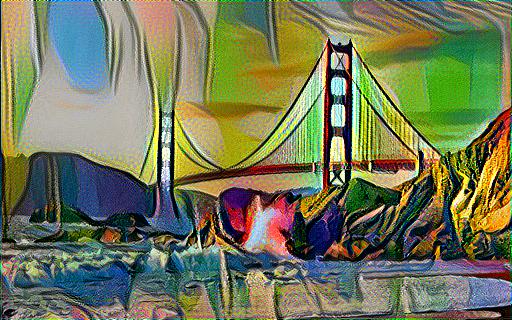}
}
\subfloat
{
\includegraphics[width=0.16\linewidth,height=0.12\linewidth]{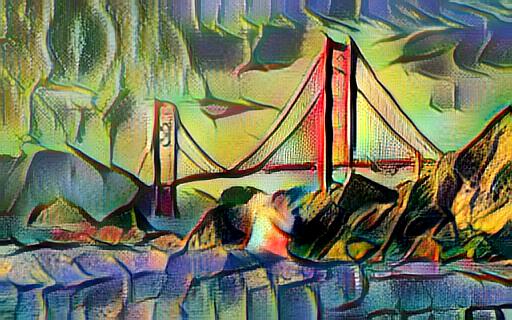}
}
\subfloat
{
\includegraphics[width=0.16\linewidth,height=0.12\linewidth]{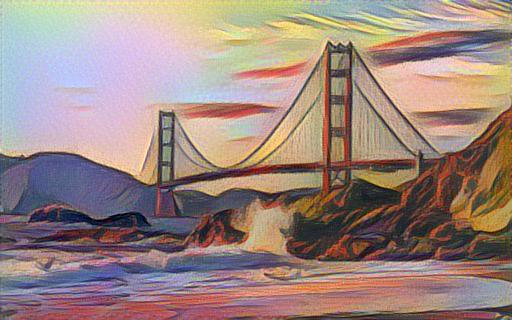}
}
\vspace{-0.8em}
%%%%%%%%%%%%%%%%%%%%%%%%%%
\par
\subfloat
{
  \begin{overpic}[width=0.16\linewidth,height=0.12\linewidth]{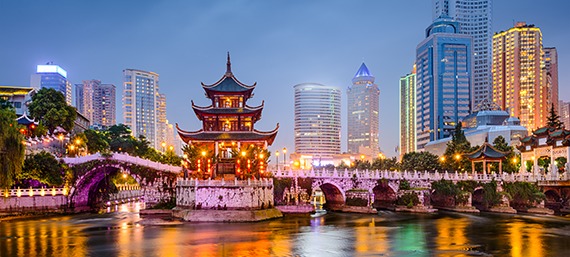}
     \put(58,32){\frame{\includegraphics[width=0.07\linewidth,height=0.07\linewidth]{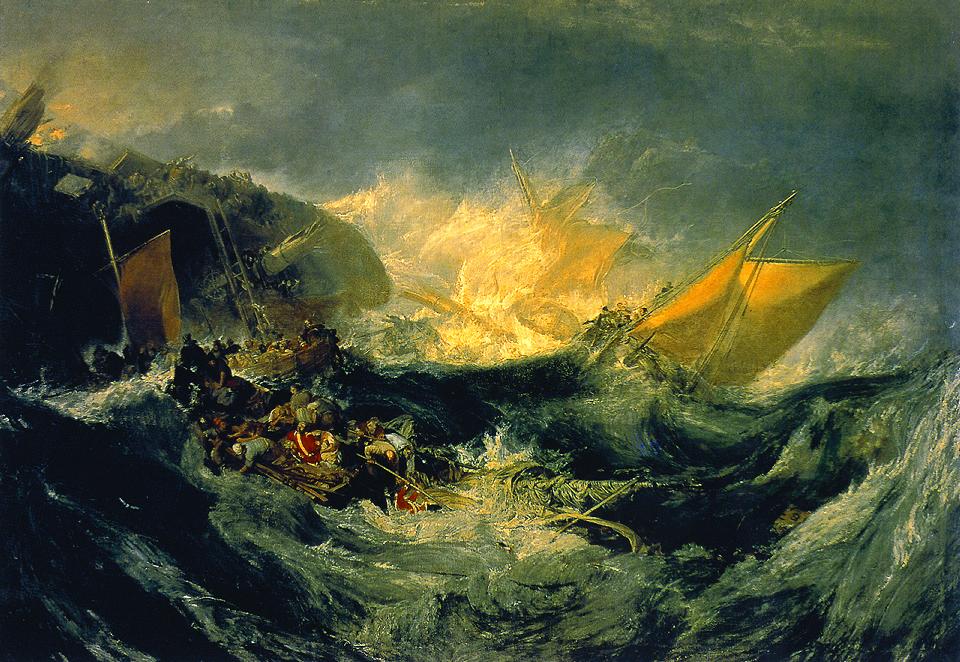}}}
  \end{overpic}
}
\subfloat
{
\includegraphics[width=0.16\linewidth,height=0.12\linewidth]{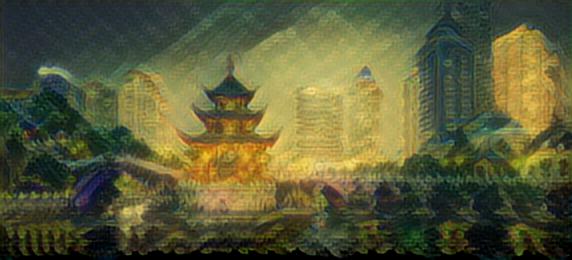}
}
\subfloat
{
\includegraphics[width=0.16\linewidth,height=0.12\linewidth]{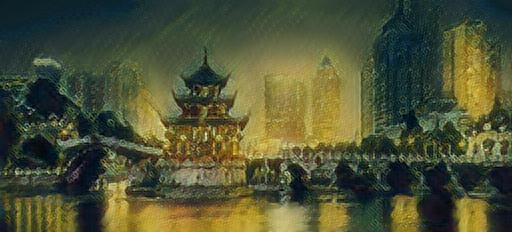}
}
\subfloat
{
\includegraphics[width=0.16\linewidth,height=0.12\linewidth]{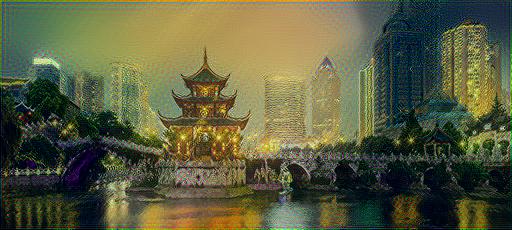}
}
\subfloat
{
\includegraphics[width=0.16\linewidth,height=0.12\linewidth]{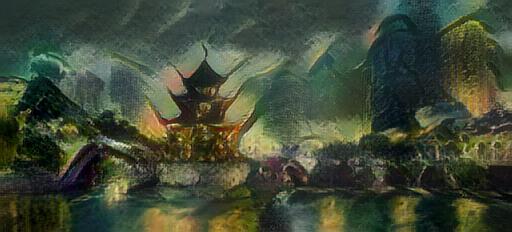}
}
\subfloat
{
\includegraphics[width=0.16\linewidth,height=0.12\linewidth]{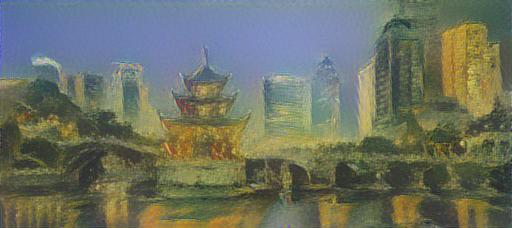}
}
\vspace{-1.8em}
%%%%%%%%%%%%%%%%%%%%%%%%%%
\setcounter{subfigure}{0}
\par
\subfloat[input]
{
  \begin{overpic}[width=0.16\linewidth,height=0.12\linewidth]{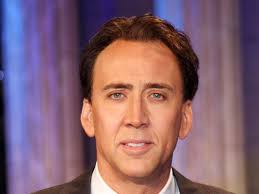}
     \put(58,32){\frame{\includegraphics[width=0.07\linewidth,height=0.07\linewidth]{figure/styles/rain_princess.jpg}}}
  \end{overpic}
}
\subfloat[Dumoulin {\it et al.}~\cite{dumoulin2016learned}]
{
\includegraphics[width=0.16\linewidth,height=0.12\linewidth]{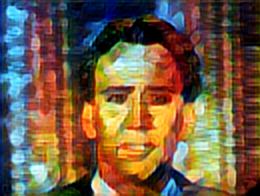}
}
\subfloat[MSG-Net (\TextRed{ours})]
{
\includegraphics[width=0.16\linewidth,height=0.12\linewidth]{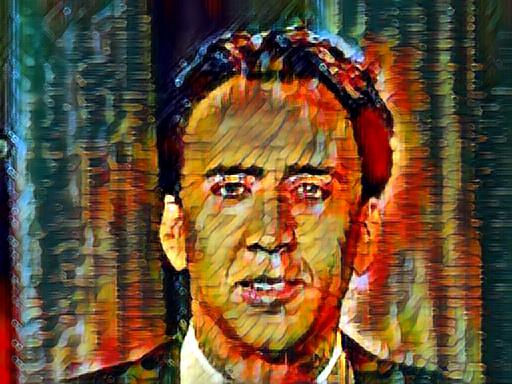}
}
\subfloat[Gatys {\it et al.}~\cite{gatys2016image}]
{
\includegraphics[width=0.16\linewidth,height=0.12\linewidth]{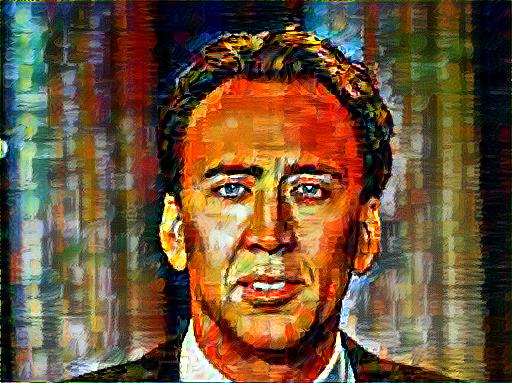}
}
\subfloat[Huang {\it et al.}~\cite{huang2017arbitrary}]
{
\includegraphics[width=0.16\linewidth,height=0.12\linewidth]{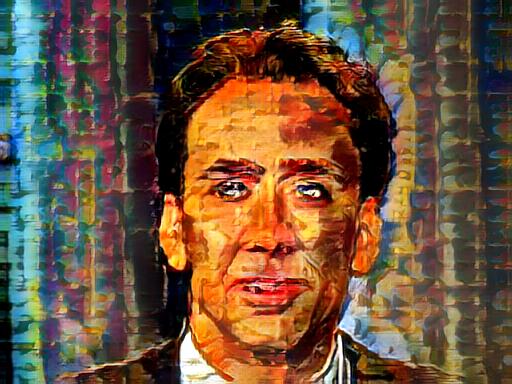}
}
\subfloat[Chen \& Schmidt~\cite{chen2016fast}]
{
\includegraphics[width=0.16\linewidth,height=0.12\linewidth]{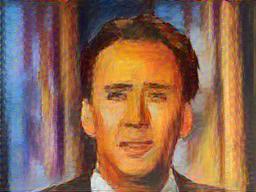}
}
%\end{adjustbox}
%%%%%%%%%%%%%%%%%%%%%%%%%%
\caption{
The tradeoff between style-flexibility and output-image quality is challenging  for generative models. Our approach enables multi-style transfer and has minimal difference in quality compared to the optimization-based Gatys approach~\cite{gatys2016image}. 
}
\vspace{-1em}
\label{fig:compare3}
\end{figure*}

\paragraph{Other Details.} We only use in-network down-sample (convolutional) and up-sample (upsampled convolution) in the transformation network. % as in previous work~\cite{radford2015unsupervised,Johnson2016Perceptual}.
We use reflectance padding to avoid artifacts at the border. Instance normalization~\cite{ulyanov2017improved} and ReLU are used after weight layers (convolution, fractionally-strided convolution and the CoMatch Layer), which improves the generated image quality and is robust to the image contrast changes.

% the figure compare with baselines of style transfer 

\subsection{Network Learning}
\label{sec:learning}

%\paragraph{\bf Loss Functions.}
Style transfer is an open problem, since there is no gold-standard ground-truth to follow.  
We follow  previous work to minimize a weighted combination of the style and content differences of the generator network outputs and the targets for a given pre-trained loss network $\mathcal{F}$~\cite{ulyanov2016texture,Johnson2016Perceptual}. Let the generator network be denoted by $G(x_c,x_s)$ parameterized by weights $W_G$. 
Learning proceeds by sampling content images $x_c\sim X_c$ and style images $x_s\sim X_s$ and then adjusting the parameters $W_G$ of the generator $G(x_c,x_s)$ in order to minimize the loss:
\begin{equation}
\begin{split}
\hat{W}_G&=\argmin_{W_G}E_{x_c,x_s}\{\\
&\lambda_c\|\mathcal{F}^c\left(G(x_c,x_s)\right)-\mathcal{F}^c(x_c)\|_F^2\\
&+\lambda_s\sum_{i=1}^K\|\mathcal{G}\left(\mathcal{F}^i(G(x_c,x_s))\right)-\mathcal{G}(\mathcal{F}^i(x_s))\|^2_F\\
&+\lambda_{TV}\ell_{TV}\left(G(x_c,x_s)\right) \} ,
\end{split}
\label{eq:loss}
\end{equation}
where $\lambda_c$ and $\lambda_s$ are the balancing weights for content and style losses. We consider image content at scale $c$ and image style at scales $i\in\{1,..K\}$. $\ell_{TV}()$ is the total variation regularization as used prior work for encouraging the smoothness of the generated images~\cite{Johnson2016Perceptual,mahendran2015understanding,zhang2010non}.

%section 5
\section{Experimental Results}
\label{sec:5}

%In this section, we make qualitative comparison of the proposed MSG-Net with existing approaches for style transfer task. We consider the gold-standard optimization based work of Gatys {\it et al.}~\cite{gatys2016image} and the state-of-the-art feed-forward approach of Johnson {\it et al.}~\cite{Johnson2016Perceptual} with Instance Normalization~\cite{ulyanov2017improved}. Additionally, we show MSG-Net can be applied to texture synthesis task. 

% texture synthesis figure
%\input{sections/textures}

\subsection{Style Transfer}
\paragraph{Baselines. }%We adopt a publicly available implementation~\cite{Johnson2015} of 
We use the implementation of the work of Gatys {\it et al.}~\cite{gatys2016image} as a gold standard baseline for style transfer approach (technical details will be included in the supplementary material). 
\begin{comment}
Given the content image $x_c$, style image $x_s$ and a pre-trained 16-layer VGG~\cite{simonyan2014very} as a descriptive network $\mathcal{F}$. Considering the content reconstruction at $c$-th scale and the style reconstruction at scales $i\in\{1,...K\}$, an image $\hat{y}$ is initialized with white noise and updated by iteratively minimizing the objective function: 
\begin{equation}
\begin{split}
\hat{y}&=\argmin_{y}\{\lambda_c\|\mathcal{F}^c(y)-\mathcal{F}^c(x_c)\|_F^2\\
&+\lambda_s\sum_{i=1}^K\|\mathcal{G}(\mathcal{F}^i(y))-\mathcal{G}(\mathcal{F}^i(x_s))\|^2_F\\
&+\lambda_{TV}\ell_{TV}(y) \} .
\end{split}
\end{equation}
The optimization is performed using L-BFGS solver for 500 iterations. %The method is slow due to forwarding and backwarding the image $y$ at each iteration.
\end{comment}
We also compare our approach with state-of-the-art multistyle or arbitrary style transfer methods, including patch-based approach~\cite{chen2016fast} and 1D style embedding~\cite{dumoulin2016learned,huang2017arbitrary}. The implementations from original authors are used in this experiments. 

%an improved version of recent feed-forward work~\cite{Johnson2016Perceptual} using instance normalization~\cite{ulyanov2017improved} as the state-of-the-art approach, where we train each feed-forward network individually for different style images using the loss function same as Equation~\ref{eq:loss}. 

\paragraph{Method Details. }
We adapt 16-layer VGG network~\cite{simonyan2014very} pre-trained on ImageNet %in the descriptive network as part of the MSG-Net and 
as the loss network in Equation~\ref{eq:loss}, because the network features learned from a diverse set of images are likely to be generic and informative. 
We consider the style representation at 4 different scales using the layers ReLU1\_2, ReLU2\_2, ReLU3\_3 and ReLU4\_3, and use the content representation at the layer ReLU2\_2.
The Microsoft COCO dataset~\cite{lin2014microsoft} is used as the content image image set $X_c$, which has around 80,000 natural images.
We collect 100 style images, choosing from  previous work in style transfer. 
Additionally 900 real paintings are selected from the open-source artistic dataset  wikiart.org~\cite{painter17} as additional style images for training MSG-Net-1K. 
%MSG-Net-100 is used in this experiment (as shown in Table~\ref{tab:msg-net}).
We follow the work~\cite{ulyanov2016texture,Johnson2016Perceptual} and adopt Adam~\cite{kingma2014adam} to train the network with a learning rate of $1\times 10^{-3}$. 
We use the loss function as described in Equation~\ref{eq:loss} with the balancing weights $\lambda_c=1,\lambda_s=5,\lambda_{TV}=1\times 10^{-6}$ for content, style and total regularization.
We resize the content images $x_c\sim X_c$ to $256\times 256$ and learn the network with a batch size of 4 for 80,000 iterations. 
We iteratively update the style image $x_s$ every iteration with size from \{256, 512, 768\} for runtime brush-size control. 
After training, the MSG-Net as a fully convolutional network~\cite{long2015fully} can accept arbitrary input image size.
For comparing the style transfer approaches, we use the same content image size, by resizing the image to 512 along the long side.
Our implementations are based on Torch~\cite{collobert2011torch7}, PyTorch~\cite{pytorch} and MXNet~\cite{chen2015mxnet}.   
It takes roughly 8 hours for training MSG-Net-100 model on a Titan Xp GPU.

\begin{table}[t]
\centering
\begin{adjustbox}{max width=0.48\textwidth}
\begin{tabular}{|l|c|c|c|c|c|c|c|c|c|}
 \hline
   & Model-size & Speed (256) & Speed (512) \\
 \hline
  Gatys {\it et al.}~\cite{gatys2016image} &  N/A & 0.07 & 0.02 \\
 \hline
 Johnson {et al.}~\cite{Johnson2016Perceptual} & 6.7MB & 91.7 & 26.3\\
 \hline
 Dumoulin {\it et al.}~\cite{dumoulin2016learned} & 6.8MB & 88.3 & 24.7\\
  \hline
 Chen {et al.}~\cite{chen2016fast} & 574MB & 5.84 & 0.31 \\
 \hline
 Huang {et al.}~\cite{huang2017arbitrary} & 28.1MB & 37.0 & 10.2\\
 \hline
 MSG-Net-100 \small{(\TextRed{ours})} & 9.6MB & 92.7 & 29.2 \\
 \hline
  MSG-Net-1K \small{(\TextRed{ours})} & 40.3MB & 47.2 &  14.3 \\
 \hline
\end{tabular}
\end{adjustbox}
\caption{
Comparing model size on disk and inference/test speed fps (frames/sec) of images with the size of 256$\times$256 and 512$\times$512 on a NVIDIA Titan Xp GPU average over 50 samples. MSG-Net-100 and MSG-Net-1K have 2.3M and 8.9M parameters respectively. 
}
\label{tab:compare}
\end{table}

\paragraph{Model Size and Speed Analysis}
For mobile applications or cloud services, the model size and test speed are crucial. We compare the model size and inference/test speed of style transfer approaches in Table~\ref{tab:compare}. 
Our proposed MSG-Net-100 has a comparable model size and speed with single style network~\cite{Johnson2016Perceptual,ulyanov2016texture}. 
The MSG-Net is faster than Arbitrary Style Transfer work~\cite{huang2017arbitrary}, because of using a learned compact encoder instead of pre-trained VGG network. 

\begin{figure*}
\begin{center}
\subfloat
{ 
\includegraphics[width=0.10\linewidth, height=0.10\linewidth]{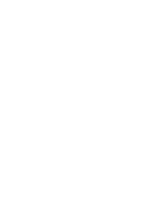}
}
\subfloat
{ 
\includegraphics[height=0.10\linewidth]{figure/content/venice-boat.jpg}
}
\subfloat
{ 
\includegraphics[height=0.10\linewidth]{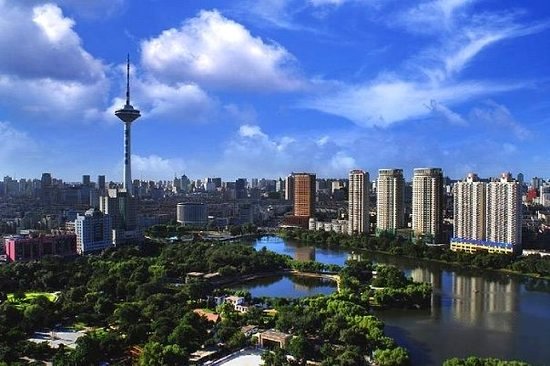}
}
\subfloat
{ 
\includegraphics[height=0.10\linewidth]{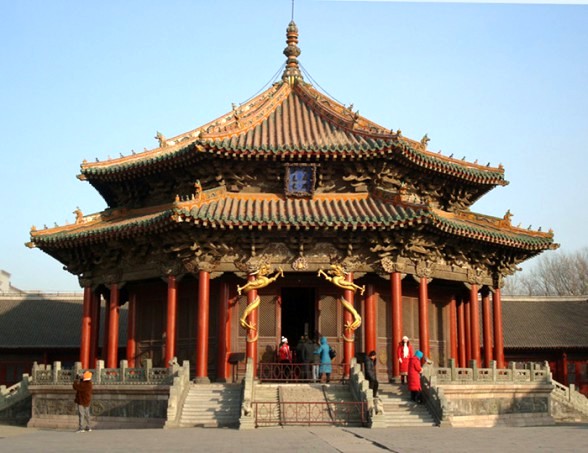}
}
\subfloat
{ 
\includegraphics[height=0.10\linewidth]{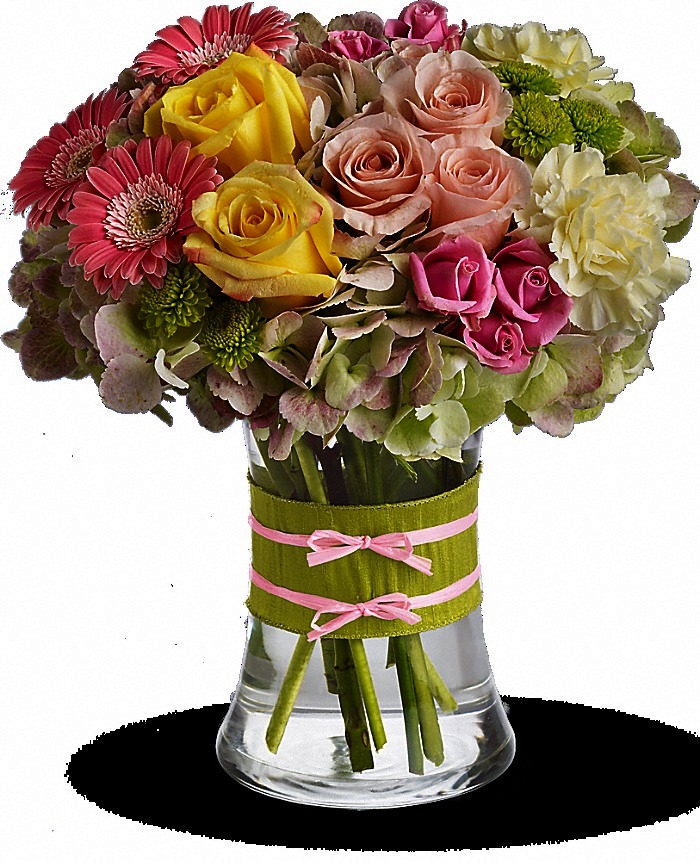}
}
\subfloat
{ 
\includegraphics[height=0.10\linewidth]{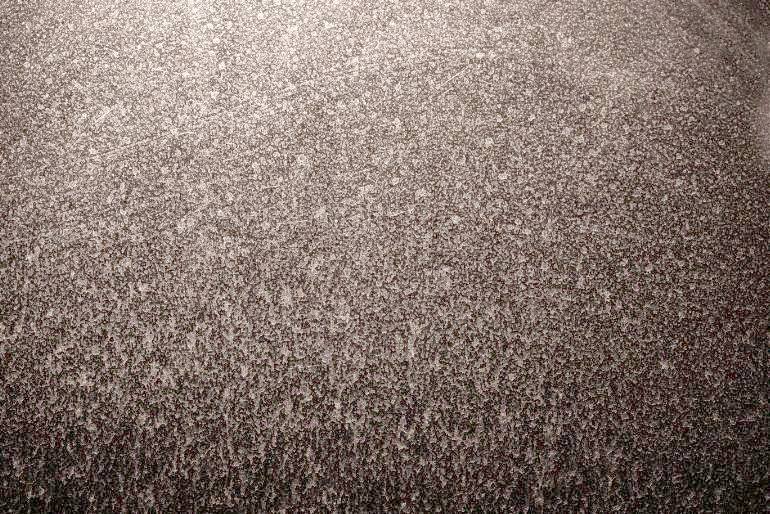}
}
\par
%%%%%%%%%%%%%%%%%%%%%%%%%%%%%%%%%%%%
\subfloat
{ 
\includegraphics[width=0.10\linewidth,height=0.10\linewidth]{figure/styles/candy.jpg}
}
\subfloat
{ 
\includegraphics[height=0.10\linewidth]{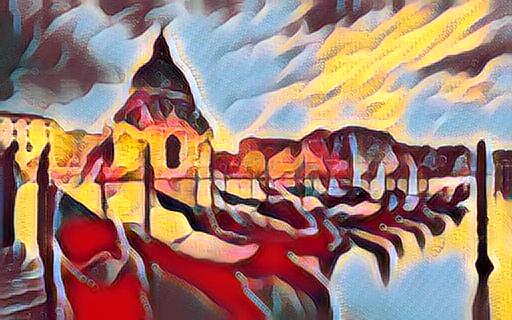}
}
\subfloat
{ 
\includegraphics[height=0.10\linewidth]{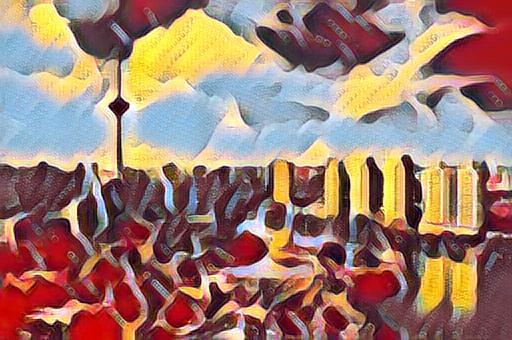}
}
\subfloat
{ 
\includegraphics[height=0.10\linewidth]{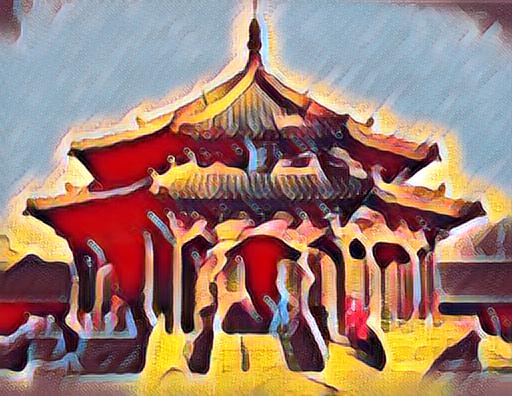}
}
\subfloat
{ 
\includegraphics[height=0.10\linewidth]{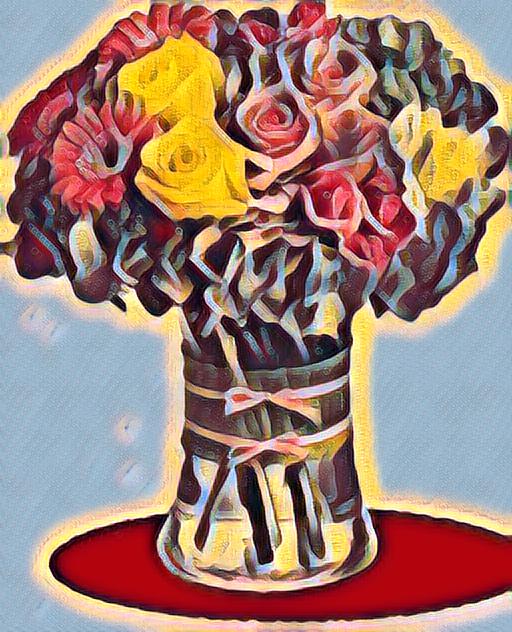}
}
\subfloat
{ 
\includegraphics[height=0.10\linewidth]{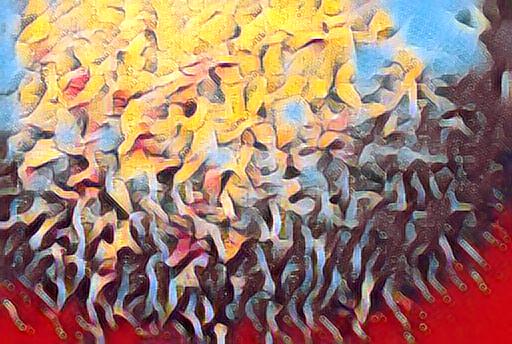}
}
\par
%%%%%%%%%%%%%%%%%%%%%%%%%%%%%%%%%%%%
\subfloat
{ 
\includegraphics[width=0.10\linewidth,height=0.10\linewidth]{figure/styles/composition_vii.jpg}
}
\subfloat
{ 
\includegraphics[height=0.10\linewidth]{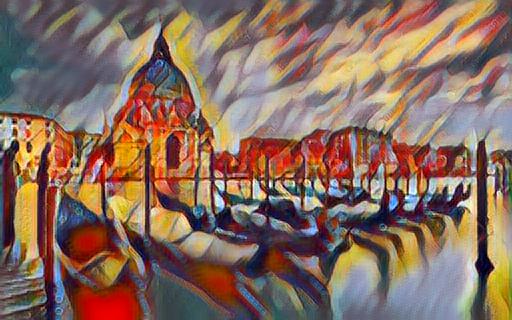}
}
\subfloat
{ 
\includegraphics[height=0.10\linewidth]{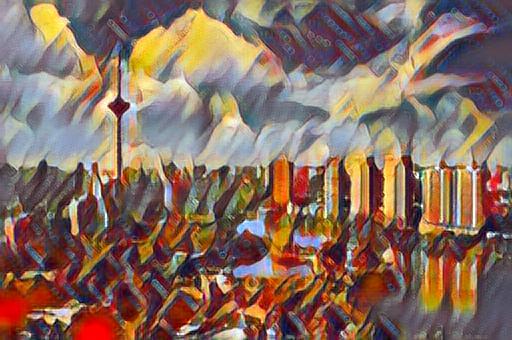}
}
\subfloat
{ 
\includegraphics[height=0.10\linewidth]{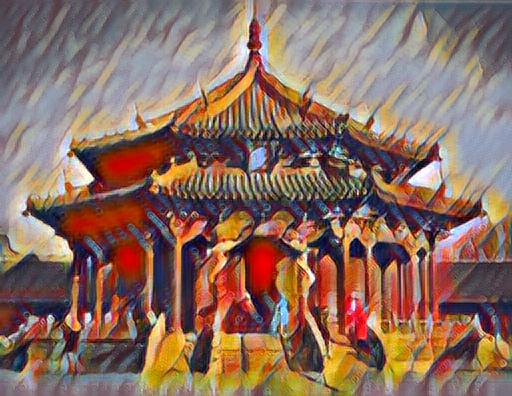}
}
\subfloat
{ 
\includegraphics[height=0.10\linewidth]{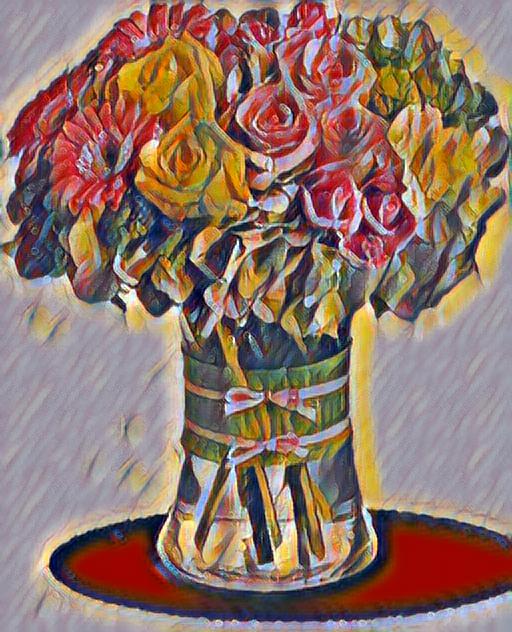}
}
\subfloat
{ 
\includegraphics[height=0.10\linewidth]{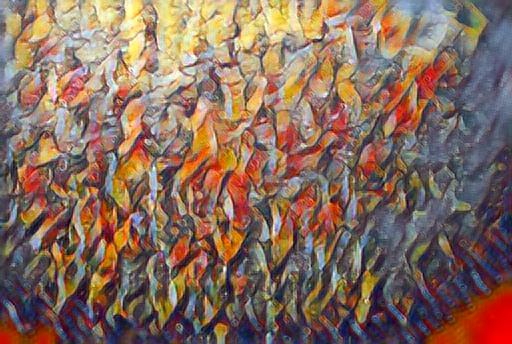}
}
\par
%%%%%%%%%%%%%%%%%%%%%%%%%%%%%%%%%%%%
% \subfloat
% { 
% \includegraphics[width=0.10\linewidth,height=0.10\linewidth]{figure/styles/feathers.jpg}
% }
% \subfloat
% { 
% \includegraphics[height=0.10\linewidth]{figure/venice_boat/feathers.jpg}
% }
% \subfloat
% { 
% \includegraphics[height=0.10\linewidth]{figure/shenyang/feathers.jpg}
% }
% \subfloat
% { 
% \includegraphics[height=0.10\linewidth]{figure/old_shenyang/feathers.jpg}
% }
% \subfloat
% { 
% \includegraphics[height=0.10\linewidth]{figure/flower/feathers.jpg}
% }
% \subfloat
% { 
% \includegraphics[height=0.10\linewidth]{figure/noise/feathers.jpg}
% }
% \par
%%%%%%%%%%%%%%%%%%%%%%%%%%%%%%%%%%%%
% \subfloat
% { 
% \includegraphics[width=0.10\linewidth,height=0.10\linewidth]{figure/styles/frida_kahlo.jpg}
% }
% \subfloat
% { 
% \includegraphics[height=0.10\linewidth]{figure/venice_boat/frida_kahlo.jpg}
% }
% \subfloat
% { 
% \includegraphics[height=0.10\linewidth]{figure/shenyang/frida_kahlo.jpg}
% }
% \subfloat
% { 
% \includegraphics[height=0.10\linewidth]{figure/old_shenyang/frida_kahlo.jpg}
% }
% \subfloat
% { 
% \includegraphics[height=0.10\linewidth]{figure/flower/frida_kahlo.jpg}
% }
% \subfloat
% { 
% \includegraphics[height=0.10\linewidth]{figure/noise/frida_kahlo.jpg}
% }
% \par
%%%%%%%%%%%%%%%%%%%%%%%%%%%%%%%%%%%%
\subfloat
{ 
\includegraphics[width=0.10\linewidth,height=0.10\linewidth]{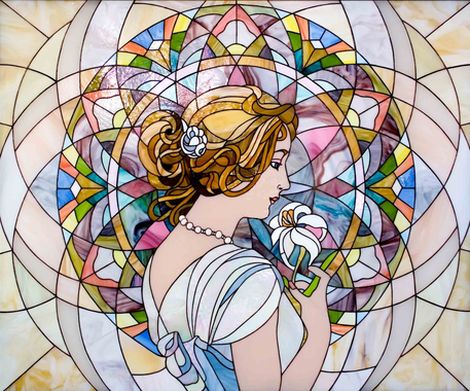}
}
\subfloat
{ 
\includegraphics[height=0.10\linewidth]{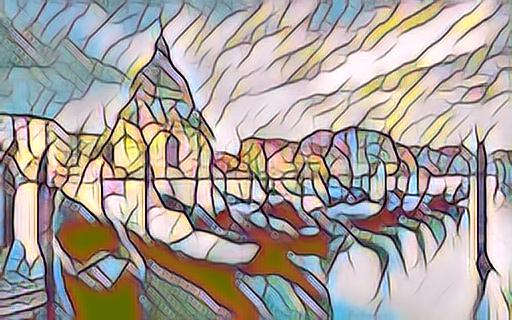}
}
\subfloat
{ 
\includegraphics[height=0.10\linewidth]{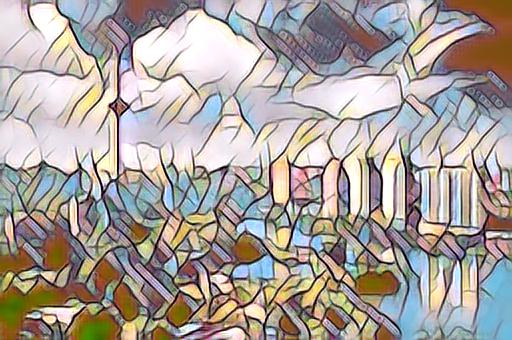}
}
\subfloat
{ 
\includegraphics[height=0.10\linewidth]{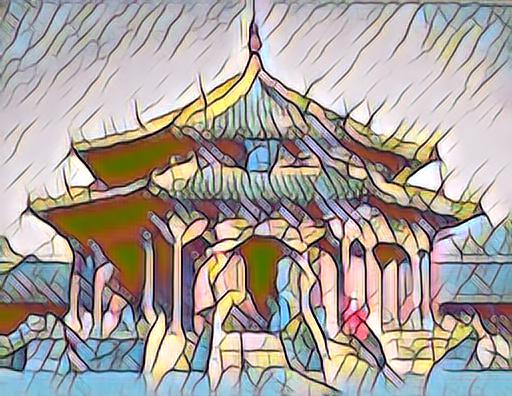}
}
\subfloat
{ 
\includegraphics[height=0.10\linewidth]{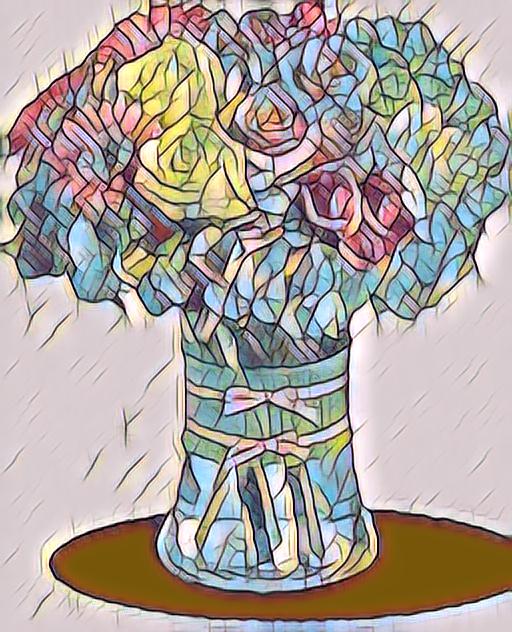}
}
\subfloat
{ 
\includegraphics[height=0.10\linewidth]{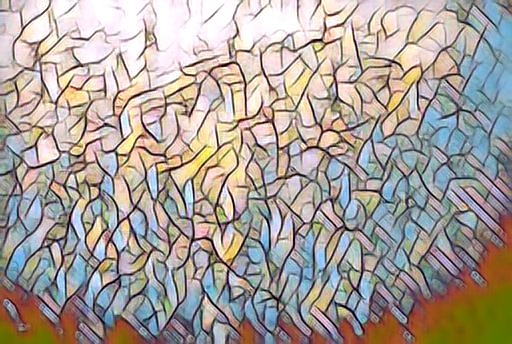}
}
\par
%%%%%%%%%%%%%%%%%%%%%%%%%%%%%%%%%%%%
\subfloat
{ 
\includegraphics[width=0.10\linewidth,height=0.10\linewidth]{figure/styles/mosaic_ducks_massimo.jpg}
}
\subfloat
{ 
\includegraphics[height=0.10\linewidth]{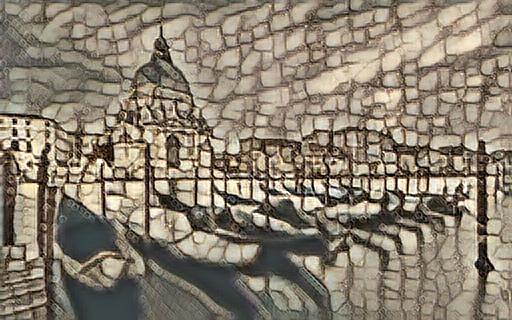}
}
\subfloat
{ 
\includegraphics[height=0.10\linewidth]{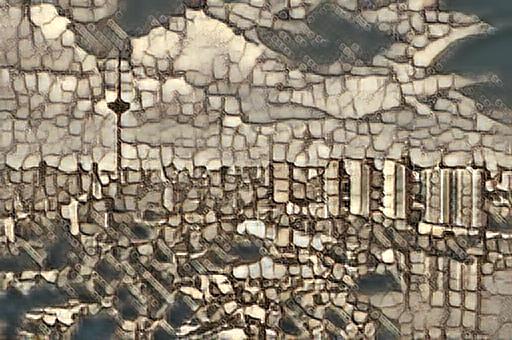}
}
\subfloat
{ 
\includegraphics[height=0.10\linewidth]{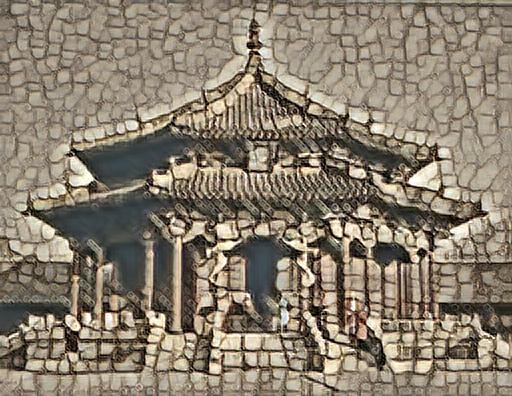}
}
\subfloat
{ 
\includegraphics[height=0.10\linewidth]{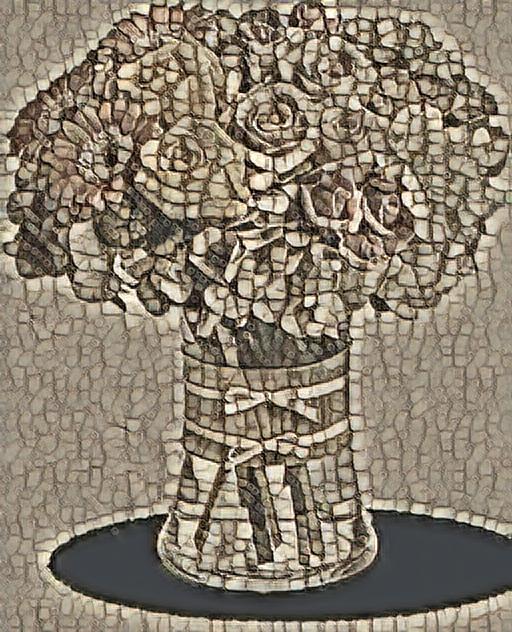}
}
\subfloat
{ 
\includegraphics[height=0.10\linewidth]{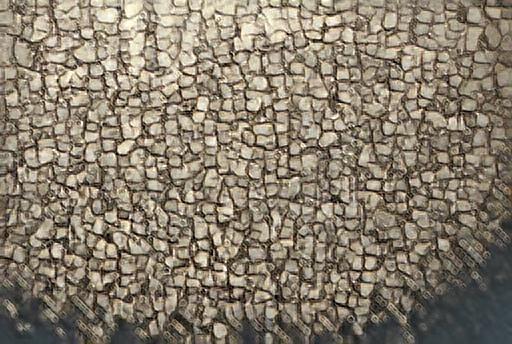}
}
\par
%%%%%%%%%%%%%%%%%%%%%%%%%%%%%%%%%%%%
\subfloat
{ 
\includegraphics[width=0.10\linewidth,height=0.10\linewidth]{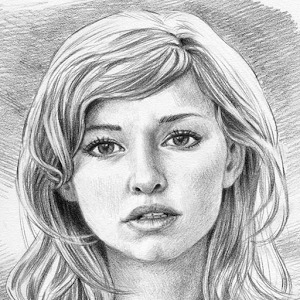}
}
\subfloat
{ 
\includegraphics[height=0.10\linewidth]{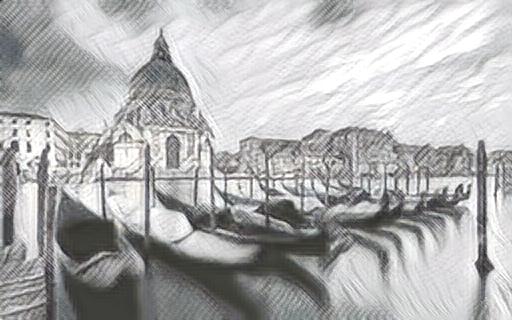}
}
\subfloat
{ 
\includegraphics[height=0.10\linewidth]{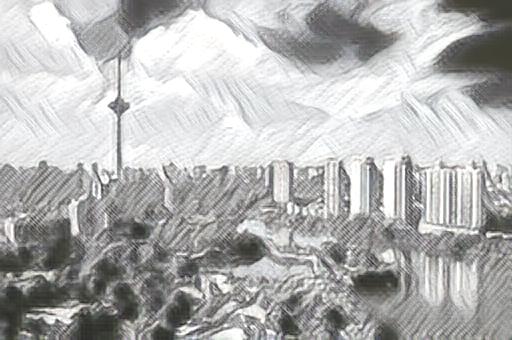}
}
\subfloat
{ 
\includegraphics[height=0.10\linewidth]{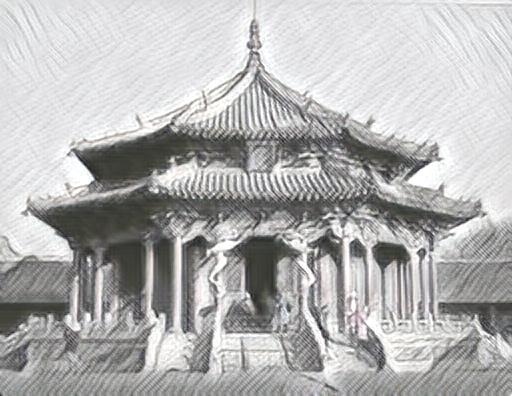}
}
\subfloat
{ 
\includegraphics[height=0.10\linewidth]{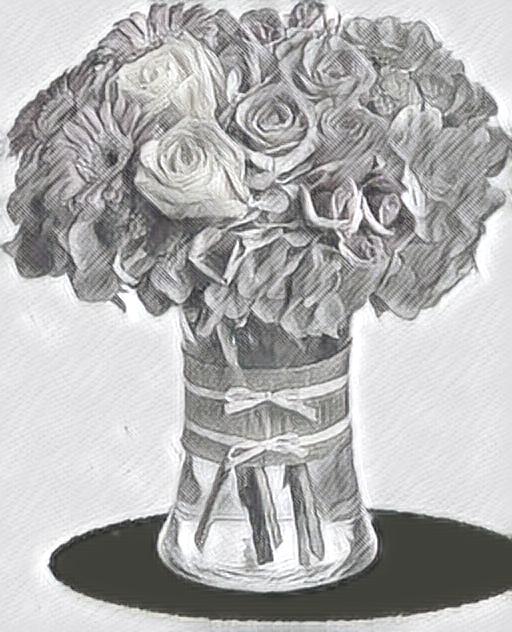}
}
\subfloat
{ 
\includegraphics[height=0.10\linewidth]{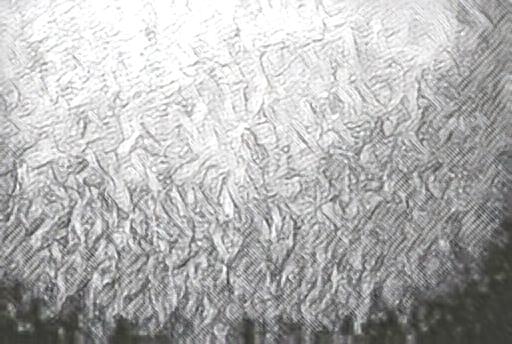}
}
\par
%%%%%%%%%%%%%%%%%%%%%%%%%%%%%%%%%%%%
\subfloat
{ 
\includegraphics[width=0.10\linewidth,height=0.10\linewidth]{figure/styles/picasso_selfport1907.jpg}
}
\subfloat
{ 
\includegraphics[height=0.10\linewidth]{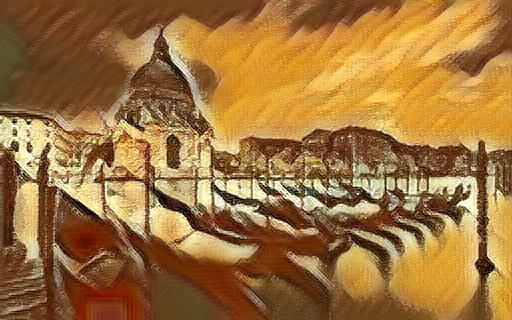}}
\subfloat
{ 
\includegraphics[height=0.10\linewidth]{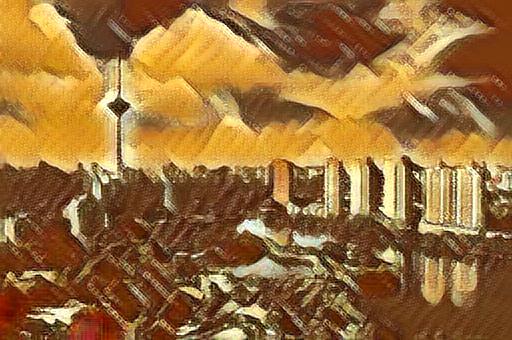}
}
\subfloat
{ 
\includegraphics[height=0.10\linewidth]{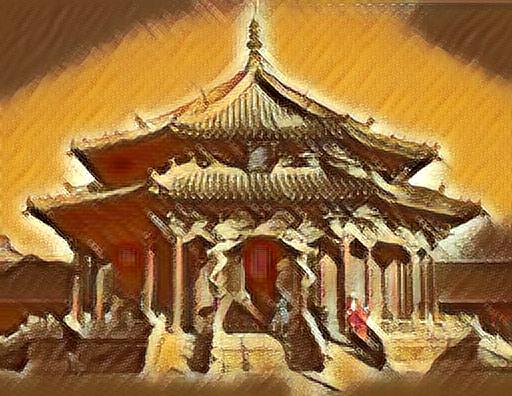}
}
\subfloat
{ 
\includegraphics[height=0.10\linewidth]{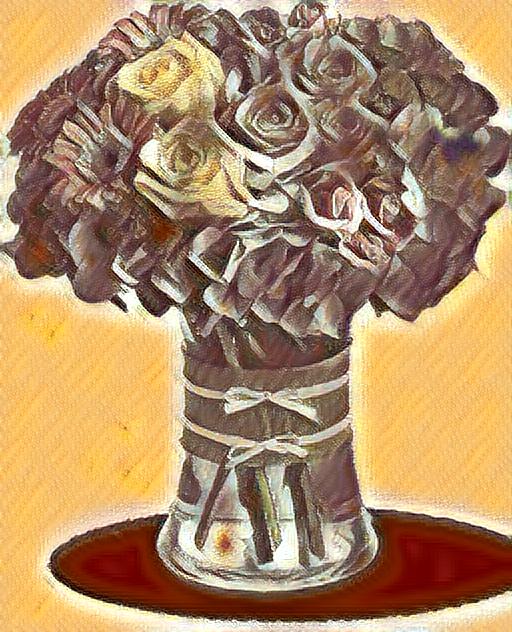}
}
\subfloat
{ 
\includegraphics[height=0.10\linewidth]{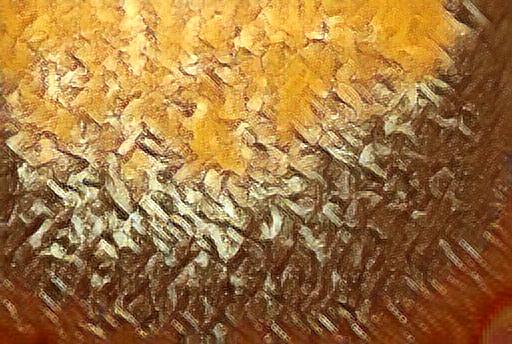}
}
\par
%%%%%%%%%%%%%%%%%%%%%%%%%%%%%%%%%%%%
\subfloat
{ 
\includegraphics[width=0.10\linewidth,height=0.10\linewidth]{figure/styles/Robert_Delaunay__1906__Portrait.jpg}
}
\subfloat
{ 
\includegraphics[height=0.10\linewidth]{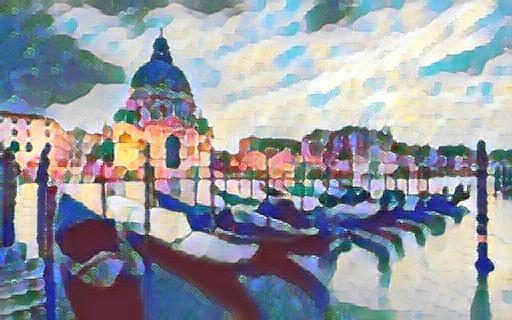}
}
\subfloat
{ 
\includegraphics[height=0.10\linewidth]{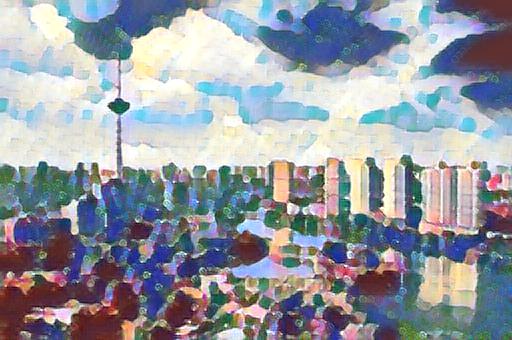}
}
\subfloat
{ 
\includegraphics[height=0.10\linewidth]{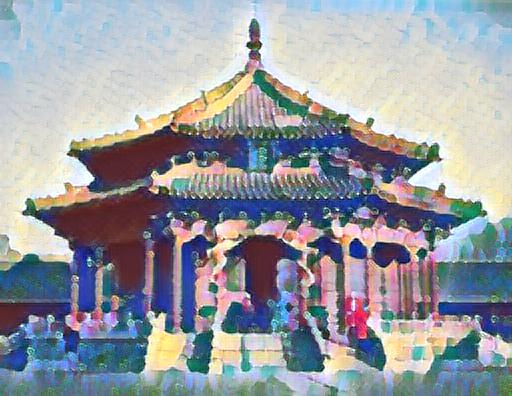}
}
\subfloat
{ 
\includegraphics[height=0.10\linewidth]{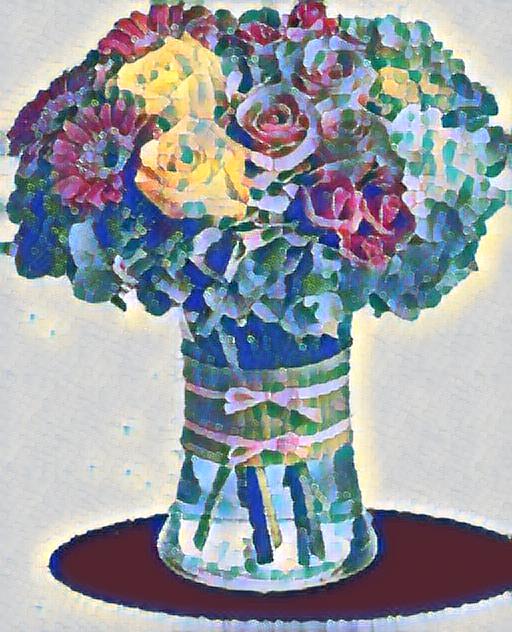}
}
\subfloat
{ 
\includegraphics[height=0.10\linewidth]{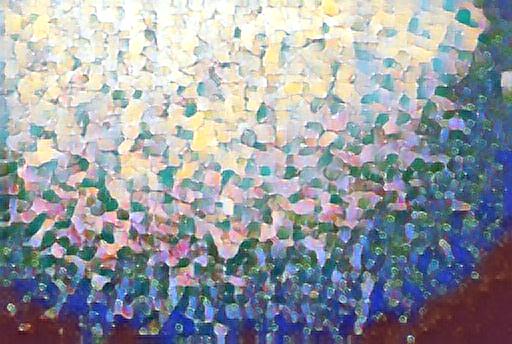}
}
\par
%%%%%%%%%%%%%%%%%%%%%%%%%%%%%%%%%%%%
% \subfloat
% { 
% \includegraphics[width=0.10\linewidth,height=0.10\linewidth]{figure/styles/seated-nude.jpg}
% }
% \subfloat
% { 
% \includegraphics[height=0.10\linewidth]{figure/venice_boat/seated-nude.jpg}
% }
% \subfloat
% { 
% \includegraphics[height=0.10\linewidth]{figure/shenyang/seated-nude.jpg}
% }
% \subfloat
% { 
% \includegraphics[height=0.10\linewidth]{figure/old_shenyang/seated-nude.jpg}
% }
% \subfloat
% { 
% \includegraphics[height=0.10\linewidth]{figure/flower/seated-nude.jpg}
% }
% \subfloat
% { 
% \includegraphics[height=0.10\linewidth]{figure/noise/seated-nude.jpg}
% }
% \par
%%%%%%%%%%%%%%%%%%%%%%%%%%%%%%%%%%%%
\subfloat
{ 
\includegraphics[width=0.10\linewidth,height=0.10\linewidth]{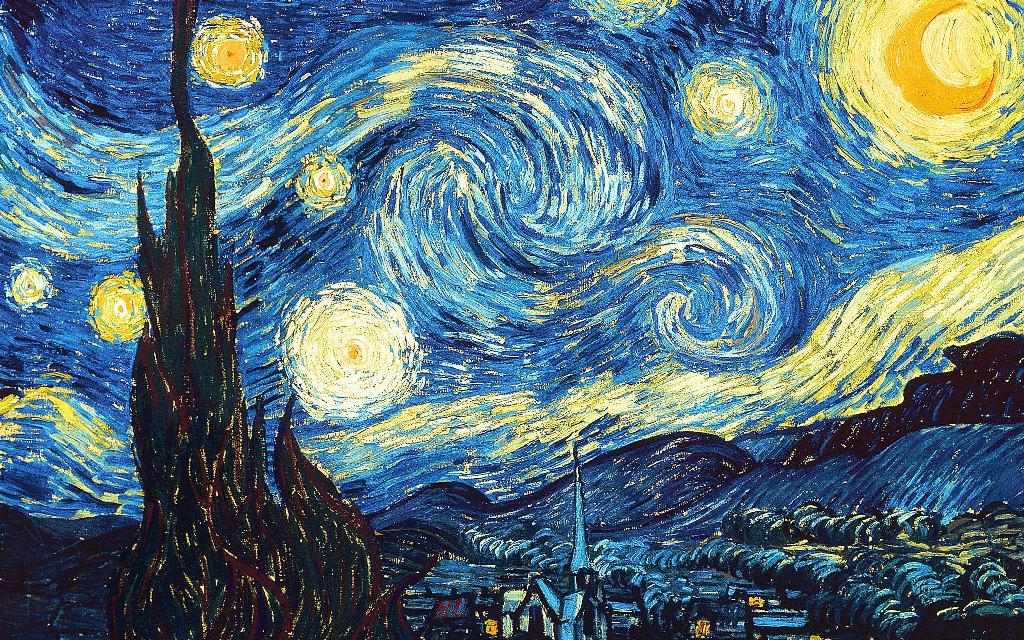}
}
\subfloat
{ 
\includegraphics[height=0.10\linewidth]{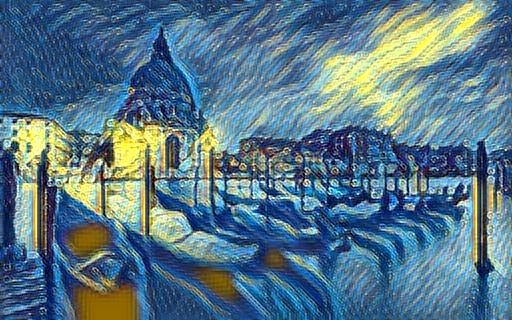}
}
\subfloat
{ 
\includegraphics[height=0.10\linewidth]{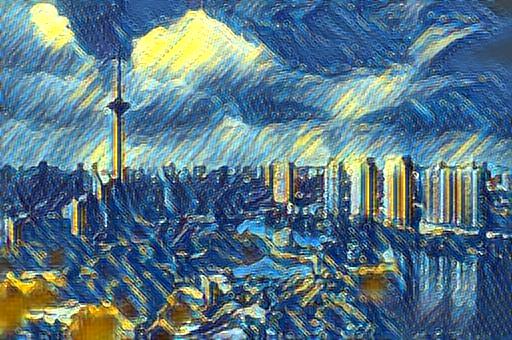}
}
\subfloat
{ 
\includegraphics[height=0.10\linewidth]{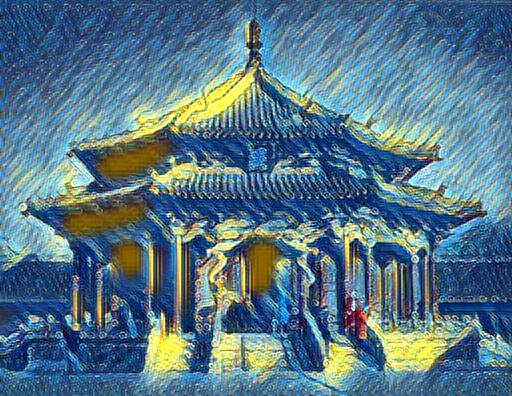}
}
\subfloat
{ 
\includegraphics[height=0.10\linewidth]{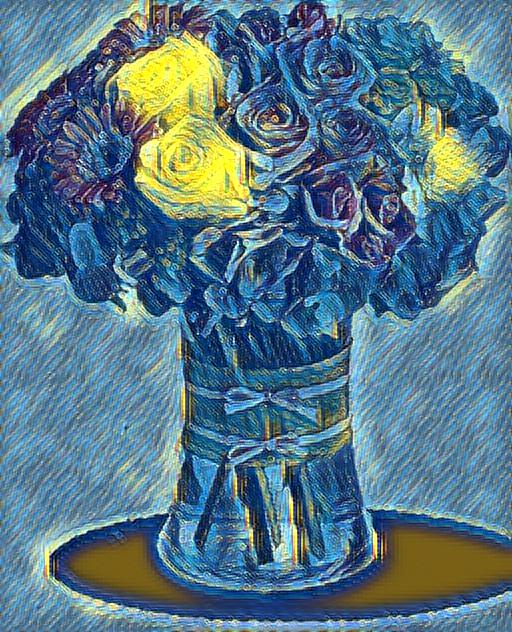}
}
\subfloat
{ 
\includegraphics[height=0.10\linewidth]{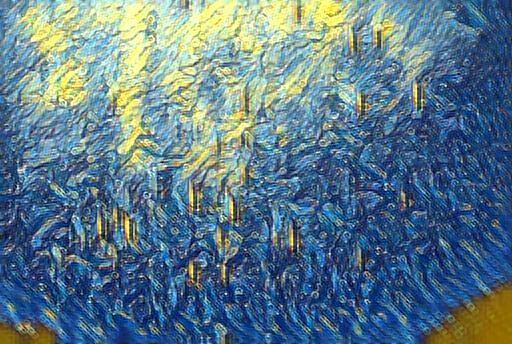}
}
%%%%%%%%%%%%%%%%%%%%%%%%%%%%%%%%%%%%
\par
\subfloat
{ 
\includegraphics[width=0.10\linewidth,height=0.10\linewidth]{figure/styles/woman-with-hat-matisse.jpg}
}
\subfloat
{ 
\includegraphics[height=0.10\linewidth]{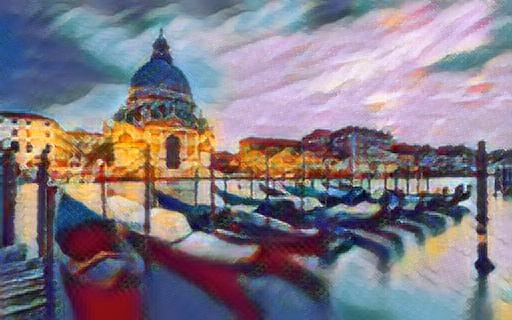}
}
\subfloat
{ 
\includegraphics[height=0.10\linewidth]{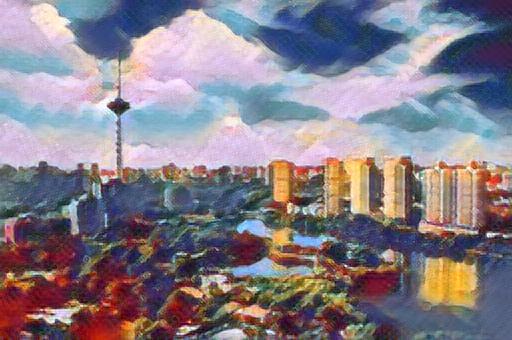}
}
\subfloat
{ 
\includegraphics[height=0.10\linewidth]{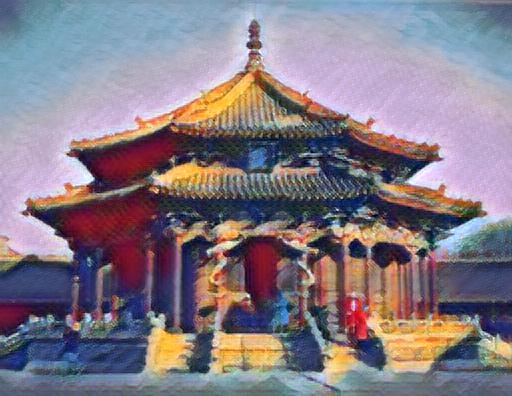}
}
\subfloat
{ 
\includegraphics[height=0.10\linewidth]{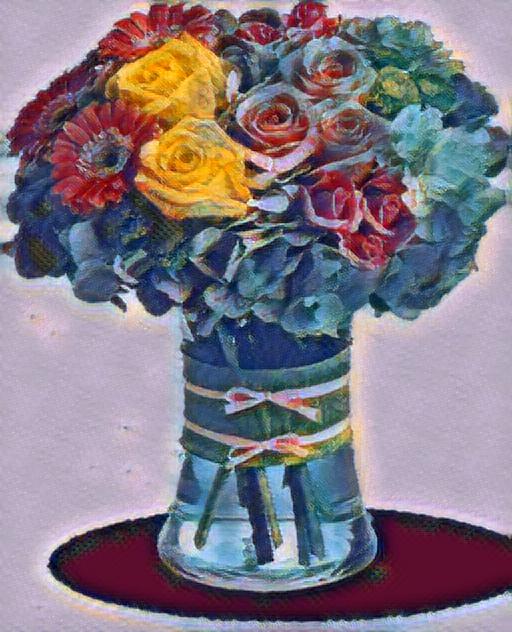}
}
\subfloat
{ 
\includegraphics[height=0.10\linewidth]{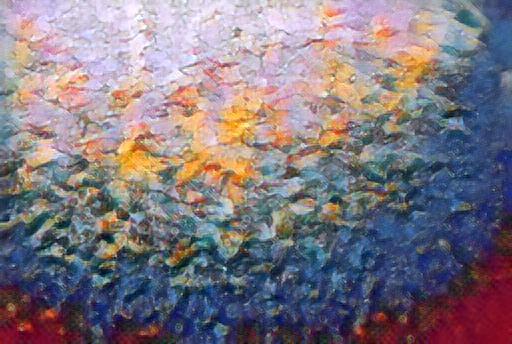}
}
%%%%%%%%%%%%%%%%%%%%%%%%%%%%%%%%%%%%
\end{center}
\caption{
Diverse images that are generated using a single MSG-Net-100 (2.3M parameters). First row shows the input content images and the other rows are generated images with different style targets (first column).}
\label{fig:9styles}
\end{figure*}

\paragraph{Qualitative Comparison} 
Our proposed MSG-Net achieves superior performance comparing to state-of-the-art generative network approaches as shown in Figure~\ref{fig:compare3}. One may argue that the arbitrary style work has better scalability/capacity~\cite{huang2017arbitrary,chen2016fast}. 
The style flexibility and image quality are always hard trade-off for generative model, and we particularly focus on the image quality in this work.  
More examples of the transfered images using MSG-Net are shown in Figure~\ref{fig:9styles}.

\begin{figure}
\subfloat
{
  \begin{overpic}[width=0.5\linewidth]{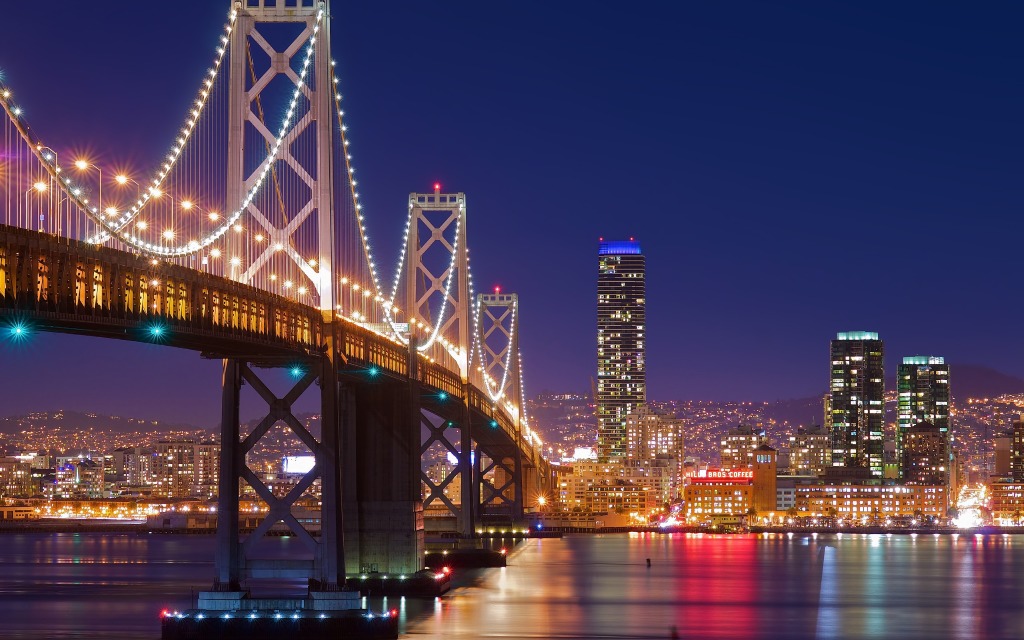}
     \put(65,25){\frame{\includegraphics[width=0.18\linewidth,height=0.2\linewidth]{figure/styles/Robert_Delaunay__1906__Portrait.jpg}} }
  \end{overpic}
}
\includegraphics[width=0.5\linewidth]{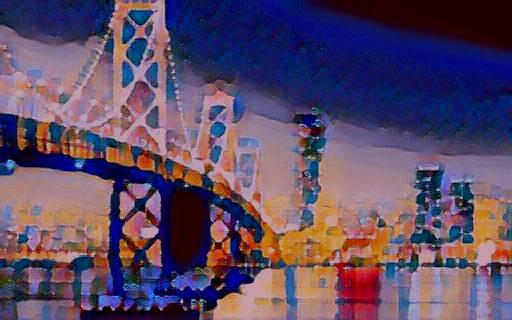}
\caption{Color control using MSG-Net, (left) content and style images, (right) color-preserved transfer result. }
\label{fig:color}
\vspace{-1em}
\end{figure}

\paragraph{Model Scalability. } Prior work using 1D style embedding has achieved  success in the scalability of style transfer towards the goal of arbitrary style transfer~\cite{huang2017arbitrary}. To test the scalability of MSG-Net, we augment the style set to 1K images, by adding 900 extra images from the wikiart.org~\cite{painter17}. We also build a larger model MSG-Net-1K with larger model capacity by increasing the width/channels of the model at mid stage (64$\times$64) by a factor of 2, resulting in 8.9M parameters. We also increase the training iterations by 4 times (320K) and follow the same training procedure as MSG-Net-100. We observe no quality degradation when increasing the number of styles (examples shown in the supplementary material).

\begin{figure}
\begin{minipage}[b]{0.36\linewidth}
\centering
\subfloat
{
\includegraphics[width=1.0\linewidth]{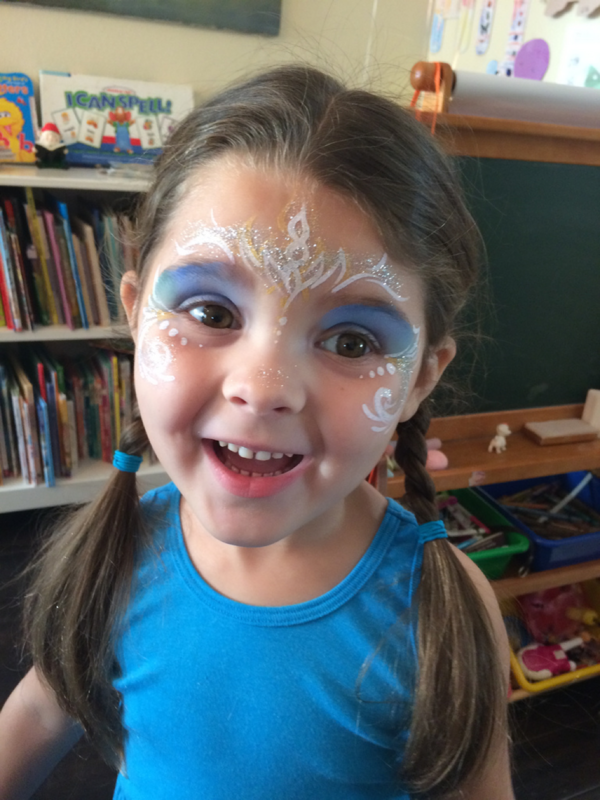}
}\\
\end{minipage}\hfill
%%%%%%%%%%%%%%%%%%%%%%%%%%%
\begin{minipage}[b]{0.26\linewidth}
\centering
\subfloat
{
\includegraphics[width=1.0\linewidth, height=\linewidth]{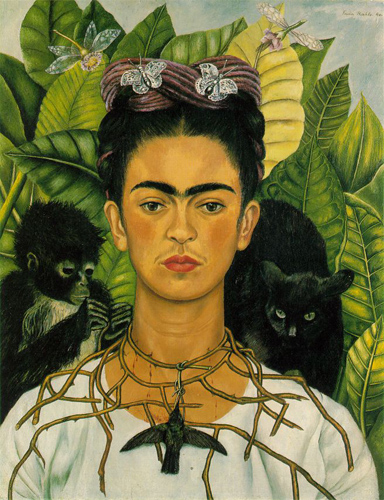}
}\\ \vspace{-0.8em}
\subfloat
{
\includegraphics[width=1.0\linewidth, height=0.8\linewidth]{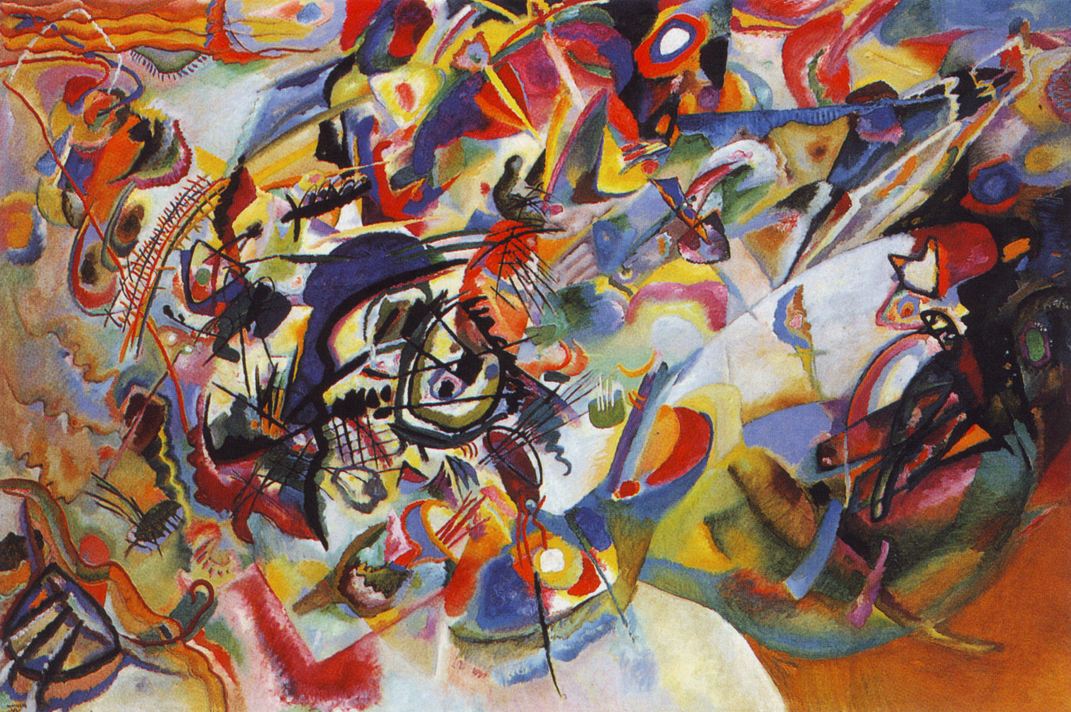}
}\\
\end{minipage}\hfill
%%%%%%%%%%%%%%%%%%%%%%%%%%%
\begin{minipage}[b]{0.36\linewidth}
\centering
\subfloat
{
\includegraphics[width=1.0\linewidth]{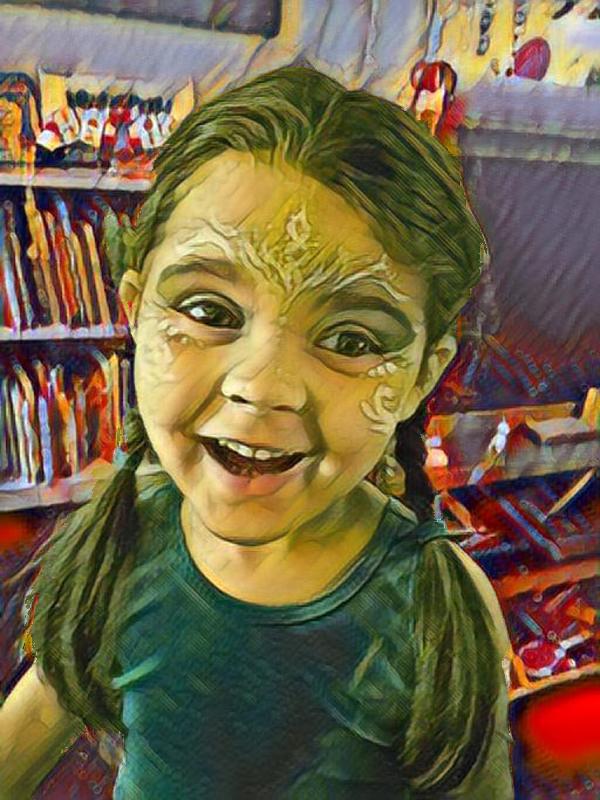}
}
\end{minipage}
\caption{Spatial control using MSG-Net. Left: input image, middle: foreground and background styles, right: style transfer result. (Input image and segmentation mask from Shen {\it et al.}~\cite{shen2016automatic}.) }
\label{fig:spatial}
\vspace{-1em}
\end{figure}

\subsection{Runtime Manipulation}

MSG-Net as a general approach for real-time style transfer is compatible with  existing  recent progress for both feed-forward and optimization methods, including but not limited to: content-style trade-off and interpolation (Figure~\ref{fig:interp}), color-preserving transfer (Figure~\ref{fig:color}), spatial manipulation (Figure~\ref{fig:spatial}) and brush stroke size control (Figure~\ref{fig:brush}\&\ref{fig:brush2}).  
For style interpolation, we use an affine interpolation of our style embedding following the prior work~\cite{dumoulin2016learned,huang2017arbitrary}. 
For color pre-serving, we match the color of style image with the content image as Gatys {\it et. al.}~\cite{gatys2016preserving}. Brush-size control has been discussed in the Section~\ref{sec:net}. We use the segmentation mask provided by Shen {\it et al.}~\cite{shen2016automatic} for spatial control.
The source code and technical detail of runtime manipulation will be included in our PyTorch implementation.

\section{Conclusion and Discussion}

To improve the quality and flexibility of generative models in style transfer, 
we introduce a novel CoMatch Layer that learns to match the second order statistics as image style representation. 
%We introduce an Upsample Convolution which avoids the checkerboard artifacts from fractionally-strided convolution. 
Multi-style Generative Network has achieved superior image quality comparing to state-of-the-art approaches. 
In addition, the proposed MSG-Net is compatible with most existing techniques and recent progress of stye transfer including style interpolation, color-preserving and spatial control. 
Moreover, MSG-Net first enables real-time brush-size control in a fully feed-forward manor. 
The compact MSG-Net-100 model has only 2.3M parameters and runs at more than 90 fps (frame/sec) on NVIDIA Titan Xp for the input image of size 256$\times$256 and at 15 fps  on a laptop GPU (GTX 750M-2GB).

\section*{Acknowledgment}
This work was supported by National Science Foundation award IIS-1421134. A GPU used for this research was donated by the NVIDIA Corporation.

{\small
\bibliographystyle{ieee}
\bibliography{egbib}

\begin{thebibliography}{10}\itemsep=-1pt

\bibitem{Burt83}
P.~Burt and E.~Adelson.
\newblock The laplacian pyramid as a compact image code.
\newblock {\em IEEE Transactions on communications}, 31(4):532--540, 1983.

\bibitem{che2016mode}
T.~Che, Y.~Li, A.~P. Jacob, Y.~Bengio, and W.~Li.
\newblock Mode regularized generative adversarial networks.
\newblock {\em arXiv preprint arXiv:1612.02136}, 2016.

\bibitem{Chen2017StyleBank}
D.~Chen, L.~Yuan, J.~Liao, N.~Yu, and G.~Hua.
\newblock Stylebank: An explicit representation for neural image style
  transfer.
\newblock In {\em IEEE Conference on Computer Vision and Pattern Recognition
  (CVPR)}, 2017.

\bibitem{chen2015mxnet}
T.~Chen, M.~Li, Y.~Li, M.~Lin, N.~Wang, M.~Wang, T.~Xiao, B.~Xu, C.~Zhang, and
  Z.~Zhang.
\newblock Mxnet: A flexible and efficient machine learning library for
  heterogeneous distributed systems.
\newblock {\em arXiv preprint arXiv:1512.01274}, 2015.

\bibitem{chen2016fast}
T.~Q. Chen and M.~Schmidt.
\newblock Fast patch-based style transfer of arbitrary style.
\newblock {\em arXiv preprint arXiv:1612.04337}, 2016.

\bibitem{collobert2011torch7}
R.~Collobert, K.~Kavukcuoglu, and C.~Farabet.
\newblock Torch7: A matlab-like environment for machine learning.
\newblock In {\em BigLearn, NIPS Workshop}, number EPFL-CONF-192376, 2011.

\bibitem{Debonet97}
J.~S. De~Bonet.
\newblock Multiresolution sampling procedure for analysis and synthesis of
  texture images.
\newblock In {\em Proceedings of the 24th annual conference on Computer
  graphics and interactive techniques}, pages 361--368. ACM
  Press/Addison-Wesley Publishing Co., 1997.

\bibitem{painter17}
S.~Y. Duck.
\newblock Painter by numbers.
\newblock \url{https://www.kaggle.com/c/painter-by-numbers}, 2016.

\bibitem{dumoulin2016learned}
V.~Dumoulin, J.~Shlens, and M.~Kudlur.
\newblock A learned representation for artistic style.
\newblock 2016.

\bibitem{efros2001image}
A.~A. Efros and W.~T. Freeman.
\newblock Image quilting for texture synthesis and transfer.
\newblock In {\em Proceedings of the 28th annual conference on Computer
  graphics and interactive techniques}, pages 341--346. ACM, 2001.

\bibitem{efros1999texture}
A.~A. Efros and T.~K. Leung.
\newblock Texture synthesis by non-parametric sampling.
\newblock In {\em Computer Vision, 1999. The Proceedings of the Seventh IEEE
  International Conference on}, volume~2, pages 1033--1038. IEEE, 1999.

\bibitem{feichtenhofer2016convolutional}
C.~Feichtenhofer, A.~Pinz, and A.~Zisserman.
\newblock Convolutional two-stream network fusion for video action recognition.
\newblock In {\em Proceedings of the IEEE Conference on Computer Vision and
  Pattern Recognition}, pages 1933--1941, 2016.

\bibitem{gatys2015texture}
L.~Gatys, A.~S. Ecker, and M.~Bethge.
\newblock Texture synthesis using convolutional neural networks.
\newblock In {\em Advances in Neural Information Processing Systems}, pages
  262--270, 2015.

\bibitem{gatys2016preserving}
L.~A. Gatys, M.~Bethge, A.~Hertzmann, and E.~Shechtman.
\newblock Preserving color in neural artistic style transfer.
\newblock {\em arXiv preprint arXiv:1606.05897}, 2016.

\bibitem{gatys2016image}
L.~A. Gatys, A.~S. Ecker, and M.~Bethge.
\newblock Image style transfer using convolutional neural networks.
\newblock In {\em Proceedings of the IEEE Conference on Computer Vision and
  Pattern Recognition}, pages 2414--2423, 2016.

\bibitem{gatys2016controlling}
L.~A. Gatys, A.~S. Ecker, M.~Bethge, A.~Hertzmann, and E.~Shechtman.
\newblock Controlling perceptual factors in neural style transfer.
\newblock {\em arXiv preprint arXiv:1611.07865}, 2016.

\bibitem{goodfellow2014generative}
I.~Goodfellow, J.~Pouget-Abadie, M.~Mirza, B.~Xu, D.~Warde-Farley, S.~Ozair,
  A.~Courville, and Y.~Bengio.
\newblock Generative adversarial nets.
\newblock In {\em Advances in neural information processing systems}, pages
  2672--2680, 2014.

\bibitem{he2016deep}
K.~He, X.~Zhang, S.~Ren, and J.~Sun.
\newblock Deep residual learning for image recognition.
\newblock In {\em Proceedings of the IEEE Conference on Computer Vision and
  Pattern Recognition}, pages 770--778, 2016.

\bibitem{he2016identity}
K.~He, X.~Zhang, S.~Ren, and J.~Sun.
\newblock Identity mappings in deep residual networks.
\newblock In {\em European Conference on Computer Vision}, pages 630--645.
  Springer, 2016.

\bibitem{heeger1995pyramid}
D.~J. Heeger and J.~R. Bergen.
\newblock Pyramid-based texture analysis/synthesis.
\newblock In {\em Proceedings of the 22nd annual conference on Computer
  graphics and interactive techniques}, pages 229--238. ACM, 1995.

\bibitem{huang2017arbitrary}
X.~Huang and S.~Belongie.
\newblock Arbitrary style transfer in real-time with adaptive instance
  normalization.
\newblock {\em arXiv preprint arXiv:1703.06868}, 2017.

\bibitem{huang2016sgan}
X.~Huang, Y.~Li, O.~Poursaeed, J.~Hopcroft, and S.~Belongie.
\newblock Stacked generative adversarial networks.
\newblock {\em arXiv}, 2016.

\bibitem{isola2016image}
P.~Isola, J.-Y. Zhu, T.~Zhou, and A.~A. Efros.
\newblock Image-to-image translation with conditional adversarial networks.
\newblock {\em arXiv preprint arXiv:1611.07004}, 2016.

\bibitem{jing2017neural}
Y.~Jing, Y.~Yang, Z.~Feng, J.~Ye, and M.~Song.
\newblock Neural style transfer: A review.
\newblock {\em arXiv preprint arXiv:1705.04058}, 2017.

\bibitem{Johnson2016Perceptual}
J.~Johnson, A.~Alahi, and L.~Fei-Fei.
\newblock Perceptual losses for real-time style transfer and super-resolution.
\newblock In {\em European Conference on Computer Vision}, 2016.

\bibitem{kingma2014adam}
D.~Kingma and J.~Ba.
\newblock Adam: A method for stochastic optimization.
\newblock {\em arXiv preprint arXiv:1412.6980}, 2014.

\bibitem{kwatra2003graphcut}
V.~Kwatra, A.~Sch{\"o}dl, I.~Essa, G.~Turk, and A.~Bobick.
\newblock Graphcut textures: image and video synthesis using graph cuts.
\newblock In {\em ACM Transactions on Graphics (ToG)}, volume~22, pages
  277--286. ACM, 2003.

\bibitem{Li_2016_CVPR}
C.~Li and M.~Wand.
\newblock Combining markov random fields and convolutional neural networks for
  image synthesis.
\newblock In {\em The IEEE Conference on Computer Vision and Pattern
  Recognition (CVPR)}, June 2016.

\bibitem{li2016precomputed}
C.~Li and M.~Wand.
\newblock Precomputed real-time texture synthesis with markovian generative
  adversarial networks.
\newblock In {\em European Conference on Computer Vision}, pages 702--716.
  Springer, 2016.

\bibitem{lin2014microsoft}
T.-Y. Lin, M.~Maire, S.~Belongie, J.~Hays, P.~Perona, D.~Ramanan,
  P.~Doll{\'a}r, and C.~L. Zitnick.
\newblock Microsoft coco: Common objects in context.
\newblock In {\em European Conference on Computer Vision}, pages 740--755.
  Springer, 2014.

\bibitem{long2015fully}
J.~Long, E.~Shelhamer, and T.~Darrell.
\newblock Fully convolutional networks for semantic segmentation.
\newblock In {\em Proceedings of the IEEE Conference on Computer Vision and
  Pattern Recognition}, pages 3431--3440, 2015.

\bibitem{mahendran2015understanding}
A.~Mahendran and A.~Vedaldi.
\newblock Understanding deep image representations by inverting them.
\newblock In {\em Proceedings of the IEEE Conference on Computer Vision and
  Pattern Recognition}, pages 5188--5196, 2015.

\bibitem{odena2016deconvolution}
A.~Odena, V.~Dumoulin, and C.~Olah.
\newblock Deconvolution and checkerboard artifacts.
\newblock {\em Distill}, 2016.

\bibitem{pytorch}
A.~Paszke, S.~Gross, S.~Chintala, G.~Chanan, E.~Yang, Z.~DeVito, Z.~Lin,
  A.~Desmaison, L.~Antiga, and A.~Lerer.
\newblock Automatic differentiation in pytorch.
\newblock 2017.

\bibitem{portilla2000parametric}
J.~Portilla and E.~P. Simoncelli.
\newblock A parametric texture model based on joint statistics of complex
  wavelet coefficients.
\newblock {\em International journal of computer vision}, 40(1):49--70, 2000.

\bibitem{radford2015unsupervised}
A.~Radford, L.~Metz, and S.~Chintala.
\newblock Unsupervised representation learning with deep convolutional
  generative adversarial networks.
\newblock {\em arXiv preprint arXiv:1511.06434}, 2015.

\bibitem{shen2016automatic}
X.~Shen, A.~Hertzmann, J.~Jia, S.~Paris, B.~Price, E.~Shechtman, and I.~Sachs.
\newblock Automatic portrait segmentation for image stylization.
\newblock In {\em Computer Graphics Forum}, volume~35, pages 93--102. Wiley
  Online Library, 2016.

\bibitem{simoncelli1995steerable}
E.~P. Simoncelli and W.~T. Freeman.
\newblock The steerable pyramid: A flexible architecture for multi-scale
  derivative computation.
\newblock In {\em Image Processing, 1995. Proceedings., International
  Conference on}, volume~3, pages 444--447. IEEE, 1995.

\bibitem{simonyan2014very}
K.~Simonyan and A.~Zisserman.
\newblock Very deep convolutional networks for large-scale image recognition.
\newblock {\em arXiv preprint arXiv:1409.1556}, 2014.

\bibitem{sindagi2017generating}
V.~A. Sindagi and V.~M. Patel.
\newblock Generating high-quality crowd density maps using contextual pyramid
  cnns.
\newblock In {\em 2017 IEEE International Conference on Computer Vision
  (ICCV)}. IEEE, 2017.

\bibitem{ulyanov2016texture}
D.~Ulyanov, V.~Lebedev, A.~Vedaldi, and V.~Lempitsky.
\newblock Texture networks: Feed-forward synthesis of textures and stylized
  images.
\newblock In {\em Int. Conf. on Machine Learning (ICML)}, 2016.

\bibitem{ulyanov2017improved}
D.~Ulyanov, A.~Vedaldi, and V.~Lempitsky.
\newblock Improved texture networks: Maximizing quality and diversity in
  feed-forward stylization and texture synthesis.
\newblock {\em arXiv preprint arXiv:1701.02096}, 2017.

\bibitem{wei2000fast}
L.-Y. Wei and M.~Levoy.
\newblock Fast texture synthesis using tree-structured vector quantization.
\newblock In {\em Proceedings of the 27th annual conference on Computer
  graphics and interactive techniques}, pages 479--488. ACM
  Press/Addison-Wesley Publishing Co., 2000.

\bibitem{zhang2017image}
H.~Zhang, V.~Sindagi, and V.~M. Patel.
\newblock Image de-raining using a conditional generative adversarial network.
\newblock {\em arXiv preprint arXiv:1701.05957}, 2017.

\bibitem{zhang2016stackgan}
H.~Zhang, T.~Xu, H.~Li, S.~Zhang, X.~Huang, X.~Wang, and D.~Metaxas.
\newblock Stackgan: Text to photo-realistic image synthesis with stacked
  generative adversarial networks.
\newblock {\em arXiv preprint arXiv:1612.03242}, 2016.

\bibitem{zhang2010non}
H.~Zhang, J.~Yang, Y.~Zhang, and T.~S. Huang.
\newblock Non-local kernel regression for image and video restoration.
\newblock In {\em European Conference on Computer Vision}, pages 566--579.
  Springer, 2010.

\end{thebibliography}
}

\end{document}